%% file: main.tex
\documentclass[runningheads]{llncs}

\usepackage{eccvabbrv}

\usepackage{graphicx}
\usepackage{booktabs}
\usepackage[table]{xcolor}
\usepackage{multirow}
\usepackage{amsmath}
\usepackage{amssymb}
\usepackage{bm}
\usepackage{enumitem}
\usepackage{tikz}
\usetikzlibrary{shapes.geometric, arrows.meta, positioning, fit, backgrounds, calc}
\usepackage[colorlinks=true, urlcolor=blue, linkcolor=blue, citecolor=red]{hyperref}
\usepackage{algorithm}
\usepackage{algpseudocode}
\usepackage{longtable}
\usepackage{tabularx}
\usepackage{tcolorbox}
\usepackage[font=small]{caption}
\usepackage{float}

\newcommand{\mat}[1]{\mathbf{#1}}
\newcommand{\op}[1]{\operatorname{#1}}
\newcommand{\mR}{\mathbb{R}}

\usepackage{xspace}
\newcommand{\WorldSGG}{World Scene Graph Generation\xspace}
\newcommand{\WSGG}{WSGG\xspace}
\newcommand{\PWG}{PWG\xspace}
\newcommand{\MWAE}{MWAE\xspace}
\newcommand{\FDST}{4DST\xspace}
\newcommand{\PiThree}{$\pi^3$\xspace}

\usepackage{pifont}
\newcommand{\cmark}{\ding{51}}
\newcommand{\xmark}{\ding{55}}

\usepackage[accsupp]{axessibility}  

\usepackage{hyperref}

\usepackage{orcidlink}

\makeatletter
\renewcommand{\tableofcontents}{%
  \section*{\contentsname}%
  \def\l@title##1##2{}%
  \def\l@author##1##2{}%
  \def\authcount##1{}%
  \def\lastand{}%
  \def\and{}%
  \@starttoc{toc}%
}
\makeatother

\begin{document}
\raggedbottom

\title{Towards Spatio-Temporal World Scene Graph Generation from Monocular Videos} 

\titlerunning{World Scene Graph Generation}

\author{Rohith Peddi\inst{1} \and
Saurabh \inst{2} \and
Shravan Shanmugam \inst{1} \and
Likhitha Pallapothula \inst{1} \and
Yu Xiang \inst{1} \and
Parag Singla \inst{2} \and
Vibhav Gogate \inst{1}}

\authorrunning{Peddi et al.}

\institute{The University of Texas at Dallas, Richardson, TX 75080, USA \and
Indian Institute of Technology Delhi, New Delhi, India}

\maketitle

\setcounter{tocdepth}{-10}

\vspace{-6mm}
\begin{abstract}

  Spatio-temporal scene graphs provide a principled representation for modeling
  evolving object interactions, yet existing methods remain fundamentally
  frame-centric: they reason only about currently visible objects, discard
  entities upon occlusion, and operate in 2D.  To address this, we first introduce 
  \textbf{ActionGenome4D}, a dataset that upgrades Action Genome videos into 4D scenes via feed-forward 3D
  reconstruction, world-frame oriented bounding boxes for every object involved in actions, and
  dense relationship annotations including for objects that are temporarily
  unobserved due to occlusion or camera motion.  Building on this data, we
  formalize \textbf{\WorldSGG (\WSGG{})}, the task of
  constructing a world scene graph at each timestamp that encompasses all 
  interacting objects in the scene, both observed and unobserved\footnote{code: \href{https://github.com/rohithpeddi/WorldSGG}{https://github.com/rohithpeddi/WorldSGG}.}.  We then propose three
  complementary methods, each exploring a different inductive bias for
  reasoning about unobserved objects: \textbf{\PWG{}} (Persistent World Graph),
  which implements object permanence via a zero-order feature buffer;
  \textbf{\MWAE{}} (Masked World Auto-Encoder), which reframes unobserved-object
  reasoning as masked completion with cross-view associative retrieval; and
  \textbf{\FDST{}} (4D Scene Transformer), which replaces the static buffer
  with differentiable per-object temporal attention enriched by 3D motion
  and camera-pose features. We further design and evaluate the performance of strong open-source Vision--Language Models on the \WSGG{} task via a suite of 
  Graph RAG-based approaches, establishing baselines for unlocalized
  relationship prediction. \WSGG{} thus advances video scene understanding toward world-centric, 
  temporally persistent, and interpretable scene reasoning.
  \vspace{-2mm}
  \keywords{\WorldSGG \and Video Scene Graph Generation \and Scene Understanding \and Spatio-Temporal Reasoning}
\end{abstract}

\vspace{-8mm}
\section{Introduction}
\vspace{-2mm}
\input{paper_tex_files/introduction}

\vspace{-6mm}
\section{Related work}
\vspace{-2mm}
\input{paper_tex_files/related_work}

\vspace{-5mm}
\section{Notation \& Problem Description}
\vspace{-2mm}
\input{paper_tex_files/notation}

\vspace{-5mm}
\section{ActionGenome4D Dataset}
\vspace{-2mm}
\input{paper_tex_files/scene_construction}
\label{sec:scene_construction}

\vspace{-4mm}
\section{World Scene Graph Generation}
\label{sec:wsgg_methods}
\vspace{-2mm}
\input{paper_tex_files/experiments_wsgg}

\vspace{-4mm}
\section{MLLMs on Unlocalized World Scene Graph Generation}
\label{sec:mllm_wsgg}
\vspace{-2mm}
\input{paper_tex_files/experiments_mllm}

\vspace{-2mm}
\section{Conclusion \& Future Work}
\vspace{-2mm}
\input{paper_tex_files/conclusion}

\newpage

\appendix

\renewcommand{\theHsection}{appendix.\Alph{section}}
\renewcommand{\theHsubsection}{appendix.\Alph{section}.\arabic{subsection}}

\begin{center}
\Large\bfseries Supplementary Material
\end{center}
\vspace{4mm}

\setcounter{tocdepth}{3}
\tableofcontents
\newpage

\section{Motivation}
\input{sup_tex_files/sup_motivation}

\newpage

\section{Notation}
\input{sup_tex_files/sup_notation}
\newpage

\section{Related Work}
\input{sup_tex_files/sup_extended_related_work}
\newpage

\section{3D Scene Construction Details}
\input{sup_tex_files/sup_scene_construction}
\newpage

\section{Geometric Annotation Details}
\input{sup_tex_files/sup_geometric_annotation}
\newpage

\section{Semantic Annotation Details}
\input{sup_tex_files/sup_semantic_annotation}
\newpage

\section{Manual Correction of 3D BBox Annotations}
\input{sup_tex_files/sup_manual_correction_3dbbox}
\newpage

\section{Manual Correction of WSG Annotations}
\input{sup_tex_files/sup_manual_correction_wsg}
\newpage

\section{Action Genome 4D Statistics}
\input{sup_tex_files/sup_ag4d_statistics}
\newpage

\section{Monocular 3D Detection Pipeline}
\input{sup_tex_files/sup_monocular_3d}
\newpage

\section{World Scene Graph Generation}
\input{sup_tex_files/sup_experiments_wsgg}
\newpage

\section{MLLMs for Unlocalized WSG Generation}
\input{sup_tex_files/sup_mllm_wsg}
\newpage

\section{Conclusion}
\input{sup_tex_files/sup_conclusion}
\newpage

\bibliographystyle{splncs04}
\bibliography{main}
\end{document}

%% file: paper_tex_files/introduction.tex

Scene graphs offer a structured, compact, and interpretable representation of visual scenes by
encoding objects as nodes and their pairwise relationships as
edges~\cite{krishna2016visualgenomeconnectinglanguage}. While extensive progress has been made in generating scene
graphs from individual images and, more recently, from video~\cite{ji2019actiongenomeactionscomposition,cong2021spatialtemporaltransformerdynamicscene}, the predominant paradigm remains
fundamentally \emph{frame-centric}: models process each frame (or a short temporal window)
independently, predicting a flat, 2D scene graph that is neither grounded in 3D space nor temporally
consistent across the video. When an object leaves the camera's field of view or becomes occluded,
it simply vanishes from the graph---a significant limitation for downstream tasks in robotics.


\begin{figure}[!t]
    \centering
    \includegraphics[width=\textwidth]{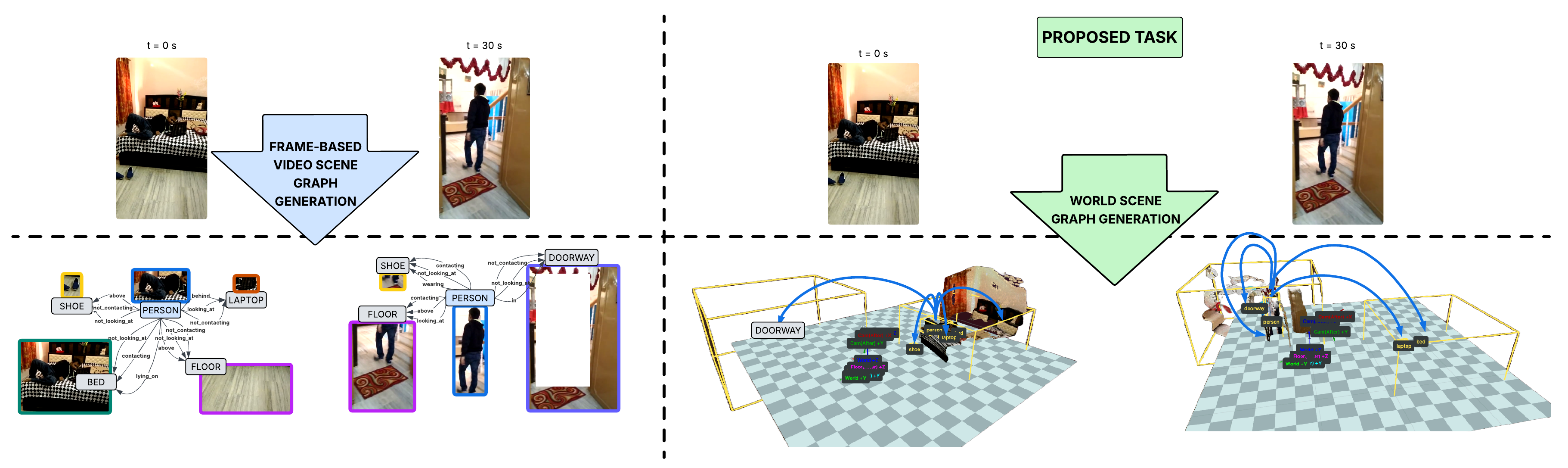}
    \vspace{-8mm}
    \caption{\textbf{World Scene Graph Generation (\WSGG{}).} Unlike standard Video Scene Graph Generation (left) 
    that is constrained to the instantaneous camera view and discards objects once they exit the frame or get occluded, 
    our proposed task (right) grounds scene understanding in a global 3D world coordinate frame. 
    Models trained on \WSGG{} output a comprehensive world scene graph at each timestamp containing all the objects 
    in the environment. As shown at $t=30$s, out-of-view objects (e.g., bed, laptop) remain perfectly 
    localized in 3D, enabling global, view-independent interpretable scene reasoning. 
    Blue curves drawn from person to objects (right) represent relationships.}
    \label{fig:world_sgg_task_overview}
    \vspace{-6mm}
\end{figure}

This frame-centric view (left, Figure~\ref{fig:world_sgg_task_overview}) is at odds with how real-world agents perceive and act. An embodied agent
moves and interacts with objects in a scene, accumulating observations over time, maintaining a
persistent semantic memory of existing objects, and reasoning about how spatial relationships evolve
even for objects it can no longer see. Developmental psychology has long recognized \emph{object
permanence}~\cite{Spelke1990ObjectPerception} as a foundational prerequisite for physical reasoning: the
understanding that objects continue to exist when they leave our direct perception. Achieving such
world-centric scene understanding (right, Figure~\ref{fig:world_sgg_task_overview}) from monocular video requires three capabilities that existing
datasets and benchmarks do not jointly provide: (i)~3D spatial grounding of every object in a common
world coordinate frame, (ii)~temporally consistent object identity and tracking across frames, and
(iii)~dense semantic annotations between all interacting objects, including \emph{unobserved}
objects that are contextually present in the scene but not visible in a given frame due to
occlusions or camera movement.


We aim to bridge this gap with two main contributions: a new dataset and a new task. We introduce
\textbf{ActionGenome4D}, upgrading Action Genome~\cite{ji2019actiongenomeactionscomposition} into
a \emph{4D spatio-temporal scene representation} via a foundation-model-driven pipeline:
(a)~\textbf{3D Scene Construction} reconstructs per-frame point clouds from monocular video
using \PiThree{}~\cite{wang2025pi};
(b)~\textbf{Geometric Annotation} produces world-frame oriented 3D bounding boxes by combining
GDINO~\cite{liu2024groundingdinomarryingdino} detection, dual-mode
SAM2~\cite{ravi2024sam2segmentimages} segmentation, and floor-aligned OBB fitting
(Section~\ref{sec:scene_construction});
(c)~\textbf{Semantic Annotation} generates dense relationship pseudo-annotations for
\emph{unobserved objects} via a RAG-based VLM pipeline with discriminative verification,
followed by manual correction (Section~\ref{sec:mllm_wsgg}).


To this end, we formalize the task of \textbf{\WSGG{}}, which generalizes conventional video scene
graph generation from per-frame graphs to temporally persistent, world-anchored scene graphs that
account for all interacting objects in the world state. Thus, \WSGG{} requires a model to
(a)~localize all objects via 3D oriented bounding boxes in a shared world frame and (b)~predict all
pairwise relationships between interacting objects, including those that remain unobserved. We
propose three complementary methods, each exploring a different inductive bias for reasoning about
the unobserved objects: (a) \textbf{\PWG{}} (Persistent World Graph): embodies the principle of
object permanence by maintaining a memory buffer that retains each object's visual features from its
last observed appearance, enabling relationship prediction over the complete world graph even when
objects leave the camera's field of view. (b) \textbf{\MWAE{}} (Masked World Auto-Encoder): treats
occlusions and camera motion as natural masking over an object-level auto-encoder. An associative
retriever trained with cross-view reconstruction and simulated-occlusion objectives, reconstructs
unobserved object representations. (c) \textbf{\FDST{}} (4D Scene Transformer): replaces the static
buffer with a differentiable temporal transformer that jointly attends over observed and unobserved
object tokens across the full video.

We further evaluate open-source VLMs on unlocalized \WSGG{}, mainly to assess their ability to
predict relationships for all interacting objects in the scene via a suite of Graph-RAG based
approaches. Thus, our contributions are:
\vspace{-2mm}
\begin{enumerate}
    \item \textbf{ActionGenome4D dataset:} A systematic extension of
    Action Genome into 4D, combining automated pseudo-label generation with
    manual correction to provide 3D-grounded oriented bounding boxes for all interacting objects,
    dense semantic annotations covering unobserved object relationships,
    and temporally aligned per-frame 3D scenes with camera poses.
    \item \textbf{\WSGG{} task:} A new task that generalizes
    video scene graph generation to world-anchored, temporally coherent
    scene graphs over all interacting objects (both observed and
    unobserved) from monocular video.
    \item \textbf{Three \WSGG{} methods (\PWG{}, \MWAE{}, and \FDST{}):} Complementary approaches
    exploring different inductive biases for unobserved object reasoning,
    with shared architectural components and comprehensive ablations.
    \item \textbf{VLM evaluation:} Systematic assessment of
    open-source VLMs on the \WSGG{} task, establishing baselines for
    unlocalized relationship prediction.
\end{enumerate}

%% file: paper_tex_files/related_work.tex


\begin{table*}[!t]
\centering
\vspace{-2mm}
\caption{\textbf{SGG Tasks.} \textbf{Bold} entries highlight the unique capabilities of \WSGG{}.}
\label{tab:sgg_task_comparison}
\small
\setlength{\tabcolsep}{4pt}
\resizebox{\textwidth}{!}{%
\begin{tabular}{@{}l c c c c c c c c@{}}
\toprule
\textbf{Task} & \textbf{Input} & \textbf{Spatial} & \textbf{Temporal} & \textbf{Localization} & \textbf{Object Scope} & \textbf{Coord.} & \textbf{Unobs.} & \textbf{Rel.} \\
 &  & \textbf{Dim.} &  &  &  & \textbf{Frame} & \textbf{Objects} & \textbf{Persistence} \\
\midrule
Image SGG         & Image       & 2D & \xmark & 2D BBox & Detected         & Image & \xmark & Per-frame \\
Video SGG         & Video       & 2D & \cmark & 2D BBox & Detected / frame & Image & \xmark & Cross-frame \\
3D SGG            & 3D Scan     & 3D & \xmark & 3D BBox & All in scan      & Scene & \xmark & Per-scan \\
4D SGG            & Video + 3D  & 3D & \cmark & 3D BBox & Detected / frame & Scene & \xmark & Cross-frame \\
Panoptic SGG      & Image       & 2D & \xmark & 2D Mask & Things + Stuff   & Image & \xmark & Per-frame \\
Panoptic VSGG     & Video       & 2D & \cmark & 2D Mask & Things + Stuff   & Image & \xmark & Cross-frame \\
\midrule
\textbf{\WSGG{} (ours)} & \textbf{Video} & \textbf{3D} & \textbf{\cmark} & \textbf{3D BBox} & \textbf{World state} & \textbf{World} & \textbf{\cmark} & \textbf{Through occ.} \\
\bottomrule
\end{tabular}%
}
\vspace{-7mm}
\end{table*}

\noindent \textbf{Structured Scene Understanding:} Scene graph generation has branched into several
task
variants. Table~\ref{tab:sgg_task_comparison} contrasts these formulations along eight
axes\footnote{Object Scope: \emph{Things}: countable objects; \emph{Stuff}: amorphous regions (e.g.,
wall, floor). 4D SGG methods typically require RGB-D or multi-view input beyond video.}. Scene
graphs were first introduced as structured representations that encode objects and their pairwise
relationships for static images~\cite{krishna2016visualgenomeconnectinglanguage,chen2019knowledgeembeddedroutingnetworkscene,zhu2022scenegraphgenerationcomprehensive}. Extending
scene graphs to the temporal domain requires modeling how relationships evolve across frames. The
Video Visual Relation Detection (VidVRD) benchmark~\cite{xindi_et_al_vid_vrd_2017} first formalised
this task, while the Action Genome (AG) dataset~\cite{ji2019actiongenomeactionscomposition}
introduced compositional spatio-temporal scene graphs grounded in human--object interactions.
Following AG, several Transformer-based~\cite{cong2021spatialtemporaltransformerdynamicscene,feng2022exploitinglongtermdependenciesgenerating,nag2023unbiasedscenegraphgeneration,peddi2025unbiasedrobustspatiotemporalscene} methods have been
proposed to generate dynamic scene graphs for VidSGG. 


Departing from the 2D setting, 3D scene graphs aim to encode objects and their relationships in a
shared world coordinate frame. These can be thought of as a generalization of 2D scene graphs to the
3D domain. 3D scene graphs were pioneered~\cite{Kim20203DSceneGraph} as sparse, semantic
representations of physical environments for intelligent agents. 3D spatial multimodal knowledge
accumulation~\cite{Feng20233DSM3DSpatialMultimodal} advanced scene graph prediction in point clouds.
Following this, several works have explored 3D scene graph generation from point
clouds~\cite{ma2024heterogeneous,koch2023sgrec3dselfsupervised3dscene,zhang2025openvocabularyfunctional3dscene,Nyffeler_2025_ICCV}. Moving forward, 4D scene graphs extend the 3D domain to the temporal dimension,
reasoning about objects and their relationships in a shared world coordinate frame across time. 4D
panoptic scene graph generation~\cite{yang20244dpanopticscenegraph} extends scene graphs to the joint
spatial-temporal-panoptic domain. Learning 4D panoptic scene graphs from rich 2D visual scenes was
explored in~\cite{wu2025learning4dpanopticscene}. RealGraph~\cite{Lin2023RealGraphMultiview} contributes a multi-view
dataset for 4D real-world context graph generation.

\noindent \textbf{Vision-Language Models (VLMs) for Spatial Understanding:} VLMs have recently been
adapted for spatial understanding tasks that require reasoning about 3D geometry and object
relationships from visual input. SpatialVLM~\cite{chen2024spatialvlmendowingvisionlanguagemodels}
endowed VLMs with spatial reasoning capabilities through internet-scale spatial data collection and
training. SpatialRGPT~\cite{cheng2024spatialrgptgroundedspatialreasoning} grounds spatial reasoning
in VLMs through depth-aware region representations.
SpatialBot~\cite{cai2025spatialbotprecisespatialunderstanding} enables precise spatial understanding
with depth integration. SD-VLM~\cite{chen2025sdvlmspatialmeasuringunderstanding} encodes depth
information directly into VLMs for spatial measuring. To further enhance spatial reasoning
capabilities of VLMs, several works focus on improving VLM spatial capabilities through specialized
training or architectural design~\cite{shen2026finegrainedpreferenceoptimizationimproves,lee2025perspectiveawarereasoningvisionlanguagemodels}. 
See\&Trek~\cite{li2025seetrektrainingfreespatialprompting} introduces training-free spatial
prompting of multimodal LLMs. Moreover, several works such as
SITE~\cite{wang2025sitespatialintelligencethorough},
STI-Bench~\cite{li2025stibenchmllmsreadyprecise}, and
CoSpace~\cite{zhu2025cospacebenchmarkingcontinuousspace} provide thorough evaluation of spatial
intelligence and test whether MLLMs are ready for precise spatial-temporal world understanding.



A growing line of work endows VLMs with explicit 3D understanding.
SpatialLLM~\cite{ma2025spatialllmcompound3dinformeddesign} is a compound 3D-informed design towards
spatially-intelligent large multimodal models.
SpatialReasoner~\cite{ma2025spatialreasonerexplicitgeneralizable3d} enables explicit and
generalizable 3D spatial reasoning. SpatialLM~\cite{mao2025spatiallmtraininglargelanguage} trains
large language models for structured indoor modeling. Scalable spatial intelligence via 2D-to-3D
data lifting was proposed in~\cite{miao2025scalablespatialintelligence2dto3d}.
LayoutVLM~\cite{sun2025layoutvlmdifferentiableoptimization3d} introduces differentiable optimization
of 3D layout via vision-language models. MindJourney~\cite{yang2025mindjourneytesttimescalingworld}
leverages test-time scaling with world models for spatial reasoning.




%% file: paper_tex_files/notation.tex
We present the proposed \WSGG{} task following the formulation of
\cite{peddi2024scenegraphanticipation}. Given an input video segment $V_{1}^{T} =
\{I^{t}\}_{t=1}^{T}$ of $T$ monocular frames, we define the \emph{world state} $\mathcal{W}^{t}$ at
timestamp $t$ as the complete set of objects that exist in the scene at that instant. The world
state is partitioned into two disjoint subsets: $\mathcal{W}^{t} = \mathcal{O}^{t} \;\cup\;
\mathcal{U}^{t}, \mathcal{O}^{t} \cap \mathcal{U}^{t} = \emptyset$, where $\mathcal{O}^{t} =
\{w_{k}^{t}\}_{k=1}^{N(t)}$ is the set of \emph{observed} objects that are visible in frame $I^{t}$,
and $\mathcal{U}^{t} = \{w_{k}^{t}\}_{k=N(t)+1}^{M(t)}$ is the set of \emph{unobserved} objects that
are not visible due to occlusions, camera motion, or other factors. Here $M(t)$ denotes the total
number of objects in the world state and $N(t)$ the number of observed objects at timestamp $t$. We
operate under the assumption that all objects both observed and unobserved always persist in the
world state. We define a binary visibility indicator $\text{vis}(k,t) \in \{0,1\}$ that equals $1$
if and only if object $w_{k}^{t} \in \mathcal{O}^{t}$ (i.e., the object is observed in frame
$I^{t}$). The camera extrinsic parameters at $t$ are given by the SE(3) transformation
$\mathbf{T}^{t} = [\mathbf{R}^{t} \mid \boldsymbol{\tau}^{t}] \in \mathbb{R}^{4 \times 4}$.

\vspace{1mm}
\noindent\textbf{Objects.} Let $\mathcal{C}$ denote the set of all object categories.
Each object $w_{k}^{t} \in \mathcal{W}^{t}$ is characterised by its category $c_{k}^{t} \in
\mathcal{C}$, a \emph{3D oriented bounding box} $\mathbf{b}_{k}^{t} \in \mathbb{R}^{8 \times 3}$
given as eight corner vertices in world coordinate frame, and for observed objects, a \emph{2D
bounding box} $\mathbf{d}_{k}^{t} \in \mathbb{R}^{4}$ in image coordinates. Because the persistent
geometric scaffold provides world-frame grounding at all times, 3D boxes are available for every
object in $\mathcal{W}^{t}$, whereas 2D boxes are defined only for $\mathcal{O}^{t}$.

\vspace{1mm}
\noindent\textbf{Relationships.} Let $\mathcal{P}$ be the set comprising all predicate classes.
Each pair of objects $\left(w_{i}^{t},\, w_{j}^{t}\right)$ in the world state may exhibit multiple
relationships, defined through predicates $\{p_{ijk}^{t}\}_{k}$ where $p_{ijk}^{t} \in \mathcal{P}$.
We define a relationship instance $r_{ijk}^{t}$ as a triplet $\left(w_{i}^{t},\, p_{ijk}^{t},\,
w_{j}^{t}\right)$ that combines two distinct objects and a predicate.

\vspace{1mm}
\noindent\textbf{Scene Graphs.} We consider scene graph as a symbolic representation of interacting
objects
and their pair-wise relationships. The \emph{frame-level scene graph} $\mathcal{G}^{t}$ is the set
of all relationship triplets among observed objects: $\mathcal{G}^{t} = \{r_{ijk}^{t} \mid
w_{i}^{t}, w_{j}^{t} \in \mathcal{O}^{t}\}$. The \emph{world scene graph}
$\mathcal{G}_{\mathcal{W}}^{t}$ is the set of all relationship triplets over the full world state:
$\mathcal{G}_{\mathcal{W}}^{t} = \{r_{ijk}^{t} \mid w_{i}^{t}, w_{j}^{t} \in \mathcal{W}^{t}\}$. For
each object $w_{i}^{t}$ and a pair of objects $\left(w_{i}^{t},\, w_{j}^{t}\right)$, we use
$\hat{\mathbf{c}}_{i}^{t} \in \left[0,1\right]^{|\mathcal{C}|}$ and $\hat{\mathbf{p}}_{ij}^{t} \in
\left[0,1\right]^{|\mathcal{P}|}$ to represent the predicted distributions over object categories
and predicate classes respectively. Here $\sum_{k} \hat{c}_{ik}^{t} = 1$ and $\sum_{k}
\hat{p}_{ijk}^{t} = 1$. Each observed object $w_{k}^{t} \in \mathcal{O}^{t}$ is associated with a
visual feature vector $\mathbf{f}_{k}^{t} \in \mathbb{R}^{d_{\text{roi}}}$.

\vspace{1mm}
\noindent\textbf{Problem Description.} We formally define the tasks of Video Scene Graph Generation
(VidSGG)
and \WorldSGG{} (\WSGG{}) as follows:

\begin{itemize}[nosep]
    \item [--] The goal of \textbf{VidSGG} is to build frame-level scene graphs
    $\{\mathcal{G}^{t}\}_{t=1}^{T}$ for the observed video segment
    $V_{1}^{T} = \{I^{t}\}_{t=1}^{T}$.  It entails the detection of observed objects
    $\{w_{k}^{t}\}_{k=1}^{N(t)}$ in each frame and the prediction of all pair-wise relationships
    $\{r_{ijk}^{t}\}_{ijk}$ between detected objects.
    \item [--] The goal of \textbf{\WSGG{}} is to construct the world scene graph
    $\mathcal{G}_{\mathcal{W}}^{t}$ at each timestamp $t$, given the monocular frames
    $V_{1}^{T}$ of a dynamic scene.  This entails:
    \begin{enumerate}[label=(\alph*)]
        \item \emph{3D Localization}: Estimating the 3D oriented bounding box $\mathbf{b}_{k}^{t}$
        in the world coordinate frame for each object $w_{k}^{t} \in \mathcal{W}^{t}$.
        \item \emph{Semantic Relationship Prediction}: Predicting all pair-wise relationships
        $\{r_{ijk}^{t}\}_{ijk}$ among all interacting objects in $\mathcal{W}^{t}$, encompassing
        interactions
        between observed-observed, observed-unobserved, and unobserved-unobserved object pairs.
    \end{enumerate}
\end{itemize}


\noindent \textbf{Graph Building Strategies.} Following VidSGG conventions, we consider two
graph construction strategies:
\begin{itemize}[nosep]
    \item [--] \textbf{With Constraint:} Each object pair
    $\left(w_{i}^{t},\, w_{j}^{t}\right)$ is assigned exactly one predicate
    $p^t_{ij}$, yielding a single-edge graph
    $\mathcal{G}_{\mathcal{W}}^t = \{r_{ij}^t\}_{ij}$.
    \item [--] \textbf{No Constraint:} Each object pair may carry multiple
    predicates, producing a multi-edge graph
    $\mathcal{G}_{\mathcal{W}}^t = \{r_{ijk}^{t}\}_{ijk}$ that captures
    richer relational structure.
\end{itemize}

%% file: paper_tex_files/scene_construction.tex

\begin{figure}[!t]
\centering
\includegraphics[width=\textwidth]{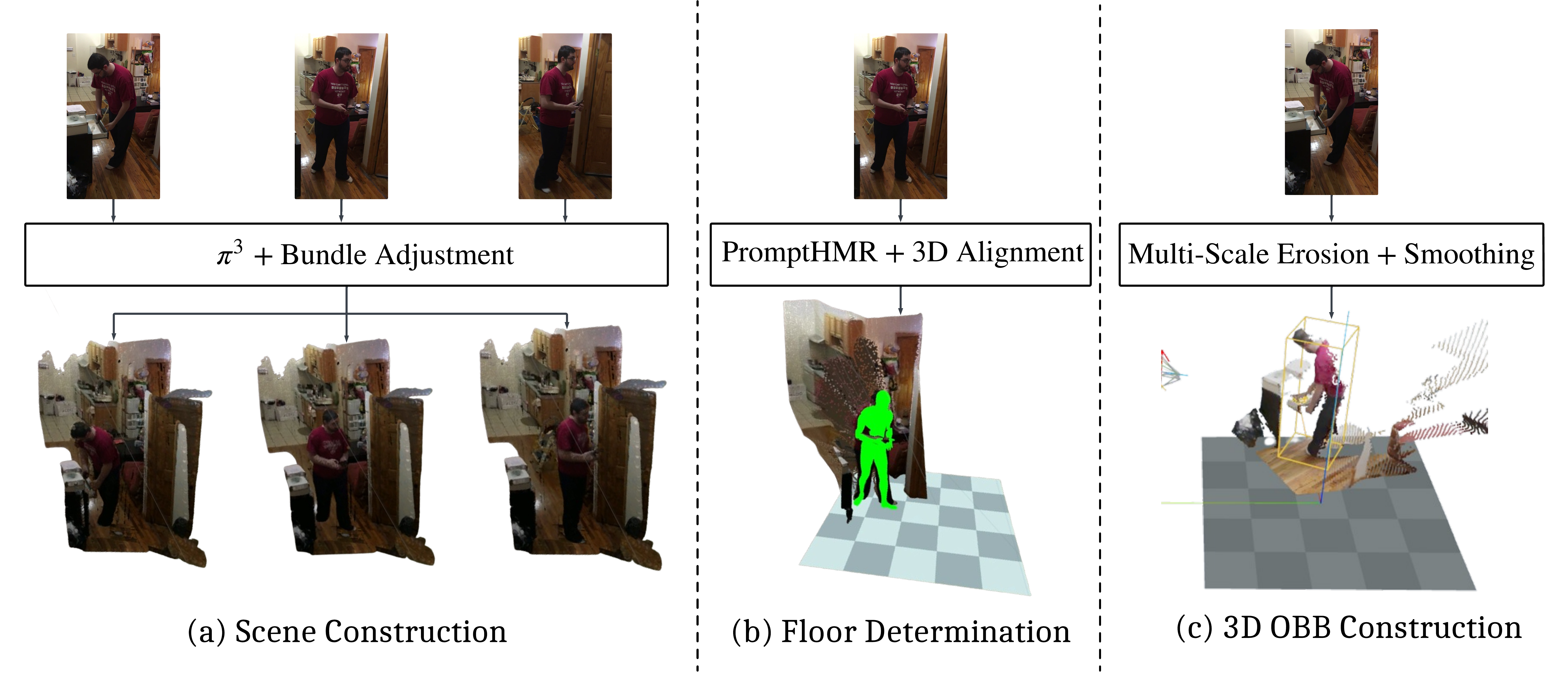}
\vspace{-8mm}
\caption{\textbf{WorldSGG Dataset \& 3D OBB construction pipeline.} Our offline framework generates persistent 3D annotations through three sequential steps: 
(a) Scene Construction recovers the global scene point cloud and camera poses using \PiThree{} and 
bundle adjustment; (b) Floor Determination establishes a canonical ground plane by aligning 3D 
human meshes extracted via PromptHMR; and (c) 3D OBB Construction refines raw object geometries 
to fit robust 3D bounding boxes in the world frame.}
\label{fig:world_sgg_dataset}
\vspace{-8mm}
\end{figure}

\noindent \textbf{3D Scene Construction.} We construct per-timestamp 3D scene 
representations from egocentric Action Genome~\cite{ji2019actiongenomeactionscomposition} 
videos using a feed-forward neural
reconstruction model followed by post-hoc geometric alignment.
The pipeline (see Figure~\ref{fig:world_sgg_dataset}) comprises three stages: (i)~frame sampling and preprocessing,
(ii)~feed-forward 3D inference via the \PiThree{}~\cite{wang2025pi} model,
and (iii)~iterative bundle adjustment (BA) for camera pose refinement.
The output provides the shared world coordinate frame in which every object's
3D oriented bounding box $\mathbf{b}_{k}^{t}$ and camera pose
$\mathbf{T}^{t}$ are expressed. For frame sampling, we use an optical
flow guided adaptive strategy: dense flow is computed between consecutive
frames, and those whose aggregated flow magnitude exceeds a threshold are
selected as keyframes, capturing significant motion while discarding
near-static frames.

The \PiThree{} model is a feed-forward neural network
for visual geometry reconstruction. A single forward pass jointly estimates per-pixel local 3D
coordinates, per-pixel confidence scores, and camera-to-world SE(3)
poses $\{\mathbf{T}^{t}\}_{t=1}^{T}$ for all $T$ input views without
requiring a fixed reference view. Although \PiThree{} produces globally consistent reconstructions, the
predicted camera poses exhibit residual drift across long sequences.
We therefore apply an iterative bundle adjustment (BA) step that
jointly refines the camera-to-world poses
$\{\mathbf{T}^{t}\}_{t=1}^{T}$ and the 3D point positions by
minimising the reprojection error over all views (temporal frames).



\vspace{0.5mm}
\noindent \textbf{Geometric Annotation.} We produce world-frame \textbf{oriented 3D bounding boxes}
$\mathbf{b}_{k}^{t}\!\in\!\mathbb{R}^{8\times 3}$ for every annotated
object $w_{k}^{t}\!\in\!\mathcal{W}^{t}$ through a five-stage pipeline: detection, segmentation, static/dynamic
classification, floor-aligned 3D point extraction, and oriented bounding
box fitting with temporal smoothing. GDINO~\cite{liu2024groundingdinomarryingdino} serves as a
zero-shot detector whose outputs are fused with ground-truth 2D
annotations for complete coverage. An LLM classifies each object as
static or dynamic. SAM2~\cite{ravi2024sam2segmentimages} then produces
per-object masks in both \emph{image} and \emph{video} modes; the
pixel-wise union of both ensures per-frame precision. 

For each object, 3D points are extracted from the \PiThree{}
reconstruction using the segmentation mask. The point cloud is
floor-aligned via RANSAC similarity estimation against estimated
SMPL~\cite{wang2025prompthmrpromptablehumanmesh} meshes, providing metric scale and floor-plane
orientation. A multiscale erosion strategy (kernels
$\{0,3,5,7,10\}$\,px) strips boundary noise, selecting the
minimum-volume candidate. We construct an unconstrained PCA-based OBB and apply Kalman filtering 
with RTS backward smoothing to yield temporally consistent trajectories. 
All boxes are converted to the shared
world coordinate frame in which $\mathbf{b}_{k}^{t}$ and
$\mathbf{T}^{t}$ are expressed. 



\vspace{0.5mm}
\noindent \textbf{Semantic Annotation.} ActionGenome4D extends the original Action Genome annotations with dense
relationship labels for \emph{all} objects in the world state
$\mathcal{W}^{t}$ on every annotated frame $I^{t}$. Following Action
Genome~\cite{ji2019actiongenomeactionscomposition}, the predicate set $\mathcal{P}$ spans three
disjoint axes: \textbf{attention} $\mathcal{P}_{\text{att}}$ (3 labels,
single-label), \textbf{spatial} $\mathcal{P}_{\text{spa}}$ (6 labels,
multi-label), and \textbf{contacting} $\mathcal{P}_{\text{con}}$ (17
labels, multi-label). Whereas the original AG annotates relationships only for observed objects
$w_{k}^{t}\!\in\!\mathcal{O}^{t}$, ActionGenome4D additionally covers
\emph{unobserved objects} $w_{k}^{t}\!\in\!\mathcal{U}^{t}$. This transforms the per-frame scene graph
$\mathcal{G}^{t}$ into a temporally complete \emph{world scene graph}
$\mathcal{G}_{\mathcal{W}}^{t}$ where every (person, object) pair
carries relationship labels at every timestamp.

To bootstrap annotations, a VLM first builds a video-level event graph with per-clip
subtitles, then predicts attention, spatial, and contacting labels for every (person,
object) pair on each annotated frame (details in Section~\ref{sec:mllm_wsgg}).
These pseudo-annotations are refined via a purpose-built correction interface;
the workflow and quality assurance are detailed in the supplementary material.


%% file: paper_tex_files/experiments_wsgg.tex
\begin{figure}[!t]
\centering
\includegraphics[width=0.95\textwidth]{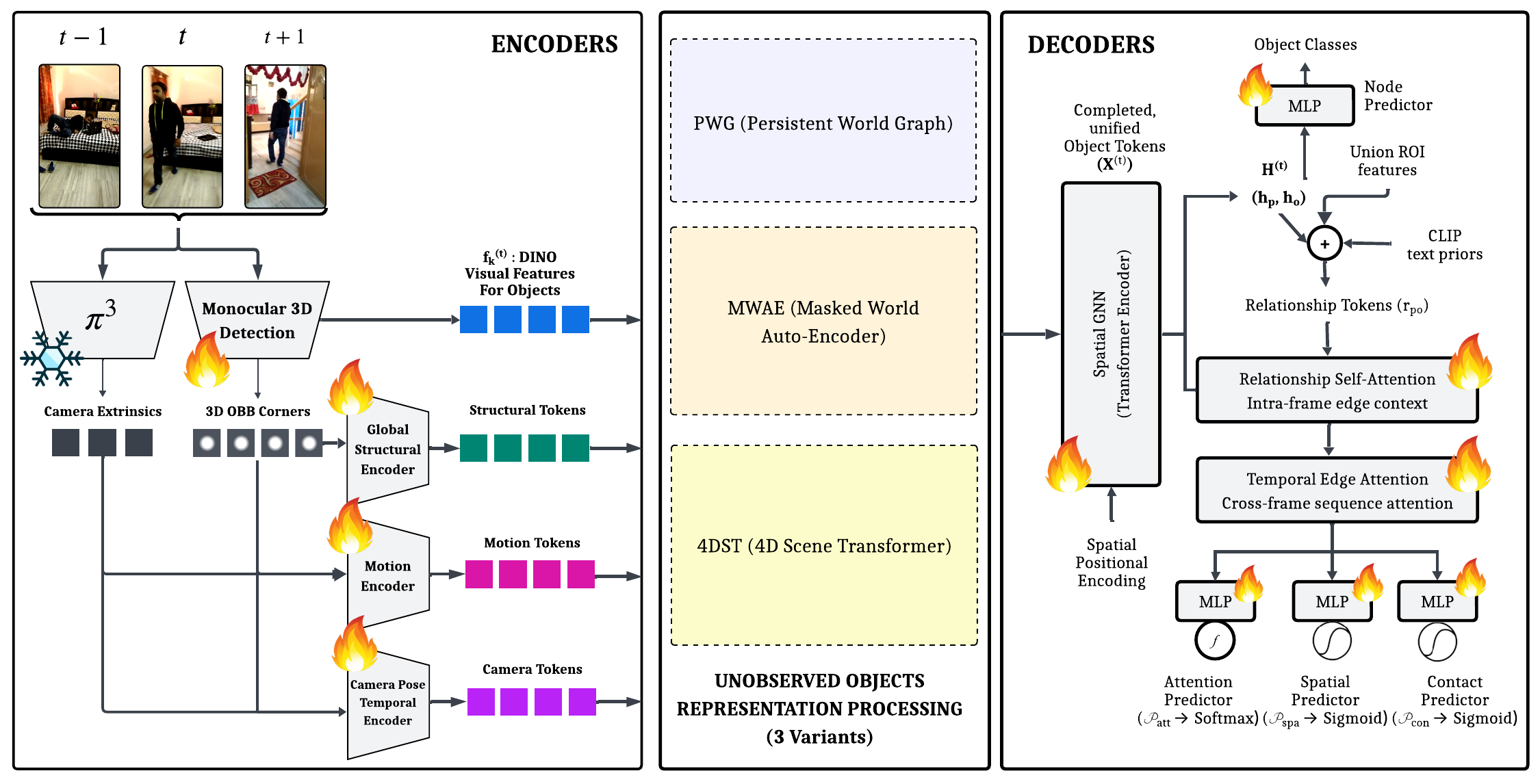}
\vspace{-4mm}
\caption{\textbf{WorldSGG Methods.} Multi-modal inputs including DINO visual features,
monocular 3D geometries, and camera extrinsics are processed into unified structural, motion,
and ego-pose tokens. (Right \& Bottom) Unobserved Object Processing: To maintain strict object
permanence for entities outside the current camera frustum, we propose and evaluate three structural
variants: (1) PWG (Persistent World Graph) utilizes a Last-Known-State (LKS) buffer to explicitly
track and propagate historical features based on temporal staleness ($\Delta_n$).
 (2) MWAE (Masked World Auto-Encoder) frames natural occlusion as a masked modeling task;
 it masks unobserved visual streams and uses an Associative Retriever with asymmetric cross-attention
 (querying all tokens against only visible ones) to reconstruct missing features of unobserved objects.
 (3) 4DST (4D Scene Transformer) fuses multi-modal tokens at a Fusion Node and applies unmasked
 bidirectional temporal self-attention followed by Spatial GNNs to output dense, globally-aware
 spatiotemporal representations ($\mathbf{H}^{(t)}$).}
\label{fig:world_sgg_methods}
\vspace{-6mm}
\end{figure}


We present three methods for \WSGG{}, the task of constructing
the world scene graph $\mathcal{G}_{\mathcal{W}}^{t}$ at each timestamp $t$ predicting relationships
over \emph{all} objects in the world state $\mathcal{W}^{t}$, including those that are temporarily
\emph{occluded} ($w_{k}^{t} \in \mathcal{U}^{t}$), have \emph{left the camera field of view}, or
have \emph{never been directly observed} in frame $I^{t}$. Unlike VidSGG methods, which construct only the graph $\mathcal{G}^{t}$ over observed
objects $\mathcal{O}^{t}$~\cite{cong2021spatialtemporaltransformerdynamicscene,chen2019knowledgeembeddedroutingnetworkscene}, \WSGG{} maintains a persistent
world state and predicts relationships for all objects in the world using 3D geometric
scaffolds, temporal memory mechanisms, and associative retrieval. We explore
complementary inductive biases for constructing dense representations of unobserved objects (see Figure~\ref{fig:world_sgg_methods}):

\vspace{-4mm}
\begin{enumerate}
    \item \textbf{\PWG{}} (Persistent World Graph, Sec.~\ref{sec:pwg}): Maintains a
          \emph{persistent object memory buffer} that freezes each object's visual features at its
          last visible appearance, drawing on principles of object
          permanence~\cite{piloto2022intuitive}.

    \item \textbf{\MWAE{}} (Masked World Auto-Encoder, Sec.~\ref{sec:mwae}): Applies a
          \emph{masked auto-encoder} framework~\cite{he2022masked,tong2022videomae} to relational
          scene understanding where occlusions and camera motion provide natural masking, and a 3D
          geometric scaffold provides the complete structural prior.

    \item \textbf{\FDST{}} (4D Scene Transformer, Sec.~\ref{sec:4dst}): Replaces the
          non-differentiable buffer with a \emph{temporal object transformer},
          extending the factorized spatial-temporal attention
          paradigm~\cite{arnab2021vivit,bertasius2021space} from visible-only 2D
          reasoning to full 4D.
\end{enumerate}

We note that all the above methods operate on the same input: pre-extracted per-object visual features $\mathbf{f}_{k}^{t}
\in \mathbb{R}^{d_{\text{roi}}}$ from frozen DINOv2/v3~\cite{oquab2024dinov2learningrobustvisual,simeoni2025dinov3} backbones, 3D oriented
bounding box corners $\mathbf{b}_{k}^{t} \in \mathbb{R}^{8 \times 3}$ from the persistent global
wireframe, and camera extrinsic matrices $\mathbf{T}^{t}$ estimated using \PiThree{}. Relationship predictions are made for the
three predicate categories: \textit{attention} $\mathcal{P}_{\text{att}}$
($|\mathcal{P}_{\text{att}}| = 3$), \textit{spatial} $\mathcal{P}_{\text{spa}}$
($|\mathcal{P}_{\text{spa}}| = 6$), and \textit{contacting} $\mathcal{P}_{\text{con}}$
($|\mathcal{P}_{\text{con}}| = 17$).

\vspace{-4mm}
\subsection{Shared Architectural Components}
\vspace{-1mm}
\label{sec:shared}
\vspace{0.5mm}
\noindent\textbf{Global Structural Encoder.}\label{sec:gse}
Each OBB $\mathbf{b}_{k}^{t}$ (8 corners $\mat{C}_k\!\in\!\mR^{8\times3}$) is centered, flattened,
and concatenated with its centroid to form a $27$-d input
$\mat{x}_n = [\op{flatten}(\tilde{\mat{c}}_{n,1},\dots,\tilde{\mat{c}}_{n,8})\;\|\;
\bar{\mat{c}}_n] \in \mR^{27}$.
A shared MLP produces per-object structural tokens $\mat{g}_n\!\in\!\mR^{d_{\text{struct}}}$, and a
permutation-invariant global token
$\mat{g}_{\text{global}} = \op{MLP}_{\text{global}}(\max_{n\in\mathcal{V}} \mat{g}_n)$ is obtained
via max-pooling over valid objects.

\vspace{0.5mm}
\noindent\textbf{Spatial Positional Encoding.}\label{sec:spe}
To ground the model in relative 3D geometry, for each object pair $(i,j)$ 
we compute a 5-d pairwise feature: Euclidean distance
$d_{ij}$, unit direction $\hat{\mat{d}}_{ij}\!\in\!\mR^{3}$, and log-volume ratio
$\rho_{ij}=\log V_i - \log V_j$. A shared MLP encodes each tuple and the results are
mean-aggregated over valid neighbors:
$\mat{s}_i = \op{Linear}\bigl(\frac{1}{|\mathcal{N}_i|}
\sum_{j}\op{MLP}_{\text{pair}}([d_{ij},\hat{\mat{d}}_{ij},\rho_{ij}])\bigr)
\in \mR^{d_{\text{model}}}$.

\vspace{0.5mm}
\noindent\textbf{Spatial GNN.}\label{sec:spatial_gnn}
Intra-frame object interactions are modeled by a Transformer
encoder~\cite{vaswani2023attentionneed} with additive spatial positional encoding ($\mat{S}^{(t)}$):
\begin{equation}
    \mat{H}^{(t)} = \op{TransformerEncoder}\bigl(\mat{X}^{(t)} + \mat{S}^{(t)},\;
    \text{mask}=\overline{\mat{v}}^{(t)}\bigr) \in \mR^{N \times d_{\text{model}}},
    \label{eq:spatial_gnn}
\end{equation}
where $\mat{X}^{(t)}$ are input tokens and $\overline{\mat{v}}^{(t)}$ the validity mask.

\vspace{0.5mm}
\noindent\textbf{Node Predictor.}\label{sec:node_predictor}
Object class prediction is performed by a two-layer MLP:
$\hat{\mat{y}}_n^{\text{obj}} = \op{MLP}_{\text{node}}(\mat{h}_n) \in \mR^{C_{\text{obj}}}$
($C_{\text{obj}} = 37$). This is only used in SGDet mode.

\vspace{0.5mm}
\noindent\textbf{Relationship Predictor.}\label{sec:rel_pred}
For each human-object pair $(p, o)$\footnote{Since ActionGenome4D centers on human-object interactions, 
we predict relationships for human--object pairs; the architecture generalizes to 
arbitrary object pairs.}, a relationship token is formed:
$\mat{r}_{po} = \op{Proj}([\mat{h}_p \;\|\; \mat{h}_o \;\|\; \mat{u}_{po} \;\|\; \mat{e}_p^{\text{txt}}
\;\|\; \mat{e}_o^{\text{txt}}]) \in \mR^{d_{\text{rel}}}$,
where $\mat{u}_{po}$ are projected union ROI features and $\mat{e}_p^{\text{txt}}, \mat{e}_o^{\text{txt}}$ are
frozen CLIP~\cite{radford2021learning} text embeddings indexed by predicted class.
All $K$ tokens within a frame then self-attend
($\mat{R}^{(t)} = \op{RelTransformer}(\{\mat{r}_{po}\})$),
and three MLP heads predict attention ($|\mathcal{P}_{\text{att}}|$ classes),
spatial ($|\mathcal{P}_{\text{spa}}|$ labels), and contacting
($|\mathcal{P}_{\text{con}}|$ labels) relationships.

\vspace{0.5mm}
\noindent\textbf{Temporal Edge Attention.}\label{sec:tea}
For each unique pair $(p,o)$ across frames $\mathcal{T}_{po}$, relationship tokens are augmented
with learned temporal embeddings and refined via a Transformer:
$\tilde{\mat{r}}_{po}^{(t)} = \op{TemporalEncoder}(\{{\mat{r}}_{po}^{(\tau)} +
\mat{e}_{\tau}\}_{\tau \in \mathcal{T}_{po}})$.

\vspace{0.5mm}
\noindent\textbf{Camera Pose Encoder.}\label{sec:cam_encoder}
The camera extrinsic $\mathbf{T}^{t}=[\mat{R}^{t}\,|\,\boldsymbol{\tau}^{t}]$ is encoded into a global token
$\mat{c}_{\text{cam}} = \op{MLP}_{\text{cam}}([\op{6D}(\mat{R}) \;\|\; \boldsymbol{\tau}])
\in \mR^{d_{\text{camera}}}$
using the 6D rotation representation~\cite{zhou2019continuity}. Per-object camera-relative
features (distance, view alignment, azimuth) are also computed and projected to
$\mR^{d_{\text{camera}}}$.

\vspace{0.5mm}
\noindent\textbf{Camera Temporal Encoder.}\label{sec:cam_temporal}
Relative poses between consecutive frames ($\mat{R}_{\text{rel}} = \mat{R}_t
\mat{R}_{t-1}^{\!\top}$, $\boldsymbol{\tau}_{\text{rel}} = \boldsymbol{\tau}_t -
\mat{R}_{\text{rel}}\boldsymbol{\tau}_{t-1}$) are MLP-encoded, augmented with temporal positional
embeddings, and processed by self-attention to capture ego-motion.

\vspace{0.5mm}
\noindent\textbf{Motion Feature Encoder.}\label{sec:motion_encoder}
Per-object 3D velocity, acceleration, and camera-relative velocity form an 11-d feature
processed by a shared MLP.


\vspace{-4mm}
\subsection{\PWG{} - Persistent World Graph}
\label{sec:pwg}

\PWG{} is the foundational baseline for \WSGG{}: it freezes each object's visual features at its
last visible appearance and reasons over the full world graph using 3D geometry as scaffolding.
This implements the cognitive science principle of \textbf{object
permanence}~\cite{piloto2022intuitive,Spelke1990ObjectPerception} at the feature level.

\vspace{0.5mm}
\noindent\textbf{LKS Memory Buffer.}\label{sec:lks_buffer}
A non-differentiable zero-order hold retrieves raw DINO features from the nearest visible frame:
\begin{equation}
    \mat{m}_n^{(t)} = \begin{cases}
        \mat{f}_n^{(t)} & \text{if visible at $t$}, \\
        \mat{f}_n^{(\tau^*)} & \tau^* = \arg\min_{\tau:\text{vis}(n,\tau)} |t - \tau|, \\
        \bm{0} & \text{if never seen}.
    \end{cases}
    \label{eq:lks_buffer}
\end{equation}
A per-object staleness $\Delta_n^{(t)} = |t - \tau^*|$ is also recorded.

\vspace{0.5mm}
\noindent\textbf{Tokenizer \& Pipeline.}
Geometry, buffered features, camera, and log-staleness are fused:
$\mat{x}_n^{(t)} = \op{Proj}([\mat{g}_n^{(t)} \;\|\; \mat{m}_n^{(t)} \;\|\; \mat{c}_n^{(t)}
\;\|\; \log(\Delta_n^{(t)} + 1)]) \in \mR^{d_{\text{model}}}$.
All $T$ frames corresponding to the video are then processed through the Spatial GNN, Node
Predictor, Relationship Predictor, and Temporal Edge Attention.

\vspace{0.5mm}
\noindent\textbf{Loss Function.}\label{sec:pwg_loss}
Pairs are split into \emph{visible} (both in $\mathcal{O}^{t}$, clean GT) and
\emph{unobserved} (at least one in $\mathcal{U}^{t}$, VLM pseudo-labels weighted by
$\lambda_{\text{vlm}}$). For each predicate axis $r \in \{\text{att, spa, con}\}$, BCE
losses are computed:
\begin{equation}
    \mathcal{L}_{\text{\PWG{}}} = \sum_{r}
    \bigl(\mathcal{L}_{r}^{\text{vis}} + \lambda_{\text{vlm}} \cdot \mathcal{L}_{r}^{\text{unobs}}\bigr) +
    \mathcal{L}_{\text{node}}.
    \label{eq:pwg_total}
\end{equation}

\input{paper_tex_files/results_tex_files/recall_comparison.tex}

\vspace{-8mm}
\subsection{\FDST{} - 4D Scene Transformer}
\label{sec:4dst}

\FDST{} addresses a fundamental limitation of \PWG{}: the memory buffer is non-differentiable.
\FDST{} replaces it with a \emph{Temporal Object Transformer} that performs per-object
bidirectional self-attention over the entire video, enabling end-to-end learning of rich contextual temporal
features.

\vspace{0.5mm}
\noindent\textbf{Temporal Object Transformer.}\label{sec:temporal_transformer}
For each object $n$, a sequence of $T$ tokens is formed by fusing visual, structural, camera,
motion, and ego-motion features:
$\mat{z}_n^{(t)} = \op{InputProj}([\op{VisProj}(\mat{f}_n^{(t)}) \;\|\; \mat{g}_n^{(t)}
\;\|\; \mat{c}_n^{(t)} \;\|\; \mat{\mu}_n^{(t)} \;\|\; \mat{e}_t])
\in \mR^{d_{\text{memory}}}$.
Sinusoidal positional encoding and a learned visibility embedding
$\mat{e}_{\text{vis}}[\mathbb{1}[\text{vis}(n,t)]]$ are added. Unobserved tokens still receive
full geometric and camera features. Each object's sequence is then processed by a bidirectional
Transformer encoder (no causal masking):
$\{\tilde{\mat{z}}_n^{(t)}\}_{t=1}^{T} =
\op{TransformerEncoder}(\{\mat{z}_n^{(t)}\}_{t=1}^{T})$.

\vspace{0.5mm}
\noindent\textbf{Pipeline \& Loss.}
Beyond the shared components, \FDST{} adds the Temporal Object Transformer. The loss mirrors \PWG{}:

\begin{equation}
    \mathcal{L}_{\text{\FDST{}}} = \sum_{r}
    (\mathcal{L}_{r}^{\text{vis}} + \lambda_{\text{vlm}} \cdot
    \mathcal{L}_{r}^{\text{unobs}}) + \mathcal{L}_{\text{node}}
    \label{eq:4dst_total}
\end{equation}

\input{paper_tex_files/results_tex_files/mean_recall_comparison.tex}

\vspace{-8mm}
\subsection{\MWAE{} - Masked World Auto-Encoder}
\label{sec:mwae}

\MWAE{} reframes \WSGG{} as a \emph{structured completion problem}: given partial observations
(visible objects) and a complete geometric scaffold (persistent 3D wireframe), the model ``fills
in'' visual representations and relationship predictions for occluded objects. The design
transposes the MAE framework~\cite{he2022masked} from the patch domain to the object/relationship
domain, where at inference occlusions and camera motion provide natural masking instead of synthetic
augmentation.

\vspace{0.5mm}
\noindent\textbf{Scaffold Tokenizer.}\label{sec:scaffold_tokenizer}
Every object receives a token at every frame:
$\mat{x}_n^{(t)} = \op{FusionMLP}([\mat{g}_n^{(t)} \;\|\; \mat{v}_n^{(t)} \;\|\;
\mat{c}_n^{(t)} \;\|\; \mat{\mu}_n^{(t)} \;\|\; \mat{e}_t])$,
where $\mat{v}_n^{(t)} = \op{VisProj}(\mat{f}_n^{(t)})$ for visible objects and a learnable
$\mat{e}_{\text{[MASK]}}$ otherwise. During training, a fraction $p_{\text{mask}}$ of visible
objects are artificially masked.

\vspace{0.5mm}
\noindent\textbf{Associative Retriever.}\label{sec:associative_retriever}
Per-object asymmetric cross-attention fills in masked tokens: queries span all tokens (visible +
masked) while keys/values are restricted to visible tokens only, preventing unobserved tokens from
attending to each other. A visibility embedding is added post-retrieval:
$\hat{\mat{x}}_n^{(t)} = \mat{x}_n^{(t)} +
\mat{e}_{\text{vis}}[\mathbb{1}[\text{vis}(n,t) \wedge \neg\text{masked}(n,t)]]$.

\vspace{0.5mm}
\noindent\textbf{Loss \& Pipeline.}\label{sec:mwae_loss}
\MWAE{} replaces the LKS buffer with the Scaffold Tokenizer and Associative Retriever, adding a
reconstruction head
($\hat{\mat{f}}_n^{(t)} = \op{Linear}(\hat{\mat{x}}_n^{(t)})$, applied only to
artificially masked objects). The triple-objective loss is:
\begin{equation}
    \mathcal{L}_{\text{\MWAE{}}} = \mathcal{L}_{\text{SG}} + \lambda_{\text{recon}} \cdot
    \lambda_{\text{dom}} \cdot \mathcal{L}_{\text{recon}} + \mathcal{L}_{\text{sim}},
    \label{eq:mwae_total}
\end{equation}
where $\mathcal{L}_{\text{SG}}$ follows the same visible/unobserved split,
$\mathcal{L}_{\text{recon}}$ is an MSE loss over artificially masked features, and
$\mathcal{L}_{\text{sim}}$ re-predicts relationships for masked-but-visible objects against clean
GT.

\vspace{-4mm}
\subsection{Results}
\label{sec:results}

\noindent\textbf{Evaluation Protocols.}
We evaluate under \textbf{PredCls} (GT labels and 3D localizations given) and \textbf{SGDet}
(end-to-end detection and prediction), each with \emph{With Constraint} (top-1) and
\emph{No Constraint} (multi-label) settings.

\textbf{Analysis of Results.} Tables~\ref{tab:recall_comparison} and~\ref{tab:mean_recall_comparison} report R@K and mR@K for
\PWG{}, \MWAE{}, and \FDST{} with DINOv2-L and DINOv3-L backbones.
\textbf{(i)}~\FDST{} delivers the most consistent gains, leading SGDet With Constraint
(R@10\,=\,42.64 DINOv2-L; R@20\,=\,58.03, R@50\,=\,71.95 DINOv3-L) and SGDet No Constraint
mR (25.70/38.02 DINOv2-L). Its differentiable temporal transformer enables end-to-end
propagation, boosting both object classification and tail-class recall.
\textbf{(ii)}~\MWAE{} excels in multi-label settings, leading PredCls No Constraint
(R@10\,=\,81.50, mR@10\,=\,55.09 DINOv3-L); its reconstruction and simulated-occlusion losses
act as complementary regularizers for diverse predicates.
\textbf{(iii)}~\PWG{} trails the best method by only 1--2 points in most PredCls settings despite
its non-differentiable design, confirming that the persistent 3D scaffold alone provides a strong
structural prior and validates the \WSGG{} formulation.



%% file: paper_tex_files/results_tex_files/recall_comparison.tex
\begin{table*}[t]
\centering
\caption{\textbf{Recall (R@K)} comparison across methods and feature backbones for PredCls and
SGDet tasks on ActionGenome4D. Best results per column are in \textbf{bold}.}
\label{tab:recall_comparison}
\setlength{\tabcolsep}{5pt}
\resizebox{\textwidth}{!}{%
\begin{tabular}{@{}ll ccc ccc ccc ccc@{}}
\toprule
 & & \multicolumn{6}{c}{\textbf{PredCls}} & \multicolumn{6}{c}{\textbf{SGDet}} \\
\cmidrule(lr){3-8} \cmidrule(l){9-14}
 & & \multicolumn{3}{c}{With Constraint} & \multicolumn{3}{c}{No Constraint}
 & \multicolumn{3}{c}{With Constraint} & \multicolumn{3}{c}{No Constraint} \\
\cmidrule(lr){3-5} \cmidrule(lr){6-8} \cmidrule(lr){9-11} \cmidrule(l){12-14}
\textbf{Method} & \textbf{Backbone}
 & R@10 & R@20 & R@50
 & R@10 & R@20 & R@50
 & R@10 & R@20 & R@50
 & R@10 & R@20 & R@50 \\
\midrule
\rowcolor{gray!6}
PWG & DINOv2-L & 65.07 & 67.99 & 68.00 & 80.06 & 94.39 & 99.59 & 41.69 & 57.14 & 69.63 & 28.62 & 40.69 & 56.66 \\
\rowcolor{gray!6}
    & DINOv3-L & 65.58 & 68.57 & 68.58 & 81.01 & 94.73 & 99.76 & 39.96 & 57.11 & 70.93 & 26.84 & 37.93 & 52.57 \\
\addlinespace[2pt]
MWAE & DINOv2-L & 65.33 & 68.30 & 68.31 & 80.96 & 94.51 & 99.65 & 41.69 & 57.06 & 69.50 & 28.91 & 40.85 & 56.96 \\
     & DINOv3-L & 65.57 & 68.58 & 68.59 & \textbf{81.50} & \textbf{95.11} & \textbf{99.78} & 39.67 & 57.06 & 70.90 & 26.89 & 38.17 & 52.86 \\
\addlinespace[2pt]
\rowcolor{gray!6}
4DST & DINOv2-L & 64.31 & 67.26 & 67.26 & 80.07 & 94.27 & 99.67 & \textbf{42.64} & 57.86 & 70.32 & \textbf{29.33} & \textbf{41.13} & \textbf{57.00} \\
\rowcolor{gray!6}
     & DINOv3-L & \textbf{66.11} & \textbf{69.11} & \textbf{69.12} & 81.04 & 94.70 & 99.72 & 40.84 & \textbf{58.03} & \textbf{71.95} & 27.04 & 38.35 & 52.90 \\
\bottomrule
\end{tabular}}
\vspace{-2mm}
\end{table*}

%% file: paper_tex_files/results_tex_files/mean_recall_comparison.tex
\begin{table*}[t]
\centering
\caption{\textbf{Mean Recall (mR@K)} comparison across methods and feature backbones for PredCls
and SGDet tasks on ActionGenome4D. Best results per column in \textbf{bold}.}
\label{tab:mean_recall_comparison}
\setlength{\tabcolsep}{5pt}
\resizebox{\textwidth}{!}{%
\begin{tabular}{@{}ll ccc ccc ccc ccc@{}}
\toprule
 & & \multicolumn{6}{c}{\textbf{PredCls}} & \multicolumn{6}{c}{\textbf{SGDet}} \\
\cmidrule(lr){3-8} \cmidrule(l){9-14}
 & & \multicolumn{3}{c}{With Constraint} & \multicolumn{3}{c}{No Constraint}
 & \multicolumn{3}{c}{With Constraint} & \multicolumn{3}{c}{No Constraint} \\
\cmidrule(lr){3-5} \cmidrule(lr){6-8} \cmidrule(lr){9-11} \cmidrule(l){12-14}
\textbf{Method} & \textbf{Backbone}
 & mR@10 & mR@20 & mR@50
 & mR@10 & mR@20 & mR@50
 & mR@10 & mR@20 & mR@50
 & mR@10 & mR@20 & mR@50 \\
\midrule
\rowcolor{gray!6}
PWG & DINOv2-L & 34.87 & 38.79 & 38.80 & 52.83 & 72.56 & 93.54 & 11.09 & 22.46 & 36.86 & 24.58 & 36.41 & 54.98 \\
\rowcolor{gray!6}
    & DINOv3-L & 36.90 & 41.28 & 41.29 & 54.27 & 74.73 & 94.27 & 9.56 & 22.61 & 39.61 & 23.25 & 35.60 & 55.51 \\
\addlinespace[2pt]
MWAE & DINOv2-L & 35.03 & 39.46 & 39.47 & 53.31 & 73.90 & 95.04 & 10.95 & 22.19 & 36.33 & 22.98 & 37.60 & 55.47 \\
     & DINOv3-L & 36.41 & 40.98 & 40.99 & \textbf{55.09} & \textbf{75.15} & 95.02 & 9.75 & 22.97 & 39.74 & 23.16 & 35.74 & 54.08 \\
\addlinespace[2pt]
\rowcolor{gray!6}
4DST & DINOv2-L & 35.39 & 39.50 & 39.50 & 51.99 & 72.18 & 94.05 & \textbf{11.31} & \textbf{24.39} & 38.23 & \textbf{25.70} & \textbf{38.02} & 55.40 \\
\rowcolor{gray!6}
     & DINOv3-L & \textbf{37.16} & \textbf{41.55} & \textbf{41.56} & 54.15 & 74.73 & \textbf{95.81} & 9.72 & 23.53 & \textbf{40.70} & 23.99 & 36.65 & \textbf{55.73} \\
\bottomrule
\end{tabular}}
\vspace{-2mm}
\end{table*}

%% file: paper_tex_files/experiments_mllm.tex
\begin{figure}[!t]
\centering
\includegraphics[width=\textwidth]{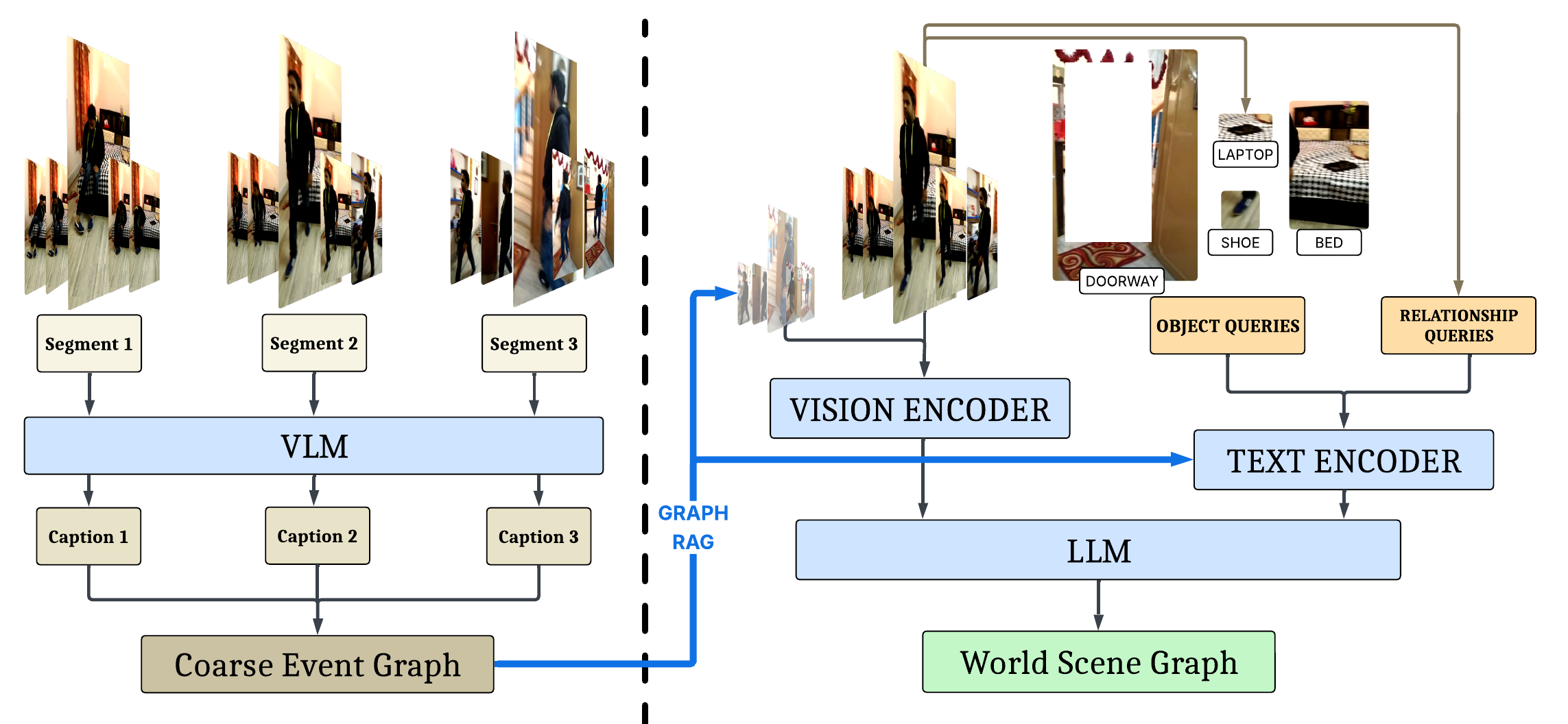}
\vspace{-4mm}
\caption{\textbf{The Graph RAG inference pipeline.} (Left) Sequential video segments are processed 
by a VLM into local captions to construct a global Coarse Event Graph, establishing a high-level 
temporal narrative. (Right) To generate the fine-grained world scene graph, our Graph RAG module
retrieves overarching spatiotemporal context from the coarse graph. This global prior is injected 
into both the visual stream (Vision Encoder) and the textual stream, where it enriches targeted 
Object and Relationship queries. An LLM then
deduces the final, offline World Scene Graph.}
\label{fig:world_sgg_mllm_pipeline}
\vspace{-4mm}
\end{figure}

We evaluate how well current VLMs perform on \WSGG{} (Figure~\ref{fig:world_sgg_mllm_pipeline}). We query
\emph{all} objects in $\mathcal{W}^{t}$ on every annotated frame, producing a complete
relationship prediction for every (person, object) pair. These VLM predictions also serve as
the pseudo-annotations that seed the manual correction process.

\vspace{0.5mm}
\noindent\textbf{Task \& Evaluation Modes.}
Given a video $V_{1}^{T}$ and world state $\mathcal{W}^{t}$, the goal is to predict
attention (single-label), spatial, and contacting (both multi-label) predicates for every
object $w_{k}^{t}\!\in\!\mathcal{W}^{t}$.
We evaluate under \emph{PredCls} (ground-truth object labels) and \emph{SGDet}
(VLM-estimated objects with vocabulary filtering and discriminative verification).

\vspace{0.5mm}
\noindent\textbf{Evaluated Models.}
Three VLMs are evaluated via vLLM with tensor parallelism:
\textbf{Kimi-VL}~\cite{kimiteam2025kimivltechnicalreport}\footnote{Inconsistent structured-output compliance; results in supplementary.},
\textbf{InternVL~2.5}~\cite{wang2025internvl35advancingopensourcemultimodal} (8B), and
\textbf{Qwen~2.5-VL}~\cite{bai2025qwen3vltechnicalreport} (7B).

\vspace{0.5mm}
\noindent\textbf{Method 1: Graph RAG.}
A 4-step pipeline over a precomputed coarse event video graph:
(1)~keyword extraction (one batched VLM call);
(2)~BGE-embedding retrieval~\cite{chen2025m3embeddingmultilingualitymultifunctionalitymultigranularity} (cosine $>$0.5, top-20; no VLM call);
(3)~binary node refinement (one batched VLM call);
(4)~frame-level answer generation with graph-node context (one batched VLM call).
Temporally sorted subtitles are prepended to both generative and verification prompts.

\vspace{0.5mm}
\noindent\textbf{Method 2: Subtitle-Only.}
Bypasses graph construction: temporally sorted subtitles are concatenated with the
relationship prompt and processed in one batched VLM call (sub-batches of 64),
followed by one bulk verification call.

\vspace{0.5mm}
\noindent\textbf{Verification \& Metrics.}
Both methods share a discriminative verification step: each predicted label is verified
via a binary Yes/No query ($p_\text{yes}$) in one bulk call.
We report precision, recall, and F1 per predicate axis, plus micro- and macro-averaged
F1. Full prompts and the algorithm are in the supplementary material.


\input{paper_tex_files/results_tex_files/combined_gt_plus_corrections_micro.tex}
\input{paper_tex_files/results_tex_files/combined_gt_plus_corrections_macro.tex}

\vspace{0.5mm}
\noindent\textbf{Results Analysis.}
Tables~\ref{tab:combined_gt_plus_corrections_micro}
and~\ref{tab:combined_gt_plus_corrections_macro} present micro- and macro-averaged results.
\textbf{(i)}~Qwen~2.5-VL is the strongest backbone: under RAG + PredCls it reaches
53.3\% micro F1 vs.\ 40.8\% for InternVL~2.5 ($+$12.5); the gap persists under SGDet
(39.0 vs.\ 24.8), confirming stronger relationship reasoning rather than detection alone.
\textbf{(ii)}~RAG consistently outperforms Subtitle-Only, though the margin
narrows for stronger VLMs (InternVL $+$3.8 vs.\ Qwen $+$2.1 PredCls micro F1),
suggesting that larger models already internalise temporal context.
\textbf{(iii)}~SGDet roughly halves recall relative to PredCls while precision degrades
gently, identifying world-level object detection as the primary bottleneck.
\textbf{(iv)}~Predicate difficulty increases from Attention to Contacting to Spatial:
attention is single-label with three classes, whereas spatial spans a larger, sparser set.

%% file: paper_tex_files/results_tex_files/combined_gt_plus_corrections_micro.tex
\definecolor{lightrow}{gray}{0.93}
\begin{table}[!t]
\centering
\caption{\small\textbf{VLM relationship prediction on corrected ActionGenome4D annotations (micro-averaged).}
  \textbf{P}\,=\,Precision, \textbf{R}\,=\,Recall, \textbf{F1}\,=\,F1-score.
  \textbf{Micro F1}: micro-average across all three predicate axes.
  \textbf{Macro F1}: mean of per-class F1 scores.}
\label{tab:combined_gt_plus_corrections_micro}
\resizebox{\textwidth}{!}{%
\setlength{\tabcolsep}{5pt}
\begin{tabular}{ll l ccc ccc ccc c c}
\toprule
\multirow{2}{*}{\textbf{Pipeline}} &
\multirow{2}{*}{\textbf{Model}} &
\multirow{2}{*}{\textbf{Mode}} &
\multicolumn{3}{c}{\textbf{Attention}} &
\multicolumn{3}{c}{\textbf{Contacting}} &
\multicolumn{3}{c}{\textbf{Spatial}} &
\textbf{Micro} &
\textbf{Macro} \\
 & & & P & R & F1 & P & R & F1 & P & R & F1 & F1 & F1 \\
\midrule
\multirow{4}{*}{Graph RAG}
  & \multirow{2}{*}{InternVL 2.5}
    & PredCls & 53.8 & 53.7 & 53.8 & 45.4 & 40.3 & 42.7 & 30.8 & 24.4 & 27.2 & 40.8 & 23.4 \\
  & & SGDet   & 49.6 & 22.3 & 30.8 & 43.5 & 18.1 & 25.5 & 35.3 & 12.9 & 18.9 & 24.8 & 15.4 \\
\cmidrule(lr){2-13}
  & \multirow{2}{*}{Qwen 2.5-VL}
    & PredCls & 61.4 & 61.3 & 61.4 & 60.7 & 53.6 & 56.9 & 47.6 & 38.5 & 42.5 & 53.3 & 26.6 \\
  & & SGDet   & 59.7 & 34.8 & 44.0 & 62.0 & 32.3 & 42.5 & 48.7 & 23.1 & 31.3 & 39.0 & 20.2 \\
\midrule
\multirow{4}{*}{Subtitle-Only}
  & \multirow{2}{*}{InternVL 2.5}
    & PredCls & 50.6 & 50.6 & 50.6 & 44.3 & 38.5 & 41.2 & 23.4 & 17.7 & 20.2 & 37.0 & 21.7 \\
  & & SGDet   & 43.5 & 19.7 & 27.1 & 40.6 & 17.2 & 24.2 & 26.4 &  9.7 & 14.2 & 21.5 & 14.5 \\
\cmidrule(lr){2-13}
  & \multirow{2}{*}{Qwen 2.5-VL}
    & PredCls & 61.9 & 61.7 & 61.8 & 57.8 & 49.0 & 53.0 & 45.4 & 35.4 & 39.8 & 51.2 & 25.9 \\
  & & SGDet   & 59.6 & 34.3 & 43.5 & 58.6 & 29.5 & 39.3 & 47.0 & 21.8 & 29.7 & 37.2 & 20.0 \\
\bottomrule
\end{tabular}}
\vspace{-4mm}
\end{table}

%% file: paper_tex_files/results_tex_files/combined_gt_plus_corrections_macro.tex
\begin{table}[!t]
\centering
\caption{\small\textbf{VLM relationship prediction on corrected ActionGenome4D annotations (macro-averaged).}
  Same setting as Table~\ref{tab:combined_gt_plus_corrections_micro};
  P, R, and F1 are computed \emph{per class} then averaged equally.
  \textbf{M-F1}: overall macro-averaged F1 across all three axes.}
\label{tab:combined_gt_plus_corrections_macro}
\resizebox{\textwidth}{!}{%
\setlength{\tabcolsep}{5pt}
\begin{tabular}{ll l ccc ccc ccc c}
\toprule
\multirow{2}{*}{\textbf{Pipeline}} &
\multirow{2}{*}{\textbf{Model}} &
\multirow{2}{*}{\textbf{Mode}} &
\multicolumn{3}{c}{\textbf{Attention}} &
\multicolumn{3}{c}{\textbf{Contacting}} &
\multicolumn{3}{c}{\textbf{Spatial}} &
\textbf{Overall} \\
 & & & P & R & F1 & P & R & F1 & P & R & F1 & M-F1 \\
\midrule
\multirow{4}{*}{Graph RAG}
  & \multirow{2}{*}{InternVL 2.5}
    & PredCls & 38.6 & 41.5 & 36.4 & 30.1 & 31.1 & 25.7 & 37.3 & 16.5 & 12.1 & 23.4 \\
  & & SGDet   & 39.7 & 19.3 & 21.9 & 31.4 & 15.4 & 17.5 & 34.4 &  6.7 &  7.5 & 15.4 \\
\cmidrule(lr){2-12}
  & \multirow{2}{*}{Qwen 2.5-VL}
    & PredCls & 38.3 & 39.6 & 38.4 & 31.0 & 27.1 & 26.0 & 40.2 & 23.6 & 23.1 & 26.6 \\
  & & SGDet   & 39.3 & 23.3 & 28.3 & 29.9 & 17.4 & 19.9 & 39.0 & 14.0 & 17.5 & 20.2 \\
\midrule
\multirow{4}{*}{Subtitle-Only}
  & \multirow{2}{*}{InternVL 2.5}
    & PredCls & 37.6 & 40.1 & 34.5 & 31.7 & 35.8 & 24.6 & 34.8 & 14.6 &  8.8 & 21.7 \\
  & & SGDet   & 39.6 & 18.6 & 19.2 & 31.5 & 17.8 & 17.1 & 30.3 &  5.9 &  5.7 & 14.5 \\
\cmidrule(lr){2-12}
  & \multirow{2}{*}{Qwen 2.5-VL}
    & PredCls & 37.1 & 39.2 & 38.1 & 30.0 & 27.5 & 25.5 & 37.1 & 21.7 & 21.9 & 25.9 \\
  & & SGDet   & 39.3 & 22.9 & 28.0 & 28.4 & 18.3 & 20.1 & 35.8 & 13.2 & 16.5 & 20.0 \\
\bottomrule
\end{tabular}}
\vspace{-4mm}
\end{table}

%% file: paper_tex_files/conclusion.tex
We presented \textbf{\WorldSGG{}} (\WSGG{}), a task that generalizes video scene graph
generation to temporally persistent, world-anchored graphs covering \emph{all} objects---both
observed and unobserved. To support it, we introduced \textbf{ActionGenome4D}, upgrading Action
Genome into a 4D representation via an automated pipeline leveraging \PiThree{} for 3D
reconstruction, Grounding~DINO + SAM2 for geometric annotation, and a multi-stage VLM
pseudo-labeling pipeline with manual correction for semantic annotation. The dataset provides
per-frame 3D scenes with camera poses, world-frame OBBs, and dense relationship annotations
across three predicate axes.

We proposed three methods exploring complementary inductive biases:
\PWG{} maintains a persistent visual-feature buffer grounded in object permanence;
\MWAE{} frames unseen-object reasoning as masked auto-encoder completion with cross-view
reconstruction objectives; and
\FDST{} uses a bidirectional per-object temporal transformer with camera-pose and motion
features, achieving the best results across all protocols.
All three share a unified component suite (Global Structural Encoder, Spatial GNN, Camera Pose
and Motion Encoders, Relationship Predictor, Temporal Edge Attention) and operate on pre-extracted
DINO features. Our VLM evaluation further established baselines for unlocalized \WSGG{}, showing
that current VLMs provide useful pseudo-annotations but struggle with fine-grained spatial and
contacting reasoning for off-screen entities.


\vspace{1mm}
\noindent\textbf{Future Work.} Several promising directions emerge from this study.
\emph{Online temporal reasoning:} adapting \FDST{} to streaming, variable-length observation
windows would enable real-time world-state tracking that can be used for applications such as error detection in procedural activities \cite{peddi2024captaincook4ddatasetunderstandingerrors,france2025position}.
\emph{End-to-end 3D grounding:} replacing the current multi-stage geometric pipeline with
unified 3D-aware detectors could improve both efficiency and accuracy. This can be potentially used for tasks such as navigation \cite{france2026chasing,allu2024modular}.
\emph{Open-vocabulary \WSGG{}:} broadening the object and predicate vocabularies through
vision-language grounding would move toward unconstrained scene understanding. This can be potentially used for tasks such as task guidance \cite{rheault2024predictive}.
\emph{Downstream applications:} assessing how world scene graphs benefit activity
recognition~\cite{peddi2024captaincook4ddatasetunderstandingerrors,france2025position,rheault2024predictive}, embodied navigation \cite{france2026chasing,allu2024modular}, and robot manipulation \cite{xiang2024grasping} remains an important validation step.
\emph{Long-tail predicate balance:} the substantial gap between micro- and macro-averaged F1
scores (e.g.\ 53.3 vs.\ 26.6) indicates that VLMs disproportionately favour frequent
predicates; mitigating this distributional skew is a key open challenge. Techniques from robust optimization literature~\cite{Peddi2022RobustLearning,peddi2022distributionally} can potentially act as data augmentation strategies that could be used to address the issue.

%% file: sup_tex_files/sup_motivation.tex
\begin{figure*}[!h]
\centering
\includegraphics[width=\textwidth]{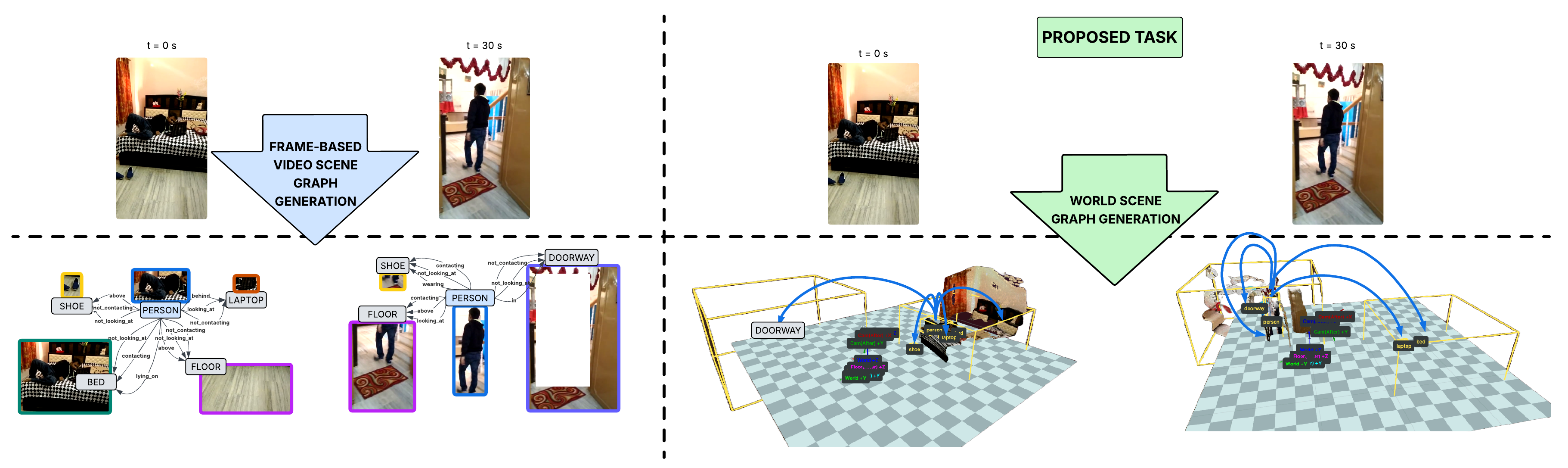}
    \caption{\textbf{World Scene Graph Generation (\WSGG{}).} \textbf{Left:} Standard Video Scene Graph Generation (VidSGG) is anchored to the instantaneous camera view. Objects are localized as 2D bounding boxes in the image plane, and relationships are predicted only for currently detected entities. When an object exits the field of view or becomes occluded, it is silently dropped from the graph along with all of its relationships, resulting in an incomplete and temporally fragmented scene understanding. \textbf{Right:} Our proposed \WSGG{} task grounds scene understanding in a persistent, global 3D world coordinate frame. Every object in the environment is represented by a 3D oriented bounding box (OBB) that persists across frames, regardless of whether the object is currently visible in the camera view. At each timestamp, the model outputs a \emph{complete} world scene graph containing \emph{all} known objects; observed ($\mathcal{O}^{t}$) and unobserved ($\mathcal{U}^{t}$); together with their pairwise semantic relationships (attention, spatial, and contacting predicates). As illustrated at $t{=}30$s, objects that have left the camera's field of view (e.g., bed, laptop) remain precisely localized in 3D world coordinates, enabling the model to continue predicting meaningful relationships for them. Blue curves drawn from person to objects denote predicted relationships. This view-independent, temporally persistent representation supports downstream tasks such as embodied navigation, robotic manipulation, and long-horizon activity understanding that require reasoning about the full state of the world, not just its currently visible slice.}
\label{sup:fig:worldsgg_task_picture}
\end{figure*}

Scene graph generation has matured considerably since the introduction of
Visual Genome~\cite{krishna2016visualgenomeconnectinglanguage}, yet the
dominant paradigm remains \emph{frame-centric}: a model observes a single
image or a short video clip and outputs a flat graph whose nodes are the
objects detected in the current view and whose edges encode pairwise semantic
relationships.  This design inherits three fundamental limitations:

\vspace{1mm}
\begin{enumerate}[nosep,leftmargin=*]
    \item \textbf{View dependence.}  Every output graph is anchored to the camera's image plane.  Object positions are expressed as 2D bounding boxes whose coordinates shift as the camera moves, offering no shared spatial reference frame.
    \item \textbf{Observation gating.}  Objects exist in the graph only if they are detected in the current frame.  When an object leaves the field of view or becomes occluded it is silently dropped from the graph.  Thus, there is no mechanism to retain its identity, location, or past relationships.
    \item \textbf{Temporal fragmentation.}  VidSGG methods that incorporate temporal modeling (e.g., STTran~\cite{cong2021spatialtemporaltransformerdynamicscene}, Tempura~\cite{nag2023unbiasedscenegraphgeneration}, ImparTail~\cite{peddi2025unbiasedrobustspatiotemporalscene}) process a sliding window of frames and produce per-frame graphs without a persistent, globally consistent world model.
\end{enumerate}

\vspace{1mm}
These limitations are tolerable for image retrieval or captioning,
where a snapshot summary suffices.  They become critical, however, for
downstream tasks that require \emph{persistent world-state reasoning}:
robotic manipulation that must track tools after they leave the camera,
embodied navigation that builds a spatial memory of traversed rooms, and
activity understanding that reasons about long-horizon human--object
interactions spanning minutes.

\vspace{1mm}
\noindent\textbf{Object Permanence as a Design Principle.}
Developmental psychology has long recognized \emph{object
permanence}~\cite{Spelke1990ObjectPerception} as a foundational cognitive
milestone: the understanding that objects continue to exist when they leave
direct perception.  Infants as young as four months exhibit surprise when
an object that was hidden behind a screen fails to reappear, implying an
internal model of the world that persists beyond the immediate sensory
input.  This principle motivates our core design decision.  A world-centric scene
graph must maintain a complete inventory of all interacting objects known to exist in
the environment; \emph{both observed} ($\mathcal{O}^{t}$) \emph{and
unobserved} ($\mathcal{U}^{t}$).  It must predict relationships involving
every interacting object at every timestamp, regardless of its current
visibility.  Concretely, the world scene graph
$\mathcal{G}_{\mathcal{W}}^{t}$ must cover observed--observed,
observed--unobserved, and unobserved--unobserved object pairs.

\subsection{Gaps in Existing Benchmarks}

\begin{table*}[!h]
\centering
\vspace{-2mm}
\caption{\textbf{SGG Tasks.} \textbf{Bold} entries highlight the unique capabilities of \WSGG{}.}
\label{sup:tab:sgg_task_comparison}
\small
\setlength{\tabcolsep}{4pt}
\resizebox{\textwidth}{!}{%
\begin{tabular}{@{}l c c c c c c c c@{}}
\toprule
\textbf{Task} & \textbf{Input} & \textbf{Spatial} & \textbf{Temporal} & \textbf{Localization} & \textbf{Object Scope} & \textbf{Coordinate} & \textbf{Unobserved} & \textbf{Relationship} \\
 &  & \textbf{Dimension} &  &  &  & \textbf{Frame} & \textbf{Objects} & \textbf{Persistence} \\
\midrule
Image SGG         & Image       & 2D & \xmark & 2D BBox & Detected         & Image & \xmark & Per-frame \\
Video SGG         & Video       & 2D & \cmark & 2D BBox & Detected / frame & Image & \xmark & Cross-frame \\
3D SGG            & 3D Scan     & 3D & \xmark & 3D BBox & All in scan      & Scene & \xmark & Per-scan \\
4D SGG            & Video + 3D  & 3D & \cmark & 3D BBox & Detected / frame & Scene & \xmark & Cross-frame \\
Panoptic SGG      & Image       & 2D & \xmark & 2D Mask & Things + Stuff   & Image & \xmark & Per-frame \\
Panoptic VSGG     & Video       & 2D & \cmark & 2D Mask & Things + Stuff   & Image & \xmark & Cross-frame \\
\midrule
\textbf{\WSGG{} (ours)} & \textbf{Video} & \textbf{3D} & \textbf{\cmark} & \textbf{3D BBox} & \textbf{World state} & \textbf{World} & \textbf{\cmark} & \textbf{Through occlusion} \\
\bottomrule
\end{tabular}%
}
\vspace{-3mm}
\end{table*}

Table~\ref{sup:tab:sgg_task_comparison}
compares existing SGG task formulations along eight diagnostic axes.  Several
observations motivate the need for \WSGG{}:
\noindent\textbf{Video SGG} (e.g., Action
Genome~\cite{ji2019actiongenomeactionscomposition}, VidVRD~\cite{xindi_et_al_vid_vrd_2017})
operates in 2D, restricts the object scope to currently detected objects, and
provides no world coordinate frame.  Relationships are defined only within
individual frames or short temporal windows.
\noindent\textbf{3D SGG} (e.g., 3D scene graphs from point
clouds~\cite{Kim20203DSceneGraph,koch2023sgrec3dselfsupervised3dscene})
localizes objects in a shared 3D coordinate frame but lacks any temporal
dimension: the graph is a static snapshot of a reconstructed scene.
\noindent\textbf{4D SGG}~\cite{yang20244dpanopticscenegraph,wu2025learning4dpanopticscene}
combines 3D localization with temporal reasoning but typically requires multi-view or RGB-D input, restricts the object scope to per-frame detections, and does not
maintain relationships for unobserved objects.
\noindent\textbf{Panoptic (V)SGG}~\cite{yang2023panopticvideoscenegraph}
enriches the object vocabulary to include both things and stuff but remains
image-plane anchored and does not reason about occluded or out-of-view entities.
No existing benchmark jointly satisfies the three requirements that
define \WSGG{}: (i)~3D spatial grounding in a world coordinate frame,
(ii)~temporally persistent object identity and tracking, and
(iii)~dense semantic annotations for \emph{all} interacting objects.

\subsection{Why Monocular Video?}

Although multi-view or RGB-D setups simplify 3D reconstruction, they impose
hardware requirements that are incompatible with many practical deployment
settings.  Egocentric and surveillance videos, internet video, and
autonomous driving footage are overwhelmingly monocular.  By grounding
\WSGG{} on monocular input, we maximize applicability while simultaneously
posing the harder perceptual challenge.  Advances in feed-forward monocular
3D reconstruction (\PiThree{}~\cite{wang2025pi}) make it feasible to recover
metrically consistent world-frame geometry from a single video, closing the
gap between monocular and multi-view pipelines for the purpose of object
localization and scene graph construction.

%% file: sup_tex_files/sup_notation.tex

For ease of reference, Table~\ref{sup:tab:notation} consolidates all
mathematical notation used throughout the main paper and supplementary
material.

\begingroup
\small
\renewcommand{\arraystretch}{1.20}

\begin{longtable}{@{}p{0.27\textwidth}@{\hspace{0.02\textwidth}}p{0.69\textwidth}@{}}
\caption{\textbf{Notation reference.}}
\label{sup:tab:notation}\\
\toprule
\textbf{Symbol / Notation} & \textbf{Description} \\
\midrule
\endfirsthead

\multicolumn{2}{@{}l}{\small\itshape Table~\ref{sup:tab:notation} continued from previous page} \\
\toprule
\textbf{Symbol / Notation} & \textbf{Description} \\
\midrule
\endhead

\midrule
\multicolumn{2}{r@{}}{\small\itshape Continued on next page} \\
\endfoot

\bottomrule
\endlastfoot

\multicolumn{2}{@{}l}{\textit{Task Formalization}} \\
$V_{1}^{T} = \{I^{1}, \dots, I^{T}\}$ & Video consisting of $T$ frames \\
$I^{t}$ & Video frame at timestamp $t$ \\
$\mathcal{W}^{t}$ & World state at timestamp $t$ (all objects in the scene) \\
$\mathcal{O}^{t}$ & Set of observed objects visible in frame $I^{t}$ \\
$\mathcal{U}^{t}$ & Set of unobserved objects not visible in frame $I^{t}$ \\
$w_{k}^{t}$ & Object $k$ in the world state at timestamp $t$ \\
$\mathcal{G}^{t}$ & Frame-level scene graph over $\mathcal{O}^{t}$ \\
$\mathcal{G}_{\mathcal{W}}^{t}$ & World scene graph over $\mathcal{W}^{t}$ \\
$N$ & Total number of objects in the world state \\
\midrule

\multicolumn{2}{@{}l}{\textit{Object Representation}} \\
$\mathbf{b}_{k}^{t} \in \mathbb{R}^{8 \times 3}$ & Oriented 3D bounding box (8 corners) for object $k$ at $t$ \\
$\mathbf{C}_{k} \in \mathbb{R}^{8 \times 3}$ & OBB corner coordinates for object $k$ \\
$\bar{\mathbf{c}}_{k}$ & Centroid of OBB $k$ \\
$V_{k}$ & Volume of OBB $k$ \\
$C_{\text{obj}}$ & Number of object categories (37 including person) \\
$\mathbf{f}_{k}^{t} \in \mathbb{R}^{d_{\text{roi}}}$ & Per-object visual features from DINOv2/v3 \\
\midrule

\multicolumn{2}{@{}l}{\textit{Predicates and Relationships}} \\
$\mathcal{P}_{\text{att}}$ & Attention predicate set ($|\mathcal{P}_{\text{att}}| = 3$, single-label) \\
$\mathcal{P}_{\text{spa}}$ & Spatial predicate set ($|\mathcal{P}_{\text{spa}}| = 6$, multi-label) \\
$\mathcal{P}_{\text{con}}$ & Contacting predicate set ($|\mathcal{P}_{\text{con}}| = 17$, multi-label) \\
$(p, o)$ & Human--object pair (person $p$, object $o$) \\
\midrule

\multicolumn{2}{@{}l}{\textit{Camera and Geometry}} \\
$\mathbf{T}^{t} = [\mathbf{R}^{t} \,|\, \boldsymbol{\tau}^{t}]$ & Camera-to-world SE(3) extrinsic at timestamp $t$ \\
$\mathbf{R}^{t} \in SO(3)$ & Rotation component of the camera pose \\
$\boldsymbol{\tau}^{t} \in \mathbb{R}^{3}$ & Translation component of the camera pose \\
$\mathbf{R}_{\text{rel}}, \boldsymbol{\tau}_{\text{rel}}$ & Relative rotation/translation between consecutive frames \\
$(f_x, f_y, c_x, c_y)$ & Camera intrinsics (focal lengths and principal point) \\
\midrule

\multicolumn{2}{@{}l}{\textit{Shared Architectural Components}} \\
$d_{\text{model}}$ & Model hidden dimension \\
$d_{\text{struct}}$ & Structural token dimension \\
$d_{\text{rel}}$ & Relationship token dimension \\
$d_{\text{camera}}$ & Camera feature dimension \\
$d_{\text{memory}}$ & Temporal memory dimension (4DST) \\
$\mathbf{g}_{n}^{(t)}$ & Structural token for object $n$ at frame $t$ (from GSE) \\
$\mathbf{g}_{\text{global}}$ & Global structural token (max-pooled over objects) \\
$\mathbf{s}_{i}$ & Spatial positional encoding for object $i$ \\
$d_{ij}$ & Euclidean distance between objects $i$ and $j$ \\
$\hat{\mathbf{d}}_{ij}$ & Unit direction vector from object $i$ to $j$ \\
$\rho_{ij}$ & Log-volume ratio $\log V_i - \log V_j$ \\
$\mathbf{S}^{(t)}$ & Spatial PE matrix at frame $t$ \\
$\mathbf{X}^{(t)}$ & Input token matrix at frame $t$ \\
$\mathbf{H}^{(t)} \in \mathbb{R}^{N \times d_{\text{model}}}$ & Output of the Spatial GNN at frame $t$ \\
$\overline{\mathbf{v}}^{(t)}$ & Object validity mask at frame $t$ \\
$\mathbf{h}_{n}$ & Refined object representation for object $n$ \\
$\hat{\mathbf{y}}_{n}^{\text{obj}} \in \mathbb{R}^{C_{\text{obj}}}$ & Predicted object class logits \\
$\mathbf{r}_{po}$ & Relationship token for pair $(p, o)$ \\
$\mathbf{u}_{po}$ & Union ROI features for pair $(p,o)$ \\
$\mathbf{e}_p^{\text{txt}}, \mathbf{e}_o^{\text{txt}}$ & CLIP text embeddings indexed by predicted class \\
$\mathbf{R}^{(t)}$ & Refined relationship tokens after RelTransformer \\
$\mathbf{c}_{\text{cam}}$ & Encoded camera pose token \\
$\mathbf{e}_{\tau}$ & Temporal positional embedding at position $\tau$ \\
\midrule

\multicolumn{2}{@{}l}{\textit{PWG (Persistent World Graph)}} \\
$\mathbf{m}_{n}^{(t)}$ & LKS memory buffer output for object $n$ at $t$ \\
$\tau^{*}$ & Nearest visible frame for object $n$ \\
$\Delta_{n}^{(t)}$ & Staleness: $|t - \tau^{*}|$ for object $n$ \\
$\lambda_{\text{vlm}}$ & Weight for unobserved-pair VLM pseudo-label loss \\
\midrule

\multicolumn{2}{@{}l}{\textit{4DST (4D Scene Transformer)}} \\
$\mathbf{z}_{n}^{(t)}$ & Fused input token for object $n$ at $t$ \\
$\tilde{\mathbf{z}}_{n}^{(t)}$ & Temporally refined token after Temporal Object Transformer \\
$\mathbf{e}_{\text{vis}}[\cdot]$ & Learned visibility embedding \\
\midrule

\multicolumn{2}{@{}l}{\textit{MWAE (Masked World Auto-Encoder)}} \\
$\mathbf{e}_{\text{[MASK]}}$ & Learnable mask embedding for non-visible objects \\
$p_{\text{mask}}$ & Training-time masking fraction for visible objects \\
$\hat{\mathbf{x}}_{n}^{(t)}$ & Retrieved (filled-in) token after Associative Retriever \\
$\hat{\mathbf{f}}_{n}^{(t)}$ & Reconstructed visual features for masked object \\
$\lambda_{\text{recon}}$ & Reconstruction loss weight \\
$\lambda_{\text{dom}}$ & Dominance factor for reconstruction loss \\
\midrule

\multicolumn{2}{@{}l}{\textit{Loss Functions}} \\
$\mathcal{L}_{\text{PWG}}$ & Total loss for PWG method \\
$\mathcal{L}_{\text{4DST}}$ & Total loss for 4DST method \\
$\mathcal{L}_{\text{MWAE}}$ & Total loss for MWAE method \\
$\mathcal{L}_{r}^{\text{vis}}$ & Loss for predicate $r$ on visible pairs (clean GT) \\
$\mathcal{L}_{r}^{\text{unobs}}$ & Loss for predicate $r$ on unobserved pairs (pseudo-labels) \\
$\mathcal{L}_{\text{node}}$ & Node classification loss (SGDet/SGCls only) \\
$\mathcal{L}_{\text{recon}}$ & Feature reconstruction MSE loss (MWAE) \\
$\mathcal{L}_{\text{sim}}$ & Simulated-unseen relationship loss (MWAE) \\
$\mathcal{L}_{\text{SG}}$ & Scene graph prediction loss (visible + unobserved) \\
\midrule

\multicolumn{2}{@{}l}{\textit{3D Scene Construction}} \\
$\pi^3$ & Feed-forward monocular 3D reconstruction model \\
$\mathcal{P}_{\text{local}}^{t}$ & Local 3D point cloud at frame $t$ \\
$\mathcal{P}_{\text{world}}^{t}$ & World-frame 3D point cloud at frame $t$ \\
$\mathcal{S}$ & Static scene point cloud (inpainted background) \\
$\mathcal{F}^{t}$ & Dynamic scene point cloud at frame $t$ \\
\midrule

\multicolumn{2}{@{}l}{\textit{Floor Alignment and Coordinate Transforms}} \\
$s_{g}$ & Global floor similarity scale factor \\
$\mathbf{R}_{g}, \boldsymbol{\tau}_{g}$ & Global floor similarity rotation and translation \\
$\mathbf{R}_{\text{align}}, \boldsymbol{\tau}_{\text{align}}$ & World-to-floor alignment transform \\
$\mathbf{R}_{\text{final}}, \boldsymbol{\tau}_{\text{final}}$ & World-to-final coordinate transform \\
$\mathbf{T}_{\delta}$ & Manual floor correction transform (rotation $\mathbf{R}_{\delta}$, translation $\boldsymbol{\tau}_{\delta}$, scale $\mathbf{s}_{\delta}$) \\
$\mathbf{T}_{\text{XY}}$ & Automated XY-plane alignment transform \\
$\hat{\mathbf{n}}$ & Estimated floor-plane normal \\
\midrule

\multicolumn{2}{@{}l}{\textit{Geometric Annotation}} \\
Grounding DINO & Zero-shot open-vocabulary 2D detector \\
SAM2 & Segment Anything Model 2 (image and video modes) \\
SMPL & Parametric body mesh model (floor estimation) \\
\midrule

\multicolumn{2}{@{}l}{\textit{Evaluation Metrics}} \\
$\mathrm{R}@K$ & Recall at $K$ \\
$\mathrm{mR}@K$ & Mean recall at $K$ (averaged over predicate classes) \\
$\mathrm{PredCls}$ & Predicate classification (GT labels and localizations given) \\
$\mathrm{SGDet}$ & Scene graph detection (end-to-end) \\

\end{longtable}
\endgroup

%% file: sup_tex_files/sup_extended_related_work.tex

\subsection{Structured Scene Understanding}

Scene graphs were first introduced as structured representations that encode objects and their
pairwise relationships for static images~\cite{krishna2016visualgenomeconnectinglanguage,chen2019knowledgeembeddedroutingnetworkscene,zhu2022scenegraphgenerationcomprehensive}.
SGTR~\cite{li2022sgtrendtoendscenegraph} formulated end-to-end scene graph generation via
Transformers that jointly predict entity and predicate proposals, while
scene graph generation from natural language supervision~\cite{zhong2021learninggeneratescenegraph} bridges vision--language alignment with structured scene representations.  Environment-invariant curriculum relation learning~\cite{min2023environmentinvariantcurriculumrelationlearning} progressively trains on harder relational patterns for fine-grained SGG.

Extending scene graphs to the temporal domain requires modeling how relationships evolve across
frames. The Video Visual Relation Detection (VidVRD) benchmark~\cite{xindi_et_al_vid_vrd_2017}
first formalised this task, while the Action Genome (AG) dataset~\cite{ji2019actiongenomeactionscomposition} introduced compositional spatio-temporal scene
graphs grounded in human-object interactions. Following AG, several Transformer-based methods have
been proposed: STTran~\cite{cong2021spatialtemporaltransformerdynamicscene} captures intra-frame spatial context and inter-frame temporal evolution via dual spatial-temporal attention;
memory-augmented architectures~\cite{feng2022exploitinglongtermdependenciesgenerating} exploit
long-term temporal dependencies; Tempura~\cite{nag2023unbiasedscenegraphgeneration} mitigates predicate-level imbalance; and ImparTail~\cite{peddi2025unbiasedrobustspatiotemporalscene} addresses unbiased and robust spatio-temporal scene graph generation. Recent methods further pushed architectural boundaries: OED~\cite{wang2024oedonestageendtoenddynamic} is a one-stage end-to-end VidSGG pipeline; UNO~\cite{le2026unounifyingonestagevideo} unifies one-stage VidSGG via object-centric visual representation learning; HyperGLM~\cite{nguyen2025hyperglmhypergraphvideoscene} introduces a hypergraph approach; and action scene graphs~\cite{rodin2023actionscenegraphslongform} enable long-form understanding of egocentric videos. Panoptic VidSGG~\cite{yang2023panopticvideoscenegraph} extends scene graphs to jointly reason about stuff and thing regions. SceneSayer~\cite{peddi2024scenegraphanticipation} formalised Scene Graph Anticipation (SGA).

\vspace{2mm}
\noindent\textbf{Weakly-Supervised and Open-Vocabulary} approaches have also gained traction:
natural language supervision~\cite{kim2025weaklysupervisedvideoscene} for weakly-supervised VidSGG;
TRKT for temporal-enhanced weakly-supervised~\cite{xu2025trktweaklysuperviseddynamic}
knowledge transferring; LLM4SGG~\cite{kim2024llm4sgglargelanguagemodels} for LLM-guided weakly-supervised SGG; zero-shot visual relation detection via composite visual cues from
LLMs~\cite{li2023zeroshotvisualrelationdetection}; zero-shot local SGG via foundation models~\cite{zhao2023morezeroshotlocalscene}; and VLPrompt~\cite{zhou2024vlpromptvisionlanguagepromptingpanoptic} for vision-language prompting in panoptic SGG. Additional Transformer-based VidSGG methods include:
SKET~\cite{pu2023spatialtemporalknowledgeembeddedtransformervideo}, which embeds spatial-temporal
knowledge into the Transformer; EGTR~\cite{im2024egtrextractinggraphtransformer}, which extracts graphs directly from Transformers; SpGLT~\cite{wu2025spatiotemporal}, which combines global-local Transformer features with pose-aware visibility matrices; SceneLLM~\cite{zhang2025scenellmimplicitlanguagereasoning}, which leverages implicit language reasoning in LLMs; and TD$^2$-Net~\cite{lin2024td2netdenoisingdebiasingdynamic}, which jointly denoises and debiases dynamic scene graphs.

\vspace{2mm}
\noindent\textbf{3D and 4D Scene Graph Generation} Departing from the 2D setting, 3D scene graphs encode objects and their relationships in a shared
world coordinate frame. 3D scene graphs were
pioneered~\cite{Kim20203DSceneGraph} as sparse, semantic representations for intelligent agents.
3D spatial multimodal knowledge
accumulation~\cite{Feng20233DSM3DSpatialMultimodal} advanced scene graph prediction in point
clouds.
SGRec3D~\cite{koch2023sgrec3dselfsupervised3dscene} performs self-supervised 3D scene graph
learning via object-level reconstruction; open-vocabulary functional 3D scene
graphs~\cite{zhang2025openvocabularyfunctional3dscene} target real-world indoor spaces; and Nyffeler
et al.~\cite{Nyffeler_2025_ICCV} construct hierarchical 3D scene graphs outdoors.
Statistical confidence rescoring~\cite{yeo2025statisticalconfidencerescoringrobust} improves
robust 3D SGG from multi-view images, and
FROSS~\cite{hou2025frossfasterthanrealtimeonline3d} achieves faster-than-real-time online 3D
semantic SGG from RGB-D images. 4D scene graphs extend the 3D domain to the temporal dimension. 4D panoptic scene graph generation~\cite{yang20244dpanopticscenegraph} extends scene graphs to the joint
spatial-temporal-panoptic domain; learning 4D panoptic scene graphs from rich 2D visual
scenes was explored in~\cite{wu2025learning4dpanopticscene};
and RealGraph~\cite{Lin2023RealGraphMultiview} contributes a multi-view dataset for 4D real-world
context graph generation.

\vspace{2mm}
\noindent\textbf{Scene Graph Applications.} Scene graphs have been applied beyond recognition:
semantic image manipulation~\cite{dhamo2020semanticimagemanipulationusing} and
MIGS~\cite{farshad2021migsmetaimagegeneration} demonstrate their utility for controllable
synthesis; dynamic graph representations~\cite{geng2021dynamicgraphrepresentationlearning} have been used for video dialog; SSGVS~\cite{cong2022ssgvssemanticscenegraphtovideo} synthesises videos from semantic scene graphs; Relationformer~\cite{shit2022relationformerunifiedframeworkimagetograph} provides a unified framework for image-to-graph generation; and HIG~\cite{nguyen2024highierarchicalinterlacementgraph} uses hierarchical interlacement
graphs for video understanding. In robotics, Scene Graph Memory~\cite{kurenkov2023modelingdynamicenvironmentsscene} models dynamic environments,
relational predicate grounding~\cite{migimatsu2022groundingpredicatesactions} learns spatial and
contact relationships from interactive manipulation, and ESCA~\cite{huang2025escacontextualizingembodiedagents} contextualises embodied agents via SGG.

\subsection{Vision-Language Models for Spatial Understanding}

VLMs have recently been adapted for spatial understanding tasks that require reasoning about 3D
geometry and object relationships from visual input.
SpatialVLM~\cite{chen2024spatialvlmendowingvisionlanguagemodels} endows VLMs with spatial
reasoning through internet-scale spatial data;
SpatialRGPT~\cite{cheng2024spatialrgptgroundedspatialreasoning} grounds spatial reasoning through
depth-aware region representations;
SpatialBot~\cite{cai2025spatialbotprecisespatialunderstanding} enables precise spatial
understanding with depth integration;
SD-VLM~\cite{chen2025sdvlmspatialmeasuringunderstanding} encodes depth directly for spatial
measuring;
See\&Trek~\cite{li2025seetrektrainingfreespatialprompting} introduces training-free spatial
prompting of multimodal LLMs;
fine-grained preference optimization~\cite{shen2026finegrainedpreferenceoptimizationimproves}
and perspective-aware reasoning via mental imagery
simulation~\cite{lee2025perspectiveawarereasoningvisionlanguagemodels} improve VLM spatial
capabilities;
Struct2D~\cite{zhu2025struct2dperceptionguidedframeworkspatial} is a perception-guided framework
for spatial reasoning in large multimodal models;
spatial relation capture by transformers was investigated
in~\cite{wen2024transformerscapturespatialrelations}; and
SpatialMLLM~\cite{wu2025spatialmllmboostingmllmcapabilities} boosts MLLM capabilities in
visual-based spatial intelligence.
Benchmarks such as
SITE~\cite{wang2025sitespatialintelligencethorough},
STI-Bench~\cite{li2025stibenchmllmsreadyprecise}, and
CoSpace~\cite{zhu2025cospacebenchmarkingcontinuousspace} evaluate whether MLLMs are ready for
precise spatial-temporal world understanding.

A growing line of work endows VLMs with explicit 3D understanding.
SpatialLLM~\cite{ma2025spatialllmcompound3dinformeddesign} is a compound 3D-informed design
towards spatially-intelligent large multimodal models;
SpatialReasoner~\cite{ma2025spatialreasonerexplicitgeneralizable3d} enables explicit and
generalizable 3D spatial reasoning;
SpatialLM~\cite{mao2025spatiallmtraininglargelanguage} trains large language models for
structured indoor modeling;
scalable spatial intelligence via 2D-to-3D data lifting was proposed
in~\cite{miao2025scalablespatialintelligence2dto3d};
LayoutVLM~\cite{sun2025layoutvlmdifferentiableoptimization3d} introduces differentiable
optimisation of 3D layout via vision-language models;
MindJourney~\cite{yang2025mindjourneytesttimescalingworld} leverages test-time scaling with world
models for spatial reasoning;
and learning from videos with 3D vision-geometry
priors~\cite{zheng2025learningvideos3dworld} enhances multimodal LLMs.

\subsection{Neural Memory Models}

Modern Hopfield networks and dense associative memories offer exponential storage
capacity~\cite{Lucibello2024ExponentialCapacity,hu2024provablyoptimalmemorycapacity,agliari2020toleranceversussynapticnoise,clark2025transientdynamicsassociativememory}
and differentiable retrieval dynamics, making them attractive for architectures requiring persistent
state. The Hopfield framework has been extended to long sequences as attention-compatible memory
layers~\cite{chaudhry2023longsequencehopfieldmemory}, while biological
extensions~\cite{Leo2025NeuronAstrocyte} and connections to diffusion
models~\cite{hoover2024memoryplainsightsurveying,hoover2024denseassociativememorylens} have
broadened the theoretical foundations. These mechanisms have been integrated into large-scale
models: CAMELoT~\cite{he2024camelotlargelanguagemodels} serves as a training-free associative
memory for frozen LLMs, M+~\cite{wang2025mextendingmemoryllmscalable} scales knowledge retention
via long-term memory, and CLOOB~\cite{fürst2022cloobmodernhopfieldnetworks} for contrastive
vision-language pre-training.

\vspace{2mm}
\noindent\textbf{Prototypical networks.} ~\cite{snell2017prototypicalnetworksfewshot} introduced the idea of
learning a metric space in which classification is performed by computing distances to prototype
representations of each class, establishing a foundational framework for few-shot learning.
This paradigm has been extended in several directions:
Graph Prototypical Networks~\cite{ding2020graphprototypical} adapt prototypical representations
to attributed networks, enabling few-shot node classification by constructing a pool of
semi-supervised tasks that mimic the test environment;
ProFusion~\cite{zhao2025profusion} augments prototypical networks with multimodal feature fusion,
constructing image, text, and fused prototypes from vision-language pre-trained models for
robust few-shot classification;
prototype memory~\cite{smirnov2022prototypememorylargescale} scales prototypical representations
for large-scale face recognition; and
training-free prototype calibration~\cite{wang2023fewshot} addresses few-shot class-incremental
learning by adjusting prototypes without retraining.
In the context of scene graph generation,
RA-SGG~\cite{yoon2024rasggretrievalaugmentedscenegraph} proposes retrieval-augmented SGG via multi-prototype learning,
where multiple prototypes per predicate class capture the diversity of visual relationship
appearances and mitigate the long-tail distribution problem inherent in relationship annotations.

%% file: sup_tex_files/sup_scene_construction.tex
\begin{figure}[t]
    \centering
    \includegraphics[width=\textwidth]{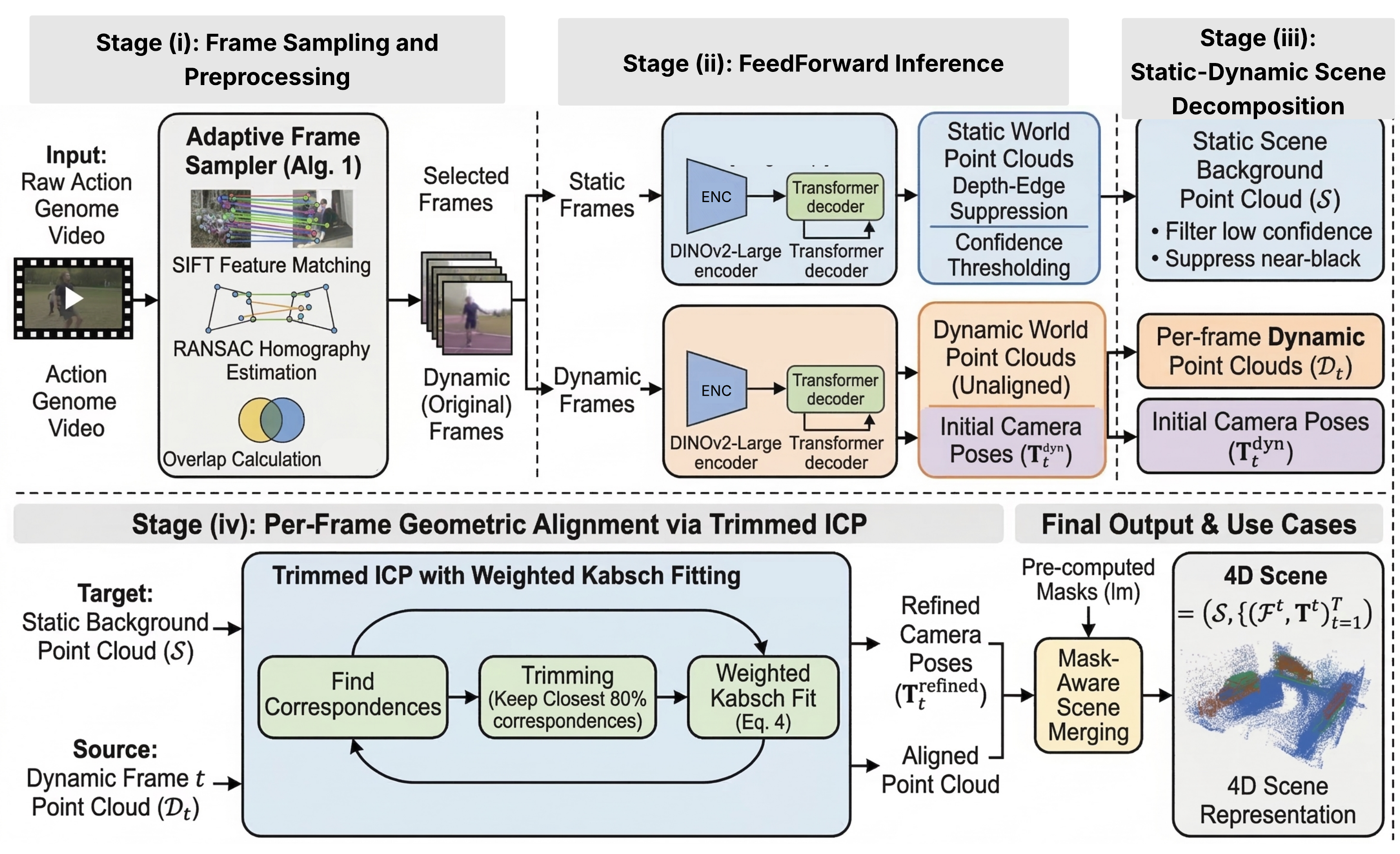}
    \caption{Pipeline for constructing a 4D scene representation from Action Genome videos. The method first selects informative frames through adaptive sampling, then predicts static and dynamic world point clouds, decomposes the scene into background and per-frame dynamic components, and finally aligns dynamic point clouds to the static scene using trimmed ICP before mask-aware merging into the final 4D output.}
    \label{sup:fig:scene_pipeline}
\end{figure}

\label{sup:sec:scene_construction}
We construct per-timestamp 3D scene representations from egocentric Action Genome~\cite{ji2019actiongenomeactionscomposition} videos using a feed-forward neural reconstruction model followed by post-hoc geometric alignment.
The pipeline comprises four stages: (i)~frame sampling and preprocessing, (ii)~feed-forward 3D inference via the \PiThree~\cite{wang2025pi} model, (iii)~static--dynamic scene decomposition, and (iv)~per-frame alignment via trimmed Iterative Closest Point (ICP) with weighted Kabsch fitting.
The full pipeline is illustrated in Figure~\ref{sup:fig:scene_pipeline}.

\paragraph{4D Scene Representation.}
The output of the pipeline is a single 4D scene per video:
\begin{equation}
    \text{4D Scene} = \Bigl(
        \underbrace{\mathcal{S}}_{\text{static background}} ,\;\;
        \bigl\{
            \underbrace{(\mathcal{F}^{t},\; \mathbf{T}^{t})}_
                {\text{per-frame dynamic}}
        \bigr\}_{t=1}^{T}
    \Bigr)
\end{equation}
where $\mathcal{S}$ is the static background point cloud (built once and shared across all frames), $\mathcal{F}^{t}$ is the per-frame dynamic point cloud for frame $I^{t}$, and $\mathbf{T}^{t} \in \mathbb{R}^{4 \times 4}$ is the ICP-refined camera-to-world pose.
This 4D scene provides the geometric substrate from which the world state $\mathcal{W}^{t}$ is populated: object detections within each frame's point cloud $\mathcal{F}^{t}$ yield the 3D oriented bounding boxes $\mathbf{b}_{k}^{t}$ for all interacting objects, while the refined poses $\mathbf{T}^{t}$ enable camera-relative reasoning for visibility determination.
This representation supports two use cases:
\begin{itemize}
    \item \textbf{World-frame 3D bounding boxes}: Per-frame foreground points provide the geometric basis for fitting OBB annotations to each object.
    \item \textbf{Scene graph construction}: The grounded full-frame representation provides complete spatial context for reasoning about relationships across time.
\end{itemize}

\subsection{Adaptive Frame Sampling via Feature Descriptor Matching}
\label{sup:sec:frame_sampling}

Raw Action Genome~\cite{ji2019actiongenomeactionscomposition} videos are captured at 24--30\,fps, producing substantial temporal redundancy.
Rather than uniform subsampling (e.g., every $k$-th frame), we employ a \emph{Feature Descriptor Sampler} that adapts its density to camera/object motion: it retains more frames during rapid motion and fewer during static shots, using SIFT feature matching~\cite{lowe2004distinctive} and RANSAC homography estimation~\cite{fischler1981random} to quantify visual overlap.

\paragraph{Homography-Based Overlap Estimation.}
\label{sup:sec:homography_overlap}
The overlap between a retained reference frame $I_A$ and a candidate frame $I_B$ is estimated in four stages:

\begin{enumerate}
    \item \textbf{SIFT extraction}: Both frames are converted to grayscale and SIFT~\cite{lowe2004distinctive} keypoints $\mathcal{K}$ and 128-d descriptors $\mat{D}$ are extracted.  If either frame yields $<4$ keypoints, the overlap defaults to 0.

    \item \textbf{Feature matching}: Descriptors in $I_A$ are matched to $I_B$ via brute-force $k$-NN ($k{=}2$) with Lowe's ratio test~\cite{lowe2004distinctive} ($\rho = 0.75$), retaining only matches where the best candidate is substantially better than the second-best:
    \begin{equation}
        \mathcal{G} = \bigl\{(i, j) : {\|\mat{d}_i^A - \mat{d}_{j_1}^B\|_2}\,/\,{\|\mat{d}_i^A - \mat{d}_{j_2}^B\|_2} < \rho\bigr\}.
        \label{sup:eq:ratio_test}
    \end{equation}

    \item \textbf{RANSAC homography}: A projective transformation $\mat{H} \in \mR^{3 \times 3}$ mapping $I_B$ into $I_A$'s coordinate frame is estimated from the good matches via RANSAC~\cite{fischler1981random} (reprojection threshold 4.0\,px, 2000 iterations, 0.995 confidence).  The homography is rejected if $<4$ inliers remain.

    \item \textbf{Polygon intersection}: The four corners of $I_B$ are warped into $I_A$'s frame via $\mat{H}$, and the visual overlap is the normalised intersection area:
    \begin{equation}
        \alpha = \frac{\op{Area}(\mathcal{P}_A \cap \mathcal{P}_B')}{\op{Area}(\mathcal{P}_A)} \in [0, 1],
        \label{sup:eq:overlap}
    \end{equation}
    where $\mathcal{P}_A$ is the frame~$A$ rectangle and $\mathcal{P}_B'$ is the warped frame~$B$ quadrilateral, computed using the Shapely geometry library.
\end{enumerate}

While Action Genome scenes are not strictly planar, the homography provides a sufficiently accurate alignment for overlap estimation in indoor environments dominated by large planar surfaces (walls, floors, tables).  The RANSAC inlier ratio also serves as an implicit quality measure; low inlier counts indicate strong parallax or dynamic content, both signals of significant visual change.

\begin{algorithm}[!t]
\caption{Greedy Frame Selection by Visual Overlap}
\label{sup:alg:greedy_sampling}
\begin{algorithmic}[1]
\Require Frames $\{I_0, \ldots, I_{T-1}\}$, threshold $\tau$
\Ensure Selected frame indices $\mathcal{S}$
\State $\mathcal{S} \gets \{0\}$; $I_{\text{ref}} \gets I_0$
\For{$t = 1, \ldots, T-1$}
    \State $\alpha \gets \op{HomographyOverlap}(I_{\text{ref}}, I_t)$
    \If{$\alpha < \tau$} \Comment{Sufficient new content}
        \State $\mathcal{S} \gets \mathcal{S} \cup \{t\}$; $I_{\text{ref}} \gets I_t$
    \EndIf
\EndFor
\State \Return $\mathcal{S}$
\end{algorithmic}
\end{algorithm}

\paragraph{Greedy Adaptive Selection.}
\label{sup:sec:greedy_selection}
Given a video with $T$ frames and an overlap threshold $\tau = 0.95$, Algorithm~\ref{sup:alg:greedy_sampling} greedily builds the selected set~$\mathcal{S}$.
The reference $I_{\text{ref}}$ advances only upon frame retention, creating a natural temporal hysteresis: the reference stays fixed during static shots (skipping many near-duplicates) and advances rapidly during camera or object motion. At $\tau = 0.95$, any frame contributing ${\geq}\,5\%$ novel visual content relative to the current reference is kept. The overlap-based criterion mirrors keyframe insertion in visual SLAM~\cite{MurArtal2015OrbSlam} and SfM~\cite{Schnberger2016StructurefromMotionR}, where a new keyframe is triggered once tracked-feature counts fall below a reference threshold; a condition equivalent to low homography overlap.
Our formulation makes this criterion geometrically explicit. Thus, the sampler compresses Action Genome videos from ${\sim}900$--$9000$ raw frames to ${\sim}30$--$150$ frames.

\textbf{Caveats:} (a) To preserve all supervised training targets, every frame carrying a ground-truth bounding-box annotation is unconditionally added after overlap filtering: $\mathcal{S} \gets \op{sort}(\mathcal{S} \cup \mathcal{A}_{\text{annotated}})$. (b) If the resulting set contains fewer than $K_{\min} = 17$ frames, the sampler reverts to uniform subsampling at stride $\Delta = \max(1, \lfloor T / K_{\min} \rfloor)$ and re-injects annotated frames, ensuring enough distinct viewpoints for reliable 3D reconstruction. 

\subsection{Feed-Forward 3D Inference via \PiThree}
\label{sup:sec:pi3_inference}

We employ the \PiThree~\cite{wang2025pi} model, a feed-forward neural network for visual geometry reconstruction.
Given $N$ unposed images, \PiThree jointly predicts per-pixel 3D points, per-pixel confidence scores, and camera-to-world SE(3) poses in a single forward pass, without requiring a designated reference view.
The model uses a frozen DINOv2-Large~\cite{oquab2024dinov2learningrobustvisual} encoder and a permutation-equivariant transformer decoder with alternating intra-view and cross-view attention, ensuring the reconstruction is invariant to the ordering of input views.
We use \PiThree as a black-box reconstruction module; we refer the reader to~\cite{wang2025pi} for full architectural details.

\paragraph{Input Frame Preparation.}
\label{sup:sec:frame_preparation}
From the selected frame set $\mathcal{S}$, we prepare two parallel sets of inputs: (a) \textbf{Static frames}: Inpainted versions in which all dynamic foreground objects (persons, interacted objects) have been removed via rectangular mask inpainting, depicting only the static background. (b) \textbf{Dynamic frames}: The original unmodified video frames at the same indices, containing all actors and objects in their natural positions. Both sets share identical indices and resolution to enable pixel-level correspondence between the reconstructions.

\paragraph{Inference Procedure.}
\label{sup:sec:inference_procedure}
The static and dynamic frame sets are processed in \textbf{two independent forward passes} in mixed precision (bfloat16/float16). Each pass takes $N$ frames batched as a tensor of shape $(1, N, 3, H, W)$ and produces all outputs simultaneously. This independence is critical: the two reconstructions inhabit different, unaligned world coordinate systems, necessitating the geometric alignment described in Section~\ref{sup:sec:icp_alignment}. For a set of $N$ input images, the model jointly estimates the quantities summarized in Table~\ref{sup:tab:pi3_outputs}.

\begin{table}[t]
\centering
\caption{Quantities estimated by the \PiThree model for $N$ input views at resolution $H \times W$.}
\label{sup:tab:pi3_outputs}
\begin{tabular}{@{}llll@{}}
\toprule
\textbf{Quantity} & \textbf{Shape} & \textbf{Type} & \textbf{Description} \\
\midrule
Local points       & $(N, H, W, 3)$ & float32 & Per-pixel 3D coordinates in camera frame \\
World points       & $(N, H, W, 3)$ & float32 & Per-pixel 3D coordinates in world frame \\
Confidence         & $(N, H, W, 1)$ & float32 & Per-pixel reliability score $\in [0,1]$ \\
Camera poses       & $(N, 4, 4)$    & float32 & Camera-to-world SE(3) matrices \\
\bottomrule
\end{tabular}
\end{table}

\paragraph{Post-Processing.}
\label{sup:sec:postprocessing}
Two filtering steps are applied to the raw predictions before downstream use:
(i)~\textbf{Depth-edge suppression}: pixels at depth discontinuities (relative depth difference $>3\%$ in a $3{\times}3$ neighborhood) have their confidence set to zero;
(ii)~\textbf{Confidence thresholding}: points below $\tau_{\text{static}} = 0.1$ (static background) or $\tau_{\text{frame}} = 0.01$ (per-frame dynamic) are discarded.
World-frame 3D points are obtained by applying the predicted camera-to-world pose $\mathbf{T}_i$ to the local points via homogeneous transformation: $\mathbf{p}_{\text{world}} = (\mathbf{T}_i \cdot \tilde{\mathbf{p}}_{\text{local}})_{1:3}$.

\subsection{Static-Dynamic Scene Decomposition}
\label{sup:sec:decomposition}

The two-pass inference strategy produces complementary scene representations:

\paragraph{Static Scene.}
\label{sup:sec:static_scene}
The static reconstruction uses inpainted frames as input, yielding a clean, object-free 3D point cloud of the background environment.
This point cloud is constructed by:
\begin{enumerate}
    \item Applying a confidence threshold $\tau_{\text{static}} = 0.1$ to filter low-quality points.
    \item Suppressing near-black pixels (per-channel intensity $\leq 8$) that coincide with low confidence ($c < 1.0$), which removes inpainting artifacts that often manifest as dark, uncertain regions.
    \item Removing all points with non-finite coordinates (NaN or Inf).
\end{enumerate}

The resulting point cloud $\mathcal{S} = \{(\mathbf{s}_j, \mathbf{c}_j^{\text{rgb}})\}_{j=1}^{M}$ provides a stable geometric reference frame for subsequent alignment.
The static scene is computed once per video and serves as the persistent, time-invariant component of the 4D scene.

\paragraph{Dynamic Scene.}
\label{sup:sec:dynamic_scene}
The dynamic reconstruction uses original (unmodified) frames, producing per-frame point clouds that include all actors and objects. Each frame $t$ yields a set of world-frame 3D points $\mathcal{D}_t = \{(\mathbf{d}_k^t, \mathbf{c}_k^{t,\text{rgb}})\}$ and camera pose $\mathbf{T}_t^{\text{dyn}}$.

\subsection{Per-Frame Geometric Alignment via Trimmed ICP}
\label{sup:sec:icp_alignment}

Since the static and dynamic scenes are reconstructed independently (through separate feed-forward passes of the \PiThree model), their world coordinate frames are not aligned.
We register each dynamic frame's point cloud to the static background via \textbf{Trimmed Iterative Closest Point (ICP)} with \textbf{confidence-weighted Kabsch fitting}.
This can be viewed as a form of post-hoc bundle adjustment that refines both the 3D point positions and camera poses to be consistent with the static reference frame.

\paragraph{Algorithm Overview.}
\label{sup:sec:icp_algorithm}
For each frame $t$, the alignment proceeds according to Algorithm~\ref{sup:alg:trimmed_icp}.
The source cloud is the dynamic frame's point set $\mathcal{D}_t$, and the target is the static background $\mathcal{S}$.

\begin{algorithm}[t]
\caption{Trimmed ICP with Weighted Kabsch Alignment}
\label{sup:alg:trimmed_icp}
\begin{algorithmic}[1]
\Require Source points $\mathbf{A} = \{\mathbf{a}_k\}_{k=1}^{K} \subset \mathbb{R}^3$ (dynamic frame)
\Require Target points $\mathbf{B} = \{\mathbf{b}_j\}_{j=1}^{M} \subset \mathbb{R}^3$ (static background)
\Require Per-point confidence weights $\{w_k\}_{k=1}^{K}$
\Require Parameters: $I_{\max}=100$, $\epsilon=10^{-5}$, $\rho=0.8$
\Ensure Rigid transform $(\mathbf{R}_{\text{total}}, \boldsymbol{\tau}_{\text{total}})$
\State Build KD-tree $\mathcal{T}$ on target $\mathbf{B}$
\State $\mathbf{R}_{\text{total}} \leftarrow \mathbf{I}_3$, $\boldsymbol{\tau}_{\text{total}} \leftarrow \mathbf{0}$
\State $\mathbf{A}' \leftarrow \mathbf{A}$ \Comment{working copy}
\For{$\text{iter} = 1$ \textbf{to} $I_{\max}$}
    \State $(\{d_k\}, \{\text{nn}_k\}) \leftarrow \mathcal{T}.\text{query}(\mathbf{A}', k=1)$ \Comment{nearest neighbors}
    \State $\mathcal{V} \leftarrow \{k : d_k < \infty\}$ \Comment{valid correspondences}
    \State $\delta \leftarrow \text{Percentile}(\{d_k\}_{k \in \mathcal{V}},\; \rho \cdot 100)$ \Comment{trimming cutoff}
    \State $\mathcal{V} \leftarrow \{k \in \mathcal{V} : d_k \leq \delta\}$ \Comment{keep closest $\rho$ fraction}
    \If{$|\mathcal{V}| < 10$}
        \State \textbf{break}
    \EndIf
    \State $(\mathbf{R}_{\text{inc}}, \boldsymbol{\tau}_{\text{inc}}) \leftarrow \text{WeightedKabsch}(\mathbf{A}'_{\mathcal{V}},\; \mathbf{B}_{\text{nn}_{\mathcal{V}}},\; \{w_k\}_{k \in \mathcal{V}})$
    \State $\mathbf{R}_{\text{total}} \leftarrow \mathbf{R}_{\text{inc}} \cdot \mathbf{R}_{\text{total}}$
    \State $\boldsymbol{\tau}_{\text{total}} \leftarrow \mathbf{R}_{\text{inc}} \cdot \boldsymbol{\tau}_{\text{total}} + \boldsymbol{\tau}_{\text{inc}}$
    \State $\mathbf{A}' \leftarrow (\mathbf{R}_{\text{inc}} \cdot \mathbf{A}'^{\top})^{\top} + \boldsymbol{\tau}_{\text{inc}}^{\top}$ \Comment{update source}
    \State $e \leftarrow \frac{1}{|\mathcal{V}|}\sum_{k \in \mathcal{V}} \|\mathbf{a}'_k - \mathbf{b}_{\text{nn}_k}\|^2$ \Comment{MSE}
    \If{$|e_{\text{prev}} - e| < \epsilon$}
        \State \textbf{break} \Comment{converged}
    \EndIf
\EndFor
\Return $(\mathbf{R}_{\text{total}}, \boldsymbol{\tau}_{\text{total}})$
\end{algorithmic}
\end{algorithm}

\paragraph{Weighted Kabsch Algorithm.}
\label{sup:sec:kabsch}
At each ICP iteration, the optimal rigid transformation aligning source correspondences $\mathbf{A}_{\mathcal{V}}$ to target correspondences $\mathbf{B}_{\mathcal{V}}$ is found by solving the weighted least-squares problem:
\begin{equation}
    \mathbf{R}^*, \boldsymbol{\tau}^* = \operatorname*{arg\,min}_{\mathbf{R} \in \text{SO}(3),\; \boldsymbol{\tau} \in \mathbb{R}^3} \sum_{k \in \mathcal{V}} w_k \| \mathbf{R} \mathbf{a}_k + \boldsymbol{\tau} - \mathbf{b}_{\text{nn}_k} \|^2
    \label{sup:eq:kabsch_objective}
\end{equation}
where $w_k$ is the per-point confidence from the \PiThree model's sigmoid-activated confidence output. The closed-form solution proceeds as follows:

\begin{enumerate}
    \item \textbf{Weighted centroids}:
    \begin{equation}
        \boldsymbol{\mu}_A = \frac{\sum_k w_k \mathbf{a}_k}{\sum_k w_k}, \quad
        \boldsymbol{\mu}_B = \frac{\sum_k w_k \mathbf{b}_k}{\sum_k w_k}
    \end{equation}
    
    \item \textbf{Cross-covariance matrix}:
    \begin{equation}
        \mathbf{H} = \sum_{k \in \mathcal{V}} w_k (\mathbf{a}_k - \boldsymbol{\mu}_A)(\mathbf{b}_k - \boldsymbol{\mu}_B)^{\top} \in \mathbb{R}^{3 \times 3}
    \end{equation}
    
    \item \textbf{SVD decomposition}: $\mathbf{H} = \mathbf{U} \boldsymbol{\Sigma} \mathbf{V}^{\top}$
    
    \item \textbf{Optimal rotation} with reflection correction:
    \begin{equation}
        \mathbf{R}^* = \mathbf{V} \cdot \text{diag}\bigl(1,\; 1,\; \det(\mathbf{V}\mathbf{U}^{\top})\bigr) \cdot \mathbf{U}^{\top}
        \label{sup:eq:kabsch_rotation}
    \end{equation}
    The diagonal correction ensures $\det(\mathbf{R}^*) > 0$, i.e., $\mathbf{R}^* \in \text{SO}(3)$.
    
    \item \textbf{Optimal translation}:
    \begin{equation}
        \boldsymbol{\tau}^* = \boldsymbol{\mu}_B - \mathbf{R}^* \boldsymbol{\mu}_A
    \end{equation}
\end{enumerate}

\paragraph{Trimming Strategy.}
\label{sup:sec:trimming}
Standard ICP is sensitive to outlier correspondences particularly problematic here because dynamic objects (persons, manipulated items) are present in the source but absent from the static target.
These points will inevitably match to incorrect static surface regions. We employ \textbf{trimmed ICP} with fraction $\rho = 0.8$: at each iteration, only the closest 80\% of correspondence pairs (ranked by Euclidean distance) are retained for the Kabsch fit. This allows up to 20\% of source points to be outliers without corrupting the alignment.

\paragraph{Camera Pose Refinement.}
\label{sup:sec:pose_refinement}
The ICP-derived rigid transform $\mathbf{T}_{\text{icp}} \in \text{SE}(3)$ is composed with the \PiThree-predicted camera pose to produce a refined pose aligned to the static reference frame.
For camera-to-world (c2w) convention poses:
\begin{equation}
    \mathbf{T}_t^{\text{refined}} = \mathbf{T}_{\text{icp}} \cdot \mathbf{T}_t^{\text{dyn}}
    \label{sup:eq:pose_update_c2w}
\end{equation}
This follows from the fact that points in the dynamic world frame are transformed as $\mathbf{p}' = \mathbf{T}_{\text{icp}} \cdot \mathbf{p}$, so the camera-to-world mapping must be left-multiplied by the same transform. For the world-to-camera (w2c) convention we follow:
\begin{equation}
    \mathbf{E}_t^{\text{refined}} = \mathbf{E}_t^{\text{dyn}} \cdot \mathbf{T}_{\text{icp}}^{-1}
    \label{sup:eq:pose_update_w2c}
\end{equation}



\subsection{Mask-Aware Scene Merging}
\label{sup:sec:mask_merging}

For the final 4D scene construction, we leverage pre-computed interaction segmentation masks to selectively merge dynamic content with the static background. Per-frame binary interaction masks $\mathbf{M}_t \in \{0, 1\}^{H \times W}$ are loaded from pre-computed segmentation outputs (from the SAM2 segmentation pipeline).

\paragraph{Point Partitioning.}
For each frame~$t$, the dynamic frame's valid points are split into two sets: (1) $\mathcal{D}_t^\text{all}$: All valid points (used for ICP alignment). (2) $\mathcal{D}_t^\text{fg} = \{k \in \mathcal{D}_t^\text{all} : \mathbf{M}_t(\text{pixel}(k)) > 0\}$: Foreground-only points (used for merging).

\paragraph{Selective Merging Procedure.}
The merging follows a two-stage approach:

\begin{enumerate}
    \item \textbf{ICP alignment (full frame)}: The ICP registration (Section~\ref{sup:sec:icp_algorithm}) uses \emph{all} valid dynamic points $\mathcal{D}_t^\text{all}$ in frame $t$---including both foreground and background---to estimate the rigid transform $\mathbf{T}_\text{icp}^{(t)}$.
    Using the full frame ensures a robust fit dominated by the abundant background correspondences.
    
    \item \textbf{Foreground-only merging}: Only the masked foreground points $\mathcal{D}_t^\text{fg}$ are transformed by $\mathbf{T}_\text{icp}^{(t)}$ and concatenated with the static background.
    This prevents duplication of background geometry (which already exists in the static cloud) while placing detected objects into the static reference frame.
\end{enumerate}

\noindent We then obtain the final merged point cloud for frame $t$ (see Eq~\ref{sup:eq:merged_scene}); then perform voxel deduplication.
\begin{equation}
    \mathcal{P}_t = \mathcal{S} \cup \bigl\{ \mathbf{T}_\text{icp}^{(t)} \cdot \mathbf{d}_k^t \;\big|\; k \in \mathcal{D}_t^\text{fg} \bigr\}
    \label{sup:eq:merged_scene}
\end{equation}

\subsection{Implementation Details}
\label{sup:sec:scene_implementation}

Table~\ref{sup:tab:scene_hyperparams} summarizes the key parameters of the scene construction pipeline. All computations were performed on a single NVIDIA GPU.

\begin{table}[t]
\centering
\caption{Scene construction pipeline parameters.}
\label{sup:tab:scene_hyperparams}
\small
\begin{tabular}{@{}llp{4.2cm}@{}}
\toprule
\textbf{Parameter} & \textbf{Value} & \textbf{Description} \\
\midrule
Backbone                 & DINOv2-L (ViT-L/14)    & Frozen encoder with registers \\
Decoder depth            & 36 layers               & Alternating intra-/cross-view attn \\
Positional encoding      & RoPE2D ($f{=}100$)      & 2D rotary position embeddings \\
Precision                & bfloat16 / float16      & Mixed-precision inference \\
Frame sampling           & 10                      & Temporal stride for selection \\
$\tau_{\text{static}}$   & 0.1                     & Confidence thr.\ (static scene) \\
$\tau_{\text{frame}}$    & 0.01                    & Confidence thr.\ (dynamic) \\
Depth-edge rtol          & 0.03                    & Rel.\ tolerance for edge suppression \\
ICP iterations           & 100 (max)               & Early stop at $\epsilon{=}10^{-5}$ \\
Trim fraction $\rho$     & 0.8                     & Keep closest 80\% of corresp. \\
Dynamic voxel            & 0.01\,m                 & Per-frame point reduction \\
Merge voxel              & 0.02\,m                 & Static--dynamic overlap removal \\
Vertical FOV             & 0.96\,rad ($\approx 55^{\circ}$)  & Pinhole camera model \\
\bottomrule
\end{tabular}
\end{table}

%% file: sup_tex_files/sup_geometric_annotation.tex
\subsection{3D Bounding Box Pipeline Overview}
\label{sup:sec:bbox_pipeline}

\begin{figure}[t]
\centering
\resizebox{\textwidth}{!}{%
\begin{tikzpicture}[
    node distance=0.4cm and 0.7cm,
    block/.style={rectangle, draw, rounded corners, minimum height=0.65cm, minimum width=1.5cm, align=center, font=\scriptsize},
    arrow/.style={-{Stealth[length=2mm]}, thick},
]
\node[block, fill=blue!10] (frames) {Video Frames\\$+$ 2D Annotations};

\node[block, fill=green!10, above right=0.35cm and 0.9cm of frames] (gdino) {Grounding DINO\\Detection};
\node[block, fill=green!10, below right=0.35cm and 0.9cm of frames] (gt) {GT Annotation\\Fusion};

\node[block, fill=orange!10, right=0.7cm of gdino, yshift=-0.5cm] (nms) {Class-wise NMS\\(Merged)};

\node[block, fill=red!10, above right=0.35cm and 0.7cm of nms] (sam2im) {SAM2\\Image Mode};
\node[block, fill=red!10, below right=0.35cm and 0.7cm of nms] (sam2vid) {SAM2\\Video Mode};

\node[block, fill=purple!10, right=0.7cm of sam2im, yshift=-0.5cm] (union) {Mask Union\\(per object)};

\node[block, fill=cyan!10, right=0.7cm of union] (bb3d) {3D Point Extraction\\$+$ Floor Alignment};

\node[block, fill=yellow!15, right=0.7cm of bb3d] (obb) {OBB Estimation\\$+$ Temporal Smoothing};

\draw[arrow] (frames) -- (gdino);
\draw[arrow] (frames) -- (gt);
\draw[arrow] (gdino) -- (nms);
\draw[arrow] (gt) -- (nms);
\draw[arrow] (nms) -- (sam2im);
\draw[arrow] (nms) -- (sam2vid);
\draw[arrow] (sam2im) -- (union);
\draw[arrow] (sam2vid) -- (union);
\draw[arrow] (union) -- (bb3d);
\draw[arrow] (bb3d) -- (obb);
\end{tikzpicture}%
}
\caption{\textbf{3D bounding box pipeline overview.} Objects are detected via Grounding DINO fused with GT annotations, segmented via SAM2 in dual mode, projected into 3D via masked point extraction from the \PiThree reconstruction, aligned to the floor plane, and enclosed with oriented bounding boxes (OBBs).}
\label{sup:fig:bbox_pipeline}
\end{figure}
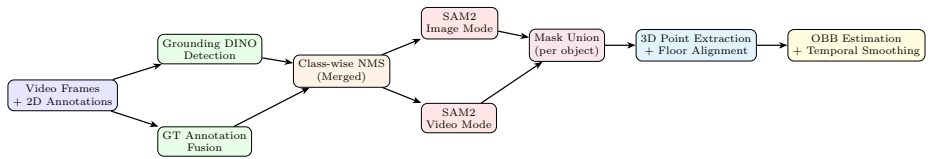

We present a multi-stage pipeline that produces world-frame \textbf{oriented 3D bounding boxes (OBBs)} for all interacting objects in Action Genome~\cite{ji2019actiongenomeactionscomposition} video clips. Starting from RGB frames and 2D annotations, the pipeline proceeds through five major stages: (i)\textbf{Detection:}~object detection via Grounding DINO~\cite{liu2024groundingdinomarryingdino} with ground-truth annotation fusion, (ii)\textbf{Segmentation:}~instance segmentation via SAM2~\cite{ravi2024sam2segmentimages} in dual image-and-video mode, (iii)\textbf{Object Classification:}~static/dynamic object classification via LLM-based reasoning, (iv)\textbf{Object Point Cloud Determination:}~floor-aligned 3D point extraction with multiscale mask erosion, and (v)\textbf{OBB Estimation:}~oriented bounding box estimation via PCA and floor-parallel minimum-area fitting, with temporal smoothing. The full pipeline is illustrated in Figure~\ref{sup:fig:bbox_pipeline}.

\subsection{Object Detection}
\label{sup:sec:detection}

The detection stage produces per-frame, per-object 2D bounding boxes by fusing model-based detections with ground-truth annotations.

\paragraph{Grounding DINO Detection.}
\label{sup:sec:gdino}
We employ Grounding DINO~\cite{liu2024groundingdinomarryingdino} as a zero-shot, open-vocabulary detector. Each frame is prompted with the video's active object labels (e.g., \texttt{"person. chair. cup."}) using a box threshold of $0.25$ and text threshold of $0.3$. Output labels are normalized to a canonical vocabulary.

\paragraph{Active Object Labels \& Static/Dynamic Classification.}
\label{sup:sec:active_objects}
Active object categories are determined from two sources: (a) labels extracted directly from Action Genome ground-truth annotations, and (b) labels produced by an LLM-based reasoning pipeline using LLaMA~\cite{touvron2023llamaopenefficientfoundation}.
\label{sup:sec:static_dynamic}
The LLM classifies objects as \textit{static} (e.g., floor, sofa, table) or \textit{dynamic} (e.g., cup, phone, book) by extracting interactions from video captions and mapping them to the candidate label set via synonym-aware matching. This drives two separate detection-segmentation tracks: a static track for background elements and a dynamic track for actively used objects.

\paragraph{Ground Truth Annotation Fusion.}
\label{sup:sec:gt_fusion}
To ensure annotated objects are not missed (especially small or occluded items), GT bounding boxes are assigned a pseudo-score of $1.001$ and concatenated with Grounding DINO predictions. Two rounds of per-class NMS (IoU threshold $0.5$) are applied; because GT scores exceed $1.0$, they are preferred over detector duplicates while still allowing new discoveries.

\subsection{Instance Segmentation via SAM2}
\label{sup:sec:segmentation}

Given per-frame detected bounding boxes, we produce per-object binary segmentation masks using SAM2~\cite{ravi2024sam2segmentimages} (hiera-base-plus) in two complementary modes.

\paragraph{Image Mode.}
\label{sup:sec:sam2_image}
SAM2 processes each frame independently: per-label bounding boxes are passed as box prompts in single-mask mode, and the resulting per-label masks are combined via boolean OR to produce one union mask per frame.

\paragraph{Video Mode.}
\label{sup:sec:sam2_video}
SAM2's video predictor propagates masks temporally from seed frames. Seeds are selected from annotated (GT) frames when available (up to 10 per label, spaced $\geq$10 frames apart, prioritized by detection score); otherwise, the first detection of each label is used. A single propagation pass produces masks for all seeded objects across all frames, binarized at threshold $0.5$.

\paragraph{Mask Union.}
\label{sup:sec:mask_union}
The final per-object mask for each frame is the pixel-wise union of image-mode and video-mode masks:
\begin{equation}
    \mathbf{M}_{\text{final}}^{(l,t)} = \mathbf{M}_{\text{image}}^{(l,t)} \;\lor\; \mathbf{M}_{\text{video}}^{(l,t)}
\end{equation}
where $l$ denotes the object label and $t$ the frame index, preserving both per-frame precision and temporal consistency.

\subsection{Floor Alignment via SMPL-Based Similarity Estimation}
\label{sup:sec:floor_alignment}

Before computing 3D bounding boxes, we establish a floor-aligned coordinate system by registering the \PiThree-reconstructed point cloud to a canonical frame where the floor plane is horizontal.

\paragraph{SMPL Mesh Correspondence.}
\label{sup:sec:smpl_corr}
We leverage estimated SMPL~\cite{loper2015smpl} body meshes as anchors for scale-and-pose recovery. For each frame, 2D body keypoints (from OpenPose~\cite{cao2017realtimemultiperson2dpose}) are paired with fitted SMPL 3D joint positions (in metric space), yielding $K$ correspondence pairs $\{(\mathbf{s}_k, \mathbf{d}_k)\}$ between SMPL points and \PiThree scene points. For each frame with $K \geq 3$ correspondences, a 7-DoF similarity transformation (scale $s$, rotation $\mathbf{R}$, translation $\boldsymbol{\tau}$) mapping scene space to metric space is estimated via RANSAC:
\begin{equation}
    \mathbf{s}_k \approx s \cdot \mathbf{R} \cdot \mathbf{d}_k + \boldsymbol{\tau}
\end{equation}

\paragraph{Global Floor Transform.}
\label{sup:sec:global_floor}
Per-frame similarity estimates are robustly averaged (weighted by inlier count) to produce a single global similarity $(s_{\text{avg}}, \mathbf{R}_{\text{avg}}, \boldsymbol{\tau}_{\text{avg}})$, requiring at least $\max(3, 20\%)$ valid frames. In the floor-aligned frame, the Y-axis is the floor normal (up) and the XZ plane is the floor surface. The floor-alignment transform maps world points to floor-aligned coordinates:
\begin{equation}
    \mathbf{p}_{\text{floor}} = \frac{1}{s_{\text{avg}}} \cdot \mathbf{R}_{\text{avg}} \cdot (\mathbf{p}_{\text{world}} - \boldsymbol{\tau}_{\text{avg}})
    \label{sup:eq:floor_align}
\end{equation}
and its inverse recovers world coordinates:
\begin{equation}
    \mathbf{p}_{\text{world}} = s_{\text{avg}} \cdot \mathbf{R}_{\text{avg}}^{\top} \cdot \mathbf{p}_{\text{floor}} + \boldsymbol{\tau}_{\text{avg}}
    \label{sup:eq:floor_to_world}
\end{equation}

\subsection{Oriented 3D Bounding Box Computation}
\label{sup:sec:bbox_computation}

For each interacting object in each frame, we extract masked 3D points from the \PiThree reconstruction and fit OBBs using two complementary methods: an unconstrained PCA-based OBB and a floor-parallel OBB.

\paragraph{Masked 3D Point Extraction.}
\label{sup:sec:point_extraction}
The 2D bounding box (GT or Grounding DINO) is resized to the \PiThree prediction resolution and intersected with the corresponding SAM2 mask (Section~\ref{sup:sec:segmentation}) to produce a tight object mask (falling back to the rectangular bbox if no mask exists). Points are then filtered by the per-pixel confidence map (thresholded at the frame's 5th-percentile, clamped $\geq 10^{-3}$) and by finite-coordinate validity.

\paragraph{Multiscale Erosion.}
\label{sup:sec:multiscale_erosion}
To strip boundary artifacts from imprecise masks, we erode the object mask with elliptical kernels of size $\{0, 3, 5, 7, 10\}$\,px, extract the surviving 3D points at each level (requiring $\geq$50 points), and select the erosion yielding the \textbf{minimum-volume} bounding box.
\label{sup:sec:candidate_volume}
Candidate volumes are computed from coordinate-wise extents in floor-aligned coordinates (Eq.~\eqref{sup:eq:floor_align}):
\begin{equation}
    V = \prod_{d \in \{x,y,z\}} (\max_k p_{k,d}^{\text{floor}} - \min_k p_{k,d}^{\text{floor}})
\end{equation}

Using the filtered 3D points from the best erosion scale, we compute two OBB variants.

\paragraph{PCA OBB.}
\label{sup:sec:obb_pca}
A full 3-DoF rotation OBB is obtained via PCA on the object point cloud: the covariance matrix $\mathbf{C} = \frac{1}{N-1}\sum_k (\mathbf{p}_k - \boldsymbol{\mu})(\mathbf{p}_k - \boldsymbol{\mu})^{\top}$ is eigen-decomposed as $\mathbf{C} = \mathbf{V} \boldsymbol{\Lambda} \mathbf{V}^{\top}$, points are projected onto the principal axes $\mathbf{q}_k = \mathbf{V}^{\top}(\mathbf{p}_k - \boldsymbol{\mu})$, and extents are taken as $e_d = \max_k q_{k,d} - \min_k q_{k,d}$. The OBB center is adjusted for asymmetric distributions:
\begin{equation}
    \mathbf{c}_{\text{obb}} = \boldsymbol{\mu} + \mathbf{V} \cdot \frac{\mathbf{q}_{\min} + \mathbf{q}_{\max}}{2}
\end{equation}
Eight corners are generated from $\pm e_d/2$ along principal axes and rotated back to world via $\mathbf{V}$.

\paragraph{Floor-Parallel OBB.}
\label{sup:sec:obb_floor}
A constrained OBB restricted to 1-DoF yaw rotation around the floor normal: points are transformed to the floor-aligned frame (Eq.~\eqref{sup:eq:floor_align}), projected onto the XZ plane, and enclosed by a minimum-area 2D rectangle. The rectangle is extruded vertically over $[y_{\min}, y_{\max}]$ to produce 8 corners, which are transformed back to world coordinates via Eq.~\eqref{sup:eq:floor_to_world}.

\subsection{Temporal Smoothing}
\label{sup:sec:temporal_smoothing}

Bounding boxes computed independently per frame exhibit temporal jitter. Per-frame minimum-volume erosion selection (Section~\ref{sup:sec:multiscale_erosion}) provides implicit scale normalization. For objects appearing across multiple frames, the 3D bounding box parameters (center, extents, yaw) are smoothed using a forward Kalman filter followed by Rauch-Tung-Striebel (RTS) backward smoothing, producing temporally consistent trajectories while preserving sharp transitions.

\subsection{World-to-Final Coordinate Transform}
\label{sup:sec:world_to_final}

For downstream consumption, all annotations are converted from the \PiThree world frame to a \textbf{final coordinate system} that is floor-aligned (XY = floor, Z = up), metrically scaled (meters, via SMPL similarity), and origin-centered. The world-to-final transform is:
\begin{equation}
    \mathbf{p}_{\text{final}} = \mathbf{A} \cdot (\mathbf{p}_{\text{world}} - \mathbf{o}_{\text{world}})
\end{equation}
where $\mathbf{A} = \frac{1}{s_{\text{avg}}} \mathbf{R}_{\text{avg}}$ and $\mathbf{o}_{\text{world}} = \boldsymbol{\tau}_{\text{avg}}$. Camera poses are similarly transformed:
\begin{equation}
    \mathbf{R}_{\text{final}} = \mathbf{A} \cdot \mathbf{R}_{\text{world}}, \quad
    \boldsymbol{\tau}_{\text{final}} = \mathbf{A} \cdot (\boldsymbol{\tau}_{\text{world}} - \mathbf{o}_{\text{world}})
\end{equation}
All annotations including OBB corners, point clouds, camera poses, and floor meshes; are converted to this final frame and stored per-video. Table~\ref{sup:tab:bbox_params} summarizes the key hyperparameters used in the pipeline.

\begin{table}[t]
\centering
\caption{3D bounding box pipeline parameters.}
\label{sup:tab:bbox_params}
\begin{tabular}{@{}lll@{}}
\toprule
\textbf{Parameter} & \textbf{Value} & \textbf{Description} \\
\midrule
\multicolumn{3}{@{}l}{\textit{Detection}} \\
Detector model         & Grounding DINO (base)   & Zero-shot open-vocabulary \\
Box threshold          & 0.25                     & Detection confidence \\
Text threshold         & 0.30                     & Text-box alignment \\
NMS IoU threshold      & 0.50                     & Per-class NMS \\
GT pseudo-score        & 1.001                    & Ensures GT survives NMS \\
\midrule
\multicolumn{3}{@{}l}{\textit{Segmentation}} \\
SAM2 model             & hiera-base-plus          & Balanced speed/quality \\
Mask threshold         & 0.50                     & Logit binarization \\
Max seeds per label    & 10                       & Video mode anchors \\
Anchor min gap         & 10 frames                & Seed spacing \\
Precision              & bfloat16                 & GPU autocast \\
\midrule
\multicolumn{3}{@{}l}{\textit{Floor Alignment}} \\
RANSAC iterations      & 500                      & Similarity estimation \\
RANSAC inlier thresh   & 0.03\,m                  & Correspondence quality \\
Scale bounds           & $[0.4, 3.0]$             & Physical plausibility \\
Min valid frames       & $\max(3, 20\%)$          & Quality gate \\
\midrule
\multicolumn{3}{@{}l}{\textit{3D Bounding Box}} \\
Erosion kernels        & $\{0, 3, 5, 7, 10\}$\,px & Multiscale mask refinement \\
Min points per scale   & 50                       & Candidate quality \\
Confidence threshold   & $\max(10^{-3}, P_5)$     & Adaptive per-frame \\
Selection criterion    & Minimum volume           & Among erosion candidates \\
\bottomrule
\end{tabular}
\end{table}

%% file: sup_tex_files/sup_semantic_annotation.tex
World Scene Graphs provide a holistic, temporally-grounded
representation of human-object relationships in video, extending
beyond single-frame annotations to capture the full spatio-temporal
dynamics of a scene.  We employ current
Vision-Language Models (VLMs) as a bridge to generate dense
\emph{pseudo semantic annotations} for
\emph{World Scene Graph Generation} (WSGG).  Specifically, we leverage
VLMs to predict human-object relationship
labels across all annotated frames of a video, following AG's annotation schema~\cite{ji2019actiongenomeactionscomposition}: (1) \textbf{Attention Relationships} (single-label): \texttt{looking\_at}, \texttt{not\_looking\_at}, \texttt{unsure}; (2) \textbf{Contacting Relationships} (multi-label): \texttt{carrying}, \texttt{covered\_by}, \texttt{drinking\_from}, \texttt{eating}, \texttt{have\_it\_on\_the\_back}, \texttt{holding}, \texttt{leaning\_on}, \texttt{lying\_on}, \\ \texttt{not\_contacting}, \texttt{other\_relationship}, \texttt{sitting\_on}, \texttt{standing\_on}, \texttt{touching}, \texttt{twisting}, \texttt{wearing}, \texttt{wiping}, \texttt{writing\_on}; (3) \textbf{Spatial Relationships} (multi-label): \texttt{above}, \texttt{beneath}, \texttt{in\_front\_of}, \texttt{behind}, \texttt{on\_the\_side\_of}, \texttt{in}.


\subsection{Models}
\label{sup:sec:models}

We employ three open-source state-of-the-art VLMs spanning different model
families and parameter counts, all deployed behind the vLLM inference
engine for efficient batched generation with tensor parallelism: (1) \textbf{Kimi-VL} (Kimi-VL-A3B-Instruct) - a lightweight mixture-of-experts VLM; (2) \textbf{InternVL 2.5} (InternVL2\_5-8B-MPO) - a mid-scale VLM with multi-modal preference optimisation; (3) \textbf{Qwen 2.5-VL} (Qwen2.5-VL-7B-Instruct) - an instruction-tuned VLM from the Qwen family. All models support variable-length multi-frame video input. A BGE embedding model (bge-large-en-v1.5) is
additionally used for text-similarity computations in the RAG-based annotation method.

\subsection{Generation Setup}
\label{sup:sec:setup}

\noindent\textbf{Video Input.} Video frames are loaded from disk and converted to a $T \times C
\times H \times W$ float tensor.  A pixel-budget constraint
(\texttt{total\_pixels}$= 128000$) is enforced to keep memory
consumption manageable.  We use annotation-driven per-frame clips. This provides fine-grained temporal context anchored to the frame being evaluated. The annotation-driven clips ensure that relationship predictions are grounded in the temporal neighbourhood of each
annotated frame.


\noindent\textbf{Relationship Query Prompt.} For each object $o \in \mathcal{O}_v$ on each annotated frame, a structured prompt elicits predictions for all three relationship
axes:

\begin{tcolorbox}[colback=gray!5, colframe=black!50, boxrule=0.5pt, arc=2pt, left=4pt, right=4pt, top=4pt, bottom=4pt]
\small
\textit{You are analyzing a video clip from a scene.  The object
``\{o\}'' is one of the objects present in this video.  It may or may
not be visible in the current frame.  A person IS visible in this
frame.}

\textit{Based on the full video context, predict the relationships
between the person and the object ``\{o\}''.}

\begin{enumerate}[nosep]
  \item ATTENTION: Pick EXACTLY ONE from: \{attention labels\}
  \item CONTACTING: Pick ONE OR MORE from: \{contacting labels\}
  \item SPATIAL: Pick ONE OR MORE from: \{spatial labels\}
\end{enumerate}
\textit{Respond ONLY in JSON format\ldots}
\end{tcolorbox}

\noindent\textbf{Discriminative Verification.} After the generative prediction step, each predicted relationship label is independently verified through a discriminative Yes/No
classification using the same VLM.  For a predicted
relationship $(o, r)$, the verification prompt varies by axis:

\begin{tcolorbox}[colback=gray!5, colframe=black!50, boxrule=0.5pt, arc=2pt, left=4pt, right=4pt, top=4pt, bottom=4pt]
  \small
  (1)~\textbf{Attention Relationship}: ``\textit{In this video, is the person \{r\} the \{o\}? Answer only Yes or No.}'';
  (2)~\textbf{Contacting Relationship}: ``\textit{In this video, is the person \{r\} the \{o\}? Answer only Yes or No.}'';
  (3)~\textbf{Spatial Relationship}: ``\textit{In this video, is the \{o\} \{r\} the person? Answer only Yes or No.}''
\end{tcolorbox}

All verification prompts across all frames and objects are collected
and dispatched in a single batched VLM call (chunked into sub-batches of 64), yielding a confidence score $p_\text{yes}$ for every predicted label.

\subsection{Annotation Methods}
\label{sup:sec:methods}

We generate pseudo-annotations using:

\paragraph{RAG Generation.}
\label{sup:sec:methods:rag}
The primary method uses Retrieval-Augmented Generation (RAG) over a precomputed scene graph from Phase~1.
Per-clip graphs are merged into a video-level directed graph, and a 4-step pipeline produces relationship predictions:
(1)~keyword extraction from query prompts via a batched VLM call;
(2)~cached node retrieval using BGE embeddings (cosine similarity $> 0.5$);
(3)~templatised node refinement via batched Yes/No verification;
(4)~frame-specific final answer using annotation-driven clips with graph-node context prepended to the prompt.
Queries are processed and deduplicated across frames, and all batched calls are chunked into sub-batches of 64.

%% file: sup_tex_files/sup_manual_correction_3dbbox.tex
\subsection{Correcting the Floor Transform}
\label{sup:sec:floor-transform}

Monocular 3D reconstruction methods such as DUSt3R~\cite{wang2024dust3rgeometric3dvision} produce an arbitrary world frame whose vertical axis does not coincide with gravity.
We align the reconstructed scene into a \emph{gravity-aligned reference frame} ($XY$-plane $=$ floor, $+Z$ $=$ up) via three stages: (i)~\emph{automatic floor alignment} from the reconstruction's floor mesh and global similarity transform, (ii)~\emph{manual interactive correction} with 6-DoF controls, and (iii)~\emph{automated XY-plane alignment} that analytically maps the corrected floor to $Z{=}0$.

\paragraph{Automatic Floor Alignment via Global Floor Similarity.}
\label{sup:sec:floor-auto-alignment}
The reconstruction pipeline extracts a floor mesh $\mathcal{F} = (\mathbf{V}_0, \mathbf{F}_0, \mathbf{C}_0)$ with vertex positions $\mathbf{V}_0 \in \mathbb{R}^{N_v \times 3}$, faces $\mathbf{F}_0$, and optional colors $\mathbf{C}_0$, together with a global floor similarity $(s_g, \mathbf{R}_g, \boldsymbol{\tau}_g)$. Floor vertices are transformed into the world frame as:
\begin{equation}
    \mathbf{v}_{\text{world}} = s_g \cdot (\mathbf{v}_0 \cdot \mathbf{R}_g^{\top}) + \boldsymbol{\tau}_g, \quad \forall\, \mathbf{v}_0 \in \mathbf{V}_0.
    \label{sup:eq:floor-world-transform}
\end{equation}

From the columns of $\mathbf{R}_g$ we extract in-plane tangent directions $\boldsymbol{\ell}_1 = \mathbf{R}_g[:, 0]$, $\boldsymbol{\ell}_2 = \mathbf{R}_g[:, 2]$ and floor normal $\mathbf{n} = \mathbf{R}_g[:, 1]$, forming the basis $\mathbf{F} = [\boldsymbol{\ell}_1 \;|\; \boldsymbol{\ell}_2 \;|\; \mathbf{n}]$. The world-to-floor rotation is $\mathbf{R}_{\text{align}} = \mathbf{F}^{\top}$ with translation $\boldsymbol{\tau}_{\text{align}} = -\mathbf{R}_{\text{align}} \cdot \boldsymbol{\tau}_g$:
\begin{equation}
    \mathbf{x}_{\text{floor}} = \mathbf{R}_{\text{align}} \cdot \mathbf{x}_{\text{world}} + \boldsymbol{\tau}_{\text{align}}.
    \label{sup:eq:world-to-floor}
\end{equation}
An optional $ZY$-plane mirror $\mathbf{M} = \text{diag}(-1, 1, 1)$ yields the final transform:
\begin{equation}
    \mathbf{R}_{\text{final}} = \mathbf{M} \cdot \mathbf{R}_{\text{align}}, \quad
    \boldsymbol{\tau}_{\text{final}} = \mathbf{M} \cdot \boldsymbol{\tau}_{\text{align}},
    \label{sup:eq:world-to-final}
\end{equation}
so that $\mathbf{x}_{\text{final}} = \mathbf{R}_{\text{final}} \cdot \mathbf{x}_{\text{world}} + \boldsymbol{\tau}_{\text{final}}$.

\paragraph{Manual Interactive Floor Correction.}
\label{sup:sec:floor-manual-correction}
The automatic alignment is often imperfect. An interactive 3D tool renders the floor mesh, point cloud, and bounding boxes, allowing the annotator to adjust a correction transform $\mathbf{T}_{\delta} = (\mathbf{R}_{\delta}, \boldsymbol{\tau}_{\delta}, \mathbf{s}_{\delta})$ via 6-DoF controls. Both original (wireframe) and corrected (solid) floor meshes are displayed simultaneously for visual feedback. The correction is persisted to Firebase per video; on reload it is applied automatically.

\paragraph{Automated XY-Plane Alignment.}
\label{sup:sec:xy-alignment}
After manual correction the floor may still not lie in the canonical $XY$-plane. We compute $\mathbf{T}_{\text{XY}} = (\mathbf{R}_{\text{XY}}, \boldsymbol{\tau}_{\text{XY}})$ to map the corrected floor to $Z{=}0$ in six steps:

\paragraph{Step 1: Apply Floor Correction.}
The floor vertices $\mathbf{V}_0$ are first transformed by the correction $\mathbf{T}_{\delta}$ (Sec.~\ref{sup:sec:floor-manual-correction}):
\begin{equation}
    \mathbf{v}_{\text{corr}} = \mathbf{R}_{\delta} \cdot (\mathbf{s}_{\delta} \odot \mathbf{v}_0) + \boldsymbol{\tau}_{\delta}, \quad \forall\, \mathbf{v}_0 \in \mathbf{V}_0.
\end{equation}

\paragraph{Step 2: Compute Floor Normal.}
The floor centroid $\bar{\mathbf{v}}$ and robust normal $\hat{\mathbf{n}}$ are estimated from triangle cross products (selecting the largest-magnitude result); the normal is flipped if needed so that $\hat{n}_z > 0$.

\paragraph{Step 3: Normal Alignment via Rodrigues' Rotation.}
The rotation $\mathbf{R}_{\text{norm}}$ aligning $\hat{\mathbf{n}}$ with $+Z$ is obtained via Rodrigues' formula with axis $\hat{\mathbf{k}} = \frac{\hat{\mathbf{n}} \times \hat{\mathbf{z}}}{\|\hat{\mathbf{n}} \times \hat{\mathbf{z}}\|}$ and angle $\phi = \arccos(\hat{\mathbf{n}} \cdot \hat{\mathbf{z}})$:
\begin{equation}
    \mathbf{R}_{\text{norm}} = \cos\phi \cdot \mathbf{I} + \sin\phi \cdot [\hat{\mathbf{k}}]_{\times} + (1 - \cos\phi) \cdot \hat{\mathbf{k}} \hat{\mathbf{k}}^{\top}.
\end{equation}

\paragraph{Step 4: In-Plane Rotation Correction.}
The residual in-plane misalignment is corrected by a $Z$-axis rotation $\mathbf{R}_z(\alpha)$ where $\alpha = -\text{atan2}(\tilde{x}_y, \tilde{x}_x)$ and $\tilde{\mathbf{x}} = \mathbf{R}_{\text{norm}} \cdot \mathbf{R}_{\delta} \cdot [1,0,0]^{\top}$. The combined rotation is $\mathbf{R}_{\text{XY}} = \mathbf{R}_z(\alpha) \cdot \mathbf{R}_{\text{norm}}$.

\paragraph{Step 5: Translation to Origin.}
The floor--origin intersection $\mathbf{p}_{\cap} = (\hat{\mathbf{n}} \cdot \bar{\mathbf{v}}) \cdot \hat{\mathbf{n}}$ is mapped to the origin: $\boldsymbol{\tau}_{\text{XY}} = -\mathbf{R}_{\text{XY}} \cdot \mathbf{p}_{\cap}$.

\paragraph{Step 6: Euler Angle Extraction.}
$\mathbf{R}_{\text{XY}}$ is decomposed into Euler angles $(\theta_x, \theta_y, \theta_z)$ ($ZYX$ convention) for storage.

\paragraph{Bounding Box Refitting.}
After applying $\mathbf{T}_{\text{XY}}$, each OBB is refitted: inlier points (identified via the separating axis test) are transformed, and a 2D PCA-based OBB is estimated on the projected $XY$ coordinates. The final 3D OBB combines the PCA box extents with the $Z$-range of the transformed inliers.

\paragraph{Final Alignment Composition.}
\label{sup:sec:final-alignment}
The three stages compose into a single rigid transformation:
\begin{equation}
    \mathbf{x}_{\text{canonical}} = \mathbf{T}_{\text{XY}} \circ \mathbf{T}_{\delta} \circ \mathbf{T}_{\text{auto}}(\mathbf{x}_{\text{world}}),
\end{equation}
where $\mathbf{T}_{\text{auto}}$ is from Eq.~\ref{sup:eq:world-to-final}, $\mathbf{T}_{\delta}$ is the manual correction, and $\mathbf{T}_{\text{XY}}$ is the XY-plane alignment. The composed transform is cached per video.

\begin{figure}[H]
    \centering
    \includegraphics[width=\textwidth]{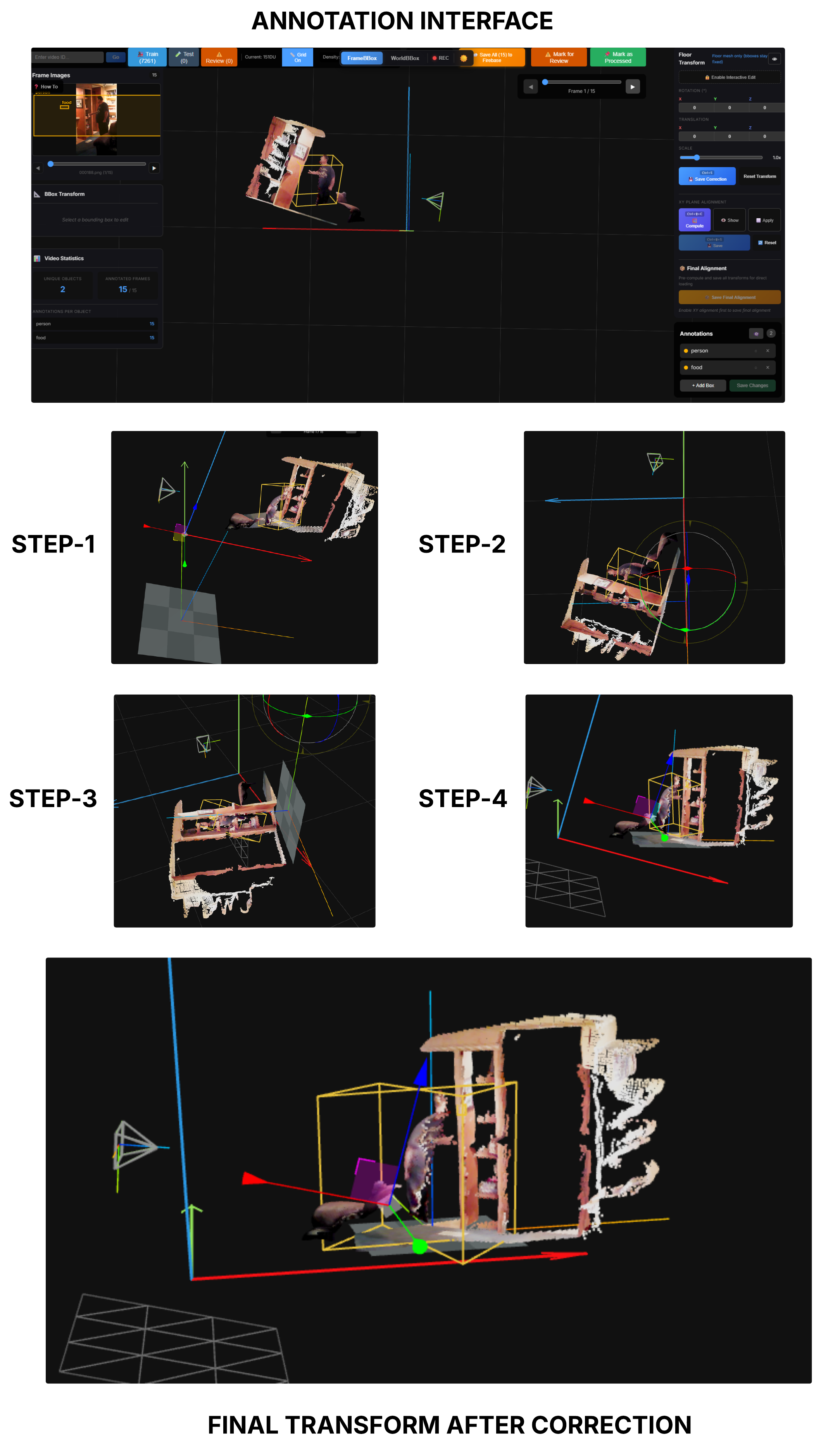}
    \caption{Overview of the annotation interface used for manual floor correction. The figure illustrates the sequential adjustment process across four steps, culminating in the final corrected transform for accurate 3D scene alignment. }
    \label{sup:fig:floor_correction_interface}
\end{figure}

\subsection{Correcting World-Level 3D BBox Annotations}
\label{sup:sec:worldbbox-correction}

With the gravity-aligned frame established (Sec.~\ref{sup:sec:floor-transform}), 3D bounding boxes are lifted into world coordinates. The automatic annotations often exhibit errors (incorrect dimensions, missing objects, misaligned orientations, wrong labels), so we provide the \emph{WorldBBox Viewer} for interactive 3D correction.

\paragraph{Automatic World-Level BBox Generation.}
\label{sup:sec:auto-world4d}
Initial world-level 4D annotations are generated automatically. Objects are classified as \emph{static} (e.g., \texttt{table}, \texttt{sofa}) or \emph{dynamic} (e.g., \texttt{person}); for static objects, missing frames are filled from the nearest known frame (object permanence). Each per-frame box ($\mathbf{C}_{\text{world}} \in \mathbb{R}^{8 \times 3}$) is transformed to the gravity-aligned frame:
\begin{equation}
    \mathbf{C}_{\text{final}} = (\mathbf{R}_{\text{final}} \cdot \mathbf{C}_{\text{world}}^{\top})^{\top} + \boldsymbol{\tau}_{\text{final}}^{\top}.
\end{equation}
For each static label, a union bounding box is computed across all frames:
\begin{equation}
    \mathbf{C}_{\text{union}} = \text{OBB}\!\left(\bigcup_{f \in \text{frames}} \mathbf{C}_{\text{final}}^{(\ell, f)}\right),
\end{equation}
ensuring spatial consistency while preserving the original world-space boxes.

\paragraph{Interactive Correction Interface}
\label{sup:sec:worldbbox-viewer}

The WorldBBox Viewer provides a 3D canvas (via Three.js/React Three Fiber) rendering the floor mesh, reconstructed point cloud, bounding box wireframes with floating text labels, and relationship arcs connecting subject-predicate-object triples.  The viewer supports orbital camera controls and frame-by-frame navigation. For each selected bounding box, the annotator can apply translate, rotate, and scale transforms via sliders with live preview.  Transforms are applied sequentially (scale $\to$ rotate $\to$ translate about box center) and committed on confirmation, with an undo stack for reverting.  The annotations panel additionally supports inline label editing, per-box visibility toggles, and adding/deleting boxes.

\paragraph{Temporal Propagation of Annotations}
\label{sup:sec:propagation}

A key challenge in world-level annotation is ensuring that object bounding boxes are consistent across all frames. The Propagation Panel provides tools to copy a bounding box annotation from the current frame to other frames: The annotator selects a bounding box and specifies: (1) \textbf{Direction}: Forward (to subsequent frames), backward (to preceding frames), or both. (2) \textbf{Frame Count}: A specific number of frames to propagate to, or ``All'' to propagate to the video's start/end. Upon clicking \textbf{Propagate}, the selected box's label, corners, center are duplicated to each target frame.

%% file: sup_tex_files/sup_manual_correction_wsg.tex
\subsection{Overview}
\label{sup:sec:manual_correction_pseudo}

The pseudo-annotations produced by the MLLM pipeline contain errors from
hallucination, ambiguous visual context, or incorrect grounding.  To produce
gold-standard annotations, we deploy a \emph{human-in-the-loop} manual
correction stage using a purpose-built web annotation tool.  Trained annotators
systematically review and correct every pseudo-annotated relationship across
three semantic axes: \textbf{Attention}, \textbf{Contacting}, and
\textbf{Spatial}.  The correction is designed as a \emph{refinement} workflow:
annotators start from MLLM predictions rather than labeling from scratch,
significantly reducing annotation time.

\subsection{Annotation Interface}
\label{sup:sec:correction:interface}

\begin{figure}[!ht]
    \centering
    \includegraphics[width=\textwidth]{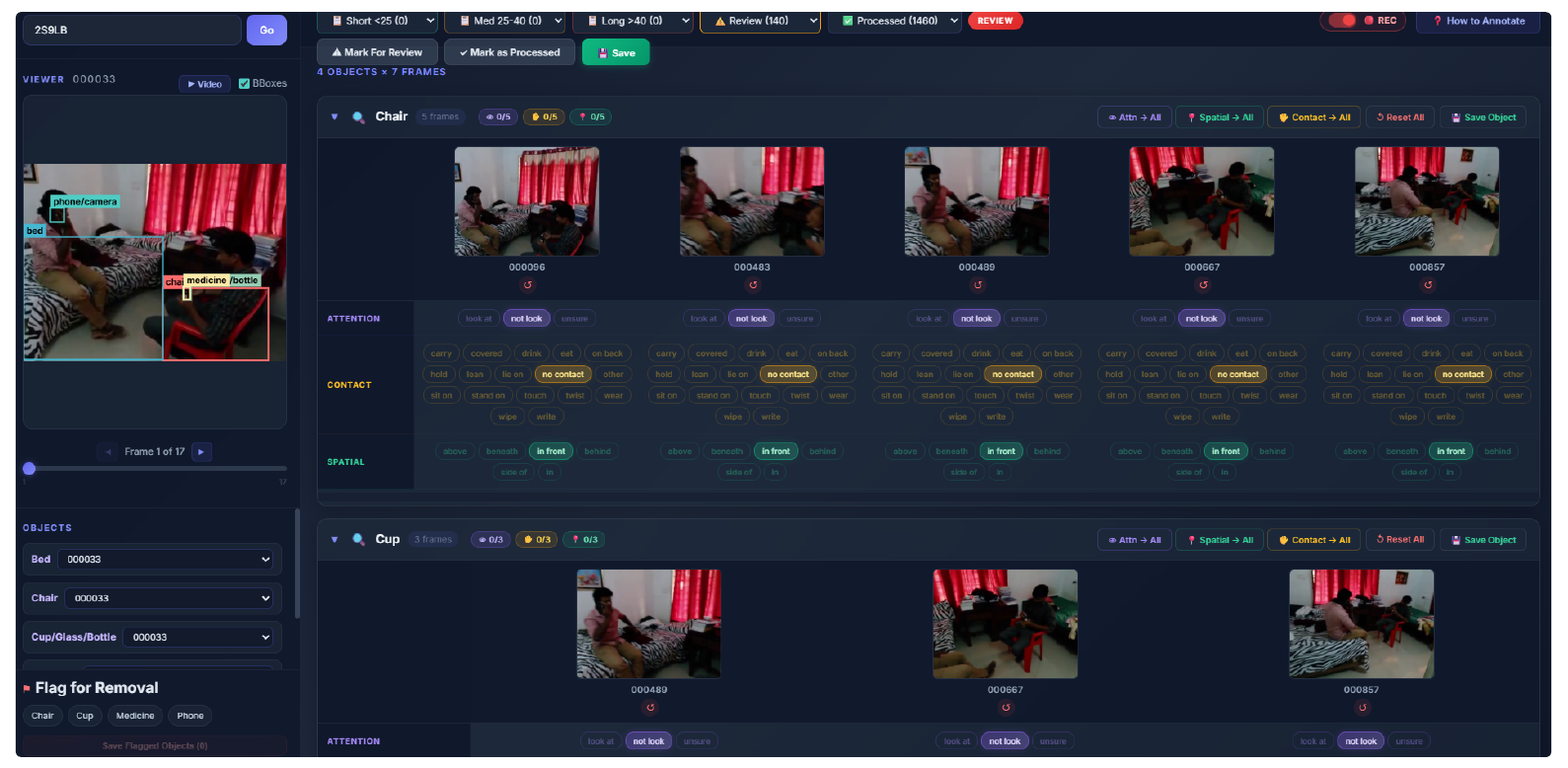}
    \caption{Interface for manual relationship correction in WorldSGG. The tool presents object-centric frame sequences and allows annotators to refine attention, contact, and spatial predicates across time for accurate scene graph construction. }
    \label{sup:fig:rel_correction_interface}
\end{figure}

The tool follows a two-panel layout. The \emph{left panel} provides a
high-fidelity frame viewer with:
\begin{itemize}[nosep]
  \item Full-resolution frame display with a slider for scrubbing through
        all video frames.
  \item An HTML5 canvas overlay rendering person bounding boxes (green) and
        object bounding boxes (15 distinct colors) with class-name labels.
  \item A lightbox modal for pixel-level inspection with zoom
        (0.5--5$\times$) and pan.
\end{itemize}

The \emph{right panel} serves as the annotation workspace:
\begin{itemize}[nosep]
  \item \textbf{Video selection}: three dropdown selectors group videos by
        status (Pending / Review / Processed).
  \item \textbf{Object accordions}: each missing object is presented as a
        collapsible section showing the object name, frame count, and
        per-category correction statistics.
  \item \textbf{Annotation grid}: a table of frames $\times$ relationship
        categories.  Each cell contains toggle buttons for every possible
        label.  Attention is single-select; Contacting and Spatial are
        multi-select.  Active labels are color-coded with abbreviated
        display names.
\end{itemize}

\subsection{Correction Workflow}
\label{sup:sec:correction:workflow}

\paragraph{Per-Video Workflow.}
The annotator (1)~selects a pending video, (2)~inspects frames using the
frame viewer and bounding-box overlay, (3)~corrects each missing object's
relationships by toggling labels in the annotation grid, (4)~saves
corrections, and (5)~marks the video as Processed or flags it for Review.

\paragraph{Propagation Tools.}
To reduce repetitive annotation when relationships are stable across frames: 
(1) Copy Forward ($\rightarrow$): copies one frame's complete 
annotation to all subsequent frames for the same object. 
(2) Apply to All: broadcasts the first frame's annotations to all frames.

\subsection{Quality Tracking}
\label{sup:sec:correction:quality}

The interface displays real-time correction statistics to monitor progress.
For each missing object~$o$ across its $N$~frame appearances, the system
computes per-category correction counts by comparing compiled (post-edit)
predictions against base (MLLM) predictions:
\begin{equation}
  n_\text{cat}^{o} = \sum_{f=1}^{N} \mathbb{1}\big[
    F_{\text{comp}}^{(f,o)}.\texttt{cat} \neq
    F_{\text{base}}^{(f,o)}.\texttt{cat}
  \big],
  \quad \text{cat} \in \{\text{att}, \text{con}, \text{spa}\}.
\end{equation}
These counts are displayed as colored badges ($n/N$) in each object
accordion header, providing at-a-glance visibility into annotator
modifications.

\subsection{Status Management}
\label{sup:sec:correction:status}

Videos follow a three-state lifecycle: \textbf{Pending} (uploaded, not yet
reviewed), \textbf{Processed} (corrections complete), and \textbf{Review}
(flagged for a second pass due to ambiguity).  Status transitions are
triggered explicitly by annotators and stored with timestamps in Firebase,
enabling dataset-level progress tracking via the dropdown counters in the
interface header.

%% file: sup_tex_files/sup_ag4d_statistics.tex

ActionGenome4D (AG4D) extends the original Action
Genome~\cite{ji2019actiongenomeactionscomposition} dataset into a 4D
spatio-temporal representation by augmenting every video with (i)~per-frame
3D scene reconstructions and camera poses, (ii)~world-frame oriented 3D
bounding boxes (OBBs) for all annotated objects, and (iii)~dense semantic
relationship annotations covering both \emph{observed} and
\emph{unobserved} objects at every annotated timestamp.
This section summarises the key statistics of the resulting dataset and
contrasts them with the original Action Genome annotations.

\vspace{2mm}
\noindent\textbf{Dataset Overview.}\label{sup:sec:ag4d:overview} Action Genome 
provides frame-level 2D bounding-box annotations and
human--object relationship labels for videos sourced from the Charades
dataset~\cite{sigurdsson2016hollywoodhomescrowdsourcingdata}.  AG4D preserves the original
train/test split and enriches every video along three axes:

\begin{enumerate}[nosep]
  \item \textbf{Geometric:} Per-frame 3D point clouds reconstructed via
        $\pi^3$~\cite{wang2025pi}, floor-aligned via SMPL-based similarity
        estimation, and enclosed by oriented 3D bounding boxes (OBBs) in a
        shared world coordinate frame
        (Section~\ref{sup:sec:bbox_pipeline}).
  \item \textbf{Semantic:} Relationship annotations expanded from
        observed-only objects ($\mathcal{O}^{t}$) to the full world state
        ($\mathcal{W}^{t} = \mathcal{O}^{t} \cup \mathcal{U}^{t}$),
        covering attention, spatial, and contacting predicates for every
        (person, object) pair at every annotated frame
        (Section~\ref{sup:sec:models}).
  \item \textbf{Camera:} Camera-to-world SE(3) poses
        $\{\mathbf{T}^{t}\}_{t=1}^{T}$ refined via iterative bundle
        adjustment, providing ego-motion trajectories for each video
        (Section~\ref{sup:sec:scene_construction}).
\end{enumerate}

\vspace{2mm}
\noindent\textbf{Annotation Strategy.}\label{sup:sec:ag4d:annotation_strategy} 
The semantic annotation strategy differs between the training and test
splits to balance scalability with annotation quality:

\begin{itemize}[nosep]
  \item \textbf{Training set - Pseudo-annotations:} Relationship labels
        for unobserved objects are generated automatically using a
        VLM-based pipeline (Section~\ref{sup:sec:models}).  A
        Graph-RAG approach retrieves relevant spatio-temporal context from
        a precomputed coarse event graph, and a discriminative verification
        step assigns per-label confidence scores.  These pseudo-labels
        serve as training targets, weighted by
        $\lambda_{\text{vlm}}$ in the loss function to account for
        potential noise.
  \item \textbf{Test set - Manual corrections:} Every pseudo-annotated
        relationship in the test split is reviewed and corrected by trained
        human annotators using a purpose-built web-based correction
        interface (Section~\ref{sup:sec:manual_correction_pseudo}).
        The correction workflow starts from the VLM predictions and
        refines them via toggle-based editing, copy-forward propagation,
        and temporal consistency enforcement.  This produces
        gold-standard annotations for reliable evaluation.
\end{itemize}

\noindent The geometric annotations (3D OBBs and camera poses) are
generated using the same automated pipeline for both splits, followed by
manual verification and correction of 3D bounding boxes via
3D annotation tool (Section~\ref{sup:sec:floor-manual-correction}).

\vspace{2mm}
\noindent\textbf{Comparison with Action Genome.}\label{sup:sec:ag4d:comparison} Table~\ref{sup:tab:ag_vs_ag4d} 
contrasts the AG dataset with ActionGenome4D across key dimensions.

\begin{table}[t]
\centering
\caption{\textbf{AG vs.\ ActionGenome4D.}
AG4D extends AG along geometric and semantic dimensions.
$\mathcal{O}^{t}$: observed objects; $\mathcal{U}^{t}$: unobserved
objects; $\mathcal{W}^{t}$: full world state.}
\label{sup:tab:ag_vs_ag4d}
\small
\begin{tabular}{@{}lcc@{}}
\toprule
\textbf{Property} & \textbf{Action Genome} & \textbf{ActionGenome4D} \\
\midrule
\multicolumn{3}{@{}l}{\textit{Scope}} \\
Object scope per frame    & $\mathcal{O}^{t}$ only   & $\mathcal{W}^{t} = \mathcal{O}^{t} \cup \mathcal{U}^{t}$ \\
Relationship coverage     & Observed pairs           & All world-state pairs \\
Coordinate frame          & 2D image                 & 3D world \\
Object localization       & 2D BBox                  & 2D BBox + 3D OBB \\
Camera poses              & \xmark                   & \cmark{} (SE(3) per frame) \\
3D scene reconstruction   & \xmark                   & \cmark{} (per-frame point clouds) \\
\midrule
\multicolumn{3}{@{}l}{\textit{Annotations}} \\
Predicate axes            & Attention, Spatial, Contacting & Attention, Spatial, Contacting \\
Attention labels          & 3 (single-label)         & 3 (single-label) \\
Spatial labels            & 6 (multi-label)          & 6 (multi-label) \\
Contacting labels         & 17 (multi-label)         & 17 (multi-label) \\
Object categories         & 36 (+~person)            & 36 (+~person) \\
Unobserved object labels  & \xmark                   & \cmark{} \\
\midrule
\multicolumn{3}{@{}l}{\textit{Scale}} \\
Videos                    & 9,250          & 9,250 \\
Train / Test videos       & 7,516 / 1,734   & 7,516 / 1,734 \\
Annotated frames          & 232,103         & 232,103 \\
\bottomrule
\end{tabular}
\end{table}

\vspace{2mm}
\noindent\textbf{Predicate Vocabulary.}\label{sup:sec:ag4d:predicates} AG4D inherits 
the Action Genome predicate vocabulary, organised into three disjoint axes:

\begin{enumerate}[nosep]
  \item \textbf{Attention} ($|\mathcal{P}_{\text{att}}| = 3$,
        single-label): \texttt{looking\_at},
        \texttt{not\_looking\_at}, \texttt{unsure}.
  \item \textbf{Spatial} ($|\mathcal{P}_{\text{spa}}| = 6$,
        multi-label): \texttt{above}, \texttt{beneath},
        \texttt{in\_front\_of}, \texttt{behind},
        \texttt{on\_the\_side\_of}, \texttt{in}.
  \item \textbf{Contacting} ($|\mathcal{P}_{\text{con}}| = 17$,
        multi-label): \texttt{carrying}, \texttt{covered\_by}, \\
        \texttt{drinking\_from}, \texttt{eating},
        \texttt{have\_it\_on\_the\_back}, \texttt{holding},
        \texttt{leaning\_on}, \\ \texttt{lying\_on}, 
        \texttt{not\_contacting}, \texttt{other\_relationship},
        \texttt{sitting\_on}, \texttt{standing\_on}, \\
        \texttt{touching}, \texttt{twisting}, \texttt{wearing},
        \texttt{wiping}, \texttt{writing\_on}.
\end{enumerate}

\noindent Attention is a single-label classification (exactly one label per
object pair per frame), while spatial and contacting are multi-label
(an object pair may simultaneously exhibit multiple spatial or contacting
relationships, e.g., a person can be both \texttt{holding} and
\texttt{touching} an object).

\vspace{2mm}
\noindent\textbf{Object Categories.}\label{sup:sec:ag4d:objects} The object vocabulary 
consists of 36 categories (excluding \texttt{person})
inherited from Action Genome.  These span common indoor objects such as
furniture (\texttt{chair}, \texttt{table}, \texttt{sofa}, \texttt{bed}),
personal items (\texttt{phone/camera}, \texttt{bag}, \texttt{book},
\texttt{laptop}), kitchenware (\texttt{cup/glass/bottle}, \texttt{dish},
\texttt{sandwich}), and room elements (\texttt{door}, \texttt{doorway},
\texttt{window}, \texttt{mirror}, \texttt{television}).
Including \texttt{person}, the full category set comprises 37 classes used
for object classification in SGDet.

\vspace{2mm}
\noindent\textbf{Geometric Annotation Statistics.}\label{sup:sec:ag4d:geometric_stats}
The 3D geometric annotation pipeline
(Section~\ref{sup:sec:auto-world4d}) produces world-frame oriented 3D
bounding boxes for every annotated object across all videos.  Key
statistics of the geometric annotations are:

\begin{itemize}[nosep]
  \item \textbf{Total 3D OBBs:} 828,213 OBBs across all videos and frames.
  \item \textbf{OBB fitting:} Each OBB is selected as the minimum-volume candidate 
  from a multiscale erosion sweep ($\{0, 3, 5, 7, 10\}$\,px kernels), with 
  temporal smoothing via Kalman filtering and RTS backward pass.
  \item \textbf{Coordinate system:} All OBBs are expressed in the final coordinate 
  system (floor-aligned, metrically scaled in meters, world-frame centered), alongside 
  per-frame camera poses.
\end{itemize}

\vspace{2mm}
\noindent\textbf{Semantic Annotation Statistics.}\label{sup:sec:ag4d:semantic_stats}
AG4D extends the original AG relationship annotations from
observed-only objects to the full world state.  The key distinction is
that every (person, object) pair receives relationship labels at every
annotated timestamp, regardless of objects' visibility.

\begin{itemize}[nosep]
  \item \textbf{Total relationship instances:} 602,668
        relationship triplets across all frames and splits (compared to
        518,895 in the original AG, which covers observed
        objects only).
  \item \textbf{Observed pairs:} 518,895 relationship
        instances where both the person and object are visible in the
        frame ($w_k^{t} \in \mathcal{O}^{t}$).
  \item \textbf{Unobserved pairs:} 83,773 relationship
        instances involving at least one unobserved object
        ($w_k^{t} \in \mathcal{U}^{t}$) (new in ActionGenome4D).
  \item \textbf{Avg.\ objects per frame (world state):}
        3.6 objects per frame in
        $\mathcal{W}^{t}$, compared to 3.2 observed objects
        per frame in the original AG.
  \item \textbf{Avg.\ relationships per frame:} 2.6
        (person, object) relationship instances per frame in
        $\mathcal{G}_{\mathcal{W}}^{t}$, compared to 2.2 in
        AG's observed-only $\mathcal{G}^{t}$.
\end{itemize}

\vspace{2mm}
\noindent\textbf{Train/Test Split Statistics.}\label{sup:sec:ag4d:splits}
Table~\ref{sup:tab:split_stats} summarises the per-split statistics.

\begin{table}[t]
\centering
\caption{\textbf{ActionGenome4D statistics.}
The train set uses VLM pseudo-annotations for unobserved object
relationships; the test set uses manually corrected annotations.}
\label{sup:tab:split_stats}
\small
\begin{tabular}{@{}lcc@{}}
\toprule
\textbf{Statistic} & \textbf{Train} & \textbf{Test} \\
\midrule
Videos                        & 7,516       & 1,734   \\
Annotated frames              & 175,751     & 56,352   \\
Object instances (observed)   & 549,315     & 201,683   \\
Object instances (unobserved) & 61,149     & 22,624   \\
Object instances (total)      & 610,464     & 224,307   \\
Relationship triplets (observed)   & 373,564 & 145,331 \\
Relationship triplets (unobserved) & 61,149 & 22,624 \\
Relationship triplets (total)      & 434,713 & 167,955 \\
3D OBBs                       & 604,128     & 224,085   \\
\midrule
Annotation type (unobserved)  & VLM pseudo-labels    & Manual corrections \\
\bottomrule
\end{tabular}
\vspace{-4mm}
\end{table}

\vspace{2mm}
\noindent\textbf{Predicate Distribution.}\label{sup:sec:ag4d:predicate_dist} The predicate 
label distributions in AG4D exhibit long-tail characteristics 
consistent with real-world relationship frequencies.
For the \textbf{attention} axis, \texttt{not\_looking\_at} dominates
(especially for unobserved objects, where the person is typically not
attending to off-screen objects), followed by \texttt{looking\_at} and
\texttt{unsure}. For the \textbf{contacting} axis, \texttt{not\_contacting} is the most
frequent label (again driven by unobserved objects with no physical
contact), while labels such as \texttt{holding}, \texttt{touching}, and
\texttt{sitting\_on} cover the active interactions. For the \textbf{spatial} axis, 
\texttt{in\_front\_of} and \texttt{on\_the\_side\_of} are common due to typical indoor scene
layouts, while \texttt{beneath} and \texttt{in} are relatively rare. The introduction 
of unobserved object annotations significantly amplifies
the representation of \texttt{not\_looking\_at} and
\texttt{not\_contacting}, since most unobserved objects are neither
attended to nor in physical contact with the person.  This shifts the
overall predicate distribution compared to the original AG and
motivates the use of label smoothing and weighted losses during
training.

%% file: sup_tex_files/sup_monocular_3d.tex
The monocular 3D object detector provides the geometric scaffolding for WorldSGG.  Given a single RGB frame, it simultaneously produces 2D bounding boxes with class labels and 8-corner oriented 3D bounding boxes (OBBs) in camera coordinates.  The 3D boxes are then lifted into the persistent world frame via extrinsic camera matrices, providing the structural prior consumed by all three WorldSGG methods (PWG, 4DST, MWAE). The system follows a Faster R-CNN~\cite{ren2016fasterrcnnrealtimeobject} architecture with three key innovations:
(i)~a \textbf{frozen DINOv2/v3 ViT backbone}~\cite{oquab2024dinov2learningrobustvisual} paired with a learnable Simple Feature Pyramid Network (SimpleFPN)~\cite{li2022exploringplainvisiontransformer};
(ii)~a \textbf{factorized 3D prediction head} decomposing 3D box regression into dimensions, rotation, depth, and center offset, reconstructed via pinhole back-projection;
(iii)~an \textbf{OVMono3D-style disentangled 3D loss}~\cite{zhang2023ovmono3dmonocular3dobject} with per-sample uncertainty weighting and Chamfer supervision.

\subsection{Model Architecture}
\label{sec:mono3d_architecture}

The full pipeline is: Image $\xrightarrow{\text{Transform}}$ Backbone $\xrightarrow{\text{FPN}}$ RPN $\xrightarrow{\text{ROI Heads}}$ 2D Detections + 3D Boxes.

\vspace{2mm}
\noindent\textbf{Backbone: Frozen DINOv2/v3 ViT.} The backbone is a Vision Transformer (ViT)~\cite{dosovitskiy2021imageworth16x16} pretrained with DINOv2/v3~\cite{oquab2024dinov2learningrobustvisual}, kept entirely \textbf{frozen} during training. Four variants are supported (Table~\ref{tab:backbone_variants}). Only the last $N_{\text{patch}} = H_p \times W_p$ spatial tokens are kept (stripping CLS and register tokens), reshaped to $\mat{F} \in \mR^{B \times C \times H_p \times W_p}$.

\begin{table}[!t]
\centering
\caption{Supported backbone variants.}
\label{tab:backbone_variants}
\small
\begin{tabular}{@{}lllrl@{}}
\toprule
\textbf{Key} & \textbf{Model} & \textbf{Hidden dim} & \textbf{Params} & \textbf{Patch size} \\
\midrule
\texttt{v2}  & DINOv2 ViT-B & 768  &  86M & 14 \\
\texttt{v2s} & DINOv2 ViT-S & 384  &  22M & 14 \\
\texttt{v2l} & DINOv2 ViT-L & 1024 & 304M & 14 \\
\texttt{v3l} & DINOv3 ViT-L & 1024 & 304M & 16 \\
\bottomrule
\end{tabular}
\end{table}

\vspace{2mm}
\noindent\textbf{Simple Feature Pyramid Network (SimpleFPN).} Following ViTDet~\cite{li2022exploringplainvisiontransformer}, the SimpleFPN converts the single-scale backbone output into a multi-scale pyramid at strides $\{P/4, P/2, P, 2P, 4P\}$ via transposed convolutions (upsampling) and max-pooling (downsampling), where $P$ is the patch size.  Each level is projected to 256 channels by a $1{\times}1$ convolution, BatchNorm, ReLU, and a $3{\times}3$ refinement convolution.  All FPN parameters are learnable.

\vspace{2mm}
\noindent\textbf{Region Proposal Network and ROI Heads.} The RPN uses standard Faster R-CNN anchors of sizes $\{32, 64, 128, 256, 512\}$ at pyramid levels $\{p2, \ldots, p6\}$ with aspect ratios $(0.5, 1.0, 2.0)$, contributing objectness and box-regression losses.  Proposals are processed by MultiScaleRoIAlign ($7{\times}7$ output, sampling ratio 2) feeding two FC layers that produce a shared 1024-d feature, from which parallel heads predict class logits and box deltas.

\vspace{2mm}
\noindent\textbf{Factorized 3D Prediction Head.} The 3D head is a lightweight branch sharing the 1024-d ROI features.  It concatenates the shared features with normalized 2D box coordinates and camera intrinsics, passed through a shared context FC (ReLU, 512-d output).  Five parallel heads produce:
\begin{align}
    \hat{\mat{d}} &= \op{softplus}(\cdot) + \epsilon \in \mR^{3}, & &\text{(dimensions: $l, w, h$)} \label{eq:dims} \\
    [\hat{s}, \hat{c}] &= \text{L2-normalize}(\cdot) \in \mR^{2}, & &\text{(rotation: $\sin\theta, \cos\theta$)} \label{eq:rot} \\
    \hat{z}_c &= \op{softplus}(\cdot) + \epsilon \in \mR^{1}, & &\text{(depth)} \label{eq:depth} \\
    \hat{\mat{\delta}} &\in \mR^{2}, & &\text{(2D center offset)} \label{eq:offset} \\
    \hat{\mu} &\in \mR^{1}, & &\text{(uncertainty)} \label{eq:mu}
\end{align}
where $\epsilon = 10^{-4}$ ensures strictly positive outputs.

\paragraph{Pinhole Back-Projection.}
The 3D camera-frame center is recovered by intersecting the offset 2D center with the predicted depth via the pinhole model:

\begin{equation}
    X_c = \frac{(u - c_x)}{f_x} \cdot \hat{z}_c, \quad Y_c = \frac{(v - c_y)}{f_y} \cdot \hat{z}_c, \quad Z_c = \hat{z}_c,
    \label{eq:cam_center}
\end{equation}

where $u = c_{x,\text{2d}} + \hat{\delta}_x$, $v = c_{y,\text{2d}} + \hat{\delta}_y$.
The 8 corners are built at $(\pm l/2, \pm w/2, \pm h/2)$, rotated by the predicted yaw, and translated to $[X_c, Y_c, Z_c]^\top$.

\paragraph{Weight Initialization.}
Default initialization causes extreme variance in the initial 3D loss.  We apply targeted initialization: Xavier-uniform for the context FC, $\mathcal{N}(0, 0.001)$ weights with sensible biases for depth (initial $\approx$1.3--1.8\,m), dimensions (initial $\approx$0.7\,m), and identity rotation ($\sin{=}0, \cos{=}1$).

\subsection{OVMono3D Disentangled 3D Loss}
\label{sec:loss_3d}

The 3D loss follows OVMono3D~\cite{zhang2023ovmono3dmonocular3dobject}, combining uncertainty-weighted supervision with geometry-level disentanglement.

\vspace{2mm}
\noindent\textbf{Disentangled Attribute Losses.} Given predicted corners $\hat{\mat{C}}$ and GT corners $\mat{C}^* \in \mR^{8 \times 3}$, both are decomposed via PCA into center $\mat{p}$, oriented dimensions $\mat{d}=(l,w,h)$, and yaw $\theta$.  For each attribute group $a \in \{xy, z, \text{dims}, r\}$, a ``mixed'' box uses the predicted attribute with GT values for all others:
\begin{alignat}{2}
    \mathcal{L}_{xy}:& \quad \mat{C}_{xy} &&= \op{build}(\hat{p}_{xy} \| p_z^*, \mat{d}^*, \theta^*), \label{eq:l_xy} \\
    \mathcal{L}_{z}:& \quad \mat{C}_{z} &&= \op{build}(p_{xy}^* \| \hat{p}_z, \mat{d}^*, \theta^*), \label{eq:l_z} \\
    \mathcal{L}_{\text{dims}}:& \quad \mat{C}_d &&= \op{build}(\mat{p}^*, \hat{\mat{d}}, \theta^*), \label{eq:l_dims} \\
    \mathcal{L}_{r}:& \quad \mat{C}_r &&= \op{build}(\mat{p}^*, \mat{d}^*, \hat{\theta}). \label{eq:l_r}
\end{alignat}
An additional holistic loss $\mathcal{L}_{\text{all}} = \op{Chamfer}_{\text{SL1}}(\hat{\mat{C}}, \mat{C}^*)$ compares the full corners directly.  All Chamfer computations use Smooth L1 distance for stability, and values are divided by the GT box diagonal for scale invariance.

\vspace{2mm}
\noindent\textbf{Per-Sample Uncertainty Weighting.} The uncertainty $\hat{\mu}_i$ is STE-clamped to $[-5, 10]$ and per-sample $\mathcal{L}_{\text{3D},i}$ is clamped at 100. The total per-sample 3D loss is:
\begin{equation}
    \mathcal{L}_i = \sqrt{2} \cdot \exp(-\hat{\mu}_i) \cdot \mathcal{L}_{\text{3D},i} + \hat{\mu}_i, \quad \mathcal{L}_{\text{3D},i} = \mathcal{L}_{xy,i} + \mathcal{L}_{z,i} + \mathcal{L}_{\text{dims},i} + \mathcal{L}_{r,i} + \mathcal{L}_{\text{all},i}.
    \label{eq:uncertainty}
\end{equation}

\vspace{2mm}
\noindent\textbf{Total Training Loss.} The total loss combines all Faster R-CNN losses and the 3D loss:
\begin{equation}
    \mathcal{L}_{\text{total}} = w_{\text{cls}} \mathcal{L}_{\text{cls}} + w_{\text{box}} \mathcal{L}_{\text{box}} + w_{\text{obj}} \mathcal{L}_{\text{rpn-obj}} + w_{\text{rpn}} \mathcal{L}_{\text{rpn-box}} + w_{\text{3d}}(e) \cdot \mathcal{L}_{\text{3D}},
    \label{eq:total_loss}
\end{equation}
where all weights default to 1.0.  To stabilize early training, $w_{\text{3d}}(e)$ follows a three-phase ramp: zero for the first $R$ epochs (2D-only), linearly increasing from $R$ to $2R$, and full weight thereafter ($R{=}5$ by default).

\begin{table}[!t]
\centering
\caption{Key training configuration parameters.}
\label{tab:config}
\small
\begin{tabular}{@{}llp{6cm}@{}}
\toprule
\textbf{Parameter} & \textbf{Default} & \textbf{Description} \\
\midrule
\texttt{model} & \texttt{v2} & Backbone variant (Table~\ref{tab:backbone_variants}) \\
\texttt{lr} & $10^{-4}$ & Peak learning rate \\
\texttt{weight\_decay} & $10^{-3}$ & AdamW weight decay \\
\texttt{batch\_size} & 128 & Per-GPU batch size \\
\texttt{epochs} & 70 & Total training epochs \\
\texttt{max\_grad\_norm} & 1.0 & Gradient clipping threshold \\
\texttt{head\_3d\_mode} & \texttt{unified} & 3D head integration (unified or separate) \\
\texttt{head\_3d\_version} & \texttt{v1} & 3D head architecture version \\
\texttt{weight\_3d} & 1.0 & Target 3D loss weight \\
\texttt{weight\_3d\_ramp\_epochs} & 5 & Staged ramp duration ($R$) \\
\texttt{pixel\_limit} & 255,000 & Max pixels for resize \\
\bottomrule
\end{tabular}
\end{table}

\subsection{Dataset and Training}
\label{sec:dataset}

\vspace{2mm}
\noindent\textbf{Data Sources.} The dataset loads 2D annotations (37 object classes) from standard Action Genome pickles, 3D annotations (8-corner OBBs in camera frame with per-video intrinsics) from per-video pickle files.  Images are resized to an aspect-ratio-preserving resolution where both dimensions are multiples of patch size $P$, with total pixels $\leq$\,255K.  Camera intrinsics are scaled proportionally.  Images are grouped into \emph{resolution buckets} sharing the same target size, enabling padding-free batching.

\vspace{2mm}
\noindent\textbf{Training Configuration.} Training uses AdamW~\cite{loshchilov2019decoupledweightdecayregularization} ($\text{lr}{=}10^{-4}$, $\text{wd}{=}10^{-3}$) with linear warmup (1\% of steps) followed by cosine annealing.  Mixed precision is enabled via \texttt{GradScaler}, with gradients clipped at norm 1.0.  Non-finite losses trigger batch skipping.  CUDA optimizations include cuDNN auto-tuning, TF32 arithmetic, and Flash Attention.  Key hyperparameters are listed in Table~\ref{tab:config}.

\vspace{2mm}
\noindent\textbf{Evaluation.} Evaluation uses a single fused forward pass over the full test set, computing 2D and 3D metrics simultaneously.

\paragraph{2D Metrics.}  We report COCO-style mAP over IoU thresholds $[0.50, 0.55, \ldots, 0.95]$, along with mAP$_{50}$ and mAP$_{75}$.

\paragraph{3D Metrics.}  3D evaluation is performed on matched prediction--GT pairs (matched via 2D IoU $\geq 0.5$, greedy, same-class).  Box-level metrics include bidirectional Chamfer distance, corner L2, and oriented 3D IoU (via Sutherland--Hodgman polygon clipping~\cite{sutherland1974reentrantpolygonclipping}).  Attribute-level metrics include center L2, dimensions L1, and wrapped rotation error.  IoU hit rates at 50\% and 75\% measure 3D regression quality for correctly detected objects.

\vspace{2mm}
\noindent\textbf{Backbone and Training Mode Comparison.} We evaluate the impact of backbone choice and training mode on 2D detection quality.  Specifically, we compare three backbones; DINOv2-B (ViT-B/14, 86M), DINOv2-L (ViT-L/14, 304M), and DINOv3-L (ViT-L/16, 304M) under two training setups:
\begin{itemize}[nosep]
    \item \textbf{Separate}: The 2D detector and 3D head are trained independently; the 3D branch does not influence 2D detection gradients.
    \item \textbf{Joint}: Both 2D and 3D heads are trained end-to-end with the staged loss ramp (Eq.~\ref{eq:total_loss}), allowing 3D supervision to refine shared representations.
\end{itemize}

\begin{table}[!t]
\centering
\small
\caption{Comparison of DINO backbones under joint and separate training setups.}
\label{tab:backbone_comparison}
\begin{tabular}{llccc}
\toprule
Backbone & Training & mAP & mAP$_{50}$ & mAP$_{75}$ \\
\midrule
DINOv2-B & Separate & 0.0988 & 0.2333 & 0.0692 \\
DINOv2-L & Separate & 0.1103 & 0.2627 & 0.0739 \\
DINOv3-L & Separate & 0.1762 & \textbf{0.3667} & 0.1461 \\
DINOv2-B & Joint & 0.0998 & 0.2377 & 0.0672 \\
DINOv2-L & Joint & 0.1086 & 0.2610 & 0.0712 \\
DINOv3-L & Joint & \textbf{0.1799} & \textbf{0.3660} & \textbf{0.1552} \\
\bottomrule
\end{tabular}
\end{table}

Results are shown in Table~\ref{tab:backbone_comparison}.  Several observations stand out.

\paragraph{DINOv3-L significantly outperforms DINOv2 variants.}  DINOv3-L achieves mAP of 0.18, a 60\% relative improvement over DINOv2-L (0.11) and 80\% over DINOv2-B (0.10).  This gap is consistent across all IoU thresholds, with DINOv3-L reaching 0.37 mAP$_{50}$ versus 0.26 for DINOv2-L.  The improvement is attributable to DINOv3's stronger spatial representations and enhanced training recipe, which yield features better suited for localization.

\paragraph{Scaling from Base to Large within DINOv2 yields marginal gains.}  DINOv2-L improves over DINOv2-B by only $\sim$1 mAP point (0.11 vs.\ 0.10), suggesting that the DINOv2 representation quality---rather than model capacity---is the bottleneck.  In contrast, the architecture and pretraining changes introduced in DINOv3 unlock substantially better performance.


Table~\ref{tab:3d_iou_results} complements the 2D analysis with 3D localization quality.

\begin{table}[!t]
\centering
\small
\caption{3D localization performance of different DINO backbones under joint and separate training setups.  Mean IoU$_{3D}$ measures the average 3D overlap, while Hit@0.50 denotes the fraction of predictions exceeding the IoU$_{3D}$ threshold.}
\label{tab:3d_iou_results}
\begin{tabular}{llcc}
\toprule
Backbone & Training & Mean IoU$_{3D}$ & Hit@0.50 \\
\midrule
DINOv2-B & Separate & \textbf{0.1446} & \textbf{0.0137} \\
DINOv2-L & Separate & 0.1377 & 0.0117 \\
DINOv3-L & Separate & 0.1295 & 0.0111 \\
DINOv2-B & Joint & 0.1411 & 0.0120 \\
DINOv2-L & Joint & 0.1390 & 0.0104 \\
DINOv3-L & Joint & 0.1285 & 0.0128 \\
\bottomrule
\end{tabular}
\end{table}

\vspace{2mm}
\noindent\textbf{Numerical Stability Notes.} All 3D loss computation is forced to float32 to avoid Chamfer overflow in float16.  The 3D loss operates on at most 64 positive proposals per batch, randomly subsampled, to bound memory under the $5\times$ disentangled expansion.  Resolution bucketing reads only JPEG/PNG headers ($\leq$32 bytes) for the ${\sim}7$K per-video dimension queries.

%% file: sup_tex_files/sup_experiments_wsgg.tex
\subsection{Overview}
\label{sec:overview}

We present a family of three methods for \emph{World Scene Graph Generation} (WorldSGG), the task of predicting a dynamic scene graph over all objects in a video---including those that are temporarily \emph{occluded}, have \emph{left the camera field of view}, or have \emph{never been directly observed} in a given frame.  Unlike conventional video scene graph generation (VidSGG)~\cite{cong2021spatialtemporaltransformerdynamicscene,ji2019actiongenomeactionscomposition}, which only reasons about \emph{currently visible} object--object relationships, WorldSGG maintains a persistent world state and predicts relationships for all entities, both seen and unseen, using 3D geometric scaffolds, temporal memory mechanisms, and associative retrieval.

Our three methods offer complementary approaches to this challenge, each exploring a distinct design philosophy for handling unseen objects while drawing on different threads of the SGG, video understanding, and representation learning literature:

\begin{enumerate}
    \item \textbf{PWG} (Persistent World Graph, Sec.~\ref{sup:sec:pwg}): Takes a \emph{memory-based} approach, maintaining a \emph{persistent object memory buffer} that freezes each object's visual features at its last visible appearance.  This design draws on principles of object permanence from cognitive science~\cite{piloto2022intuitive} and multi-object tracking~\cite{bewley2016simple}.

    \item \textbf{4DST} (4D Scene Transformer, Sec.~\ref{sup:sec:4dst}): Takes a \emph{temporal-attention} approach, using a \emph{per-object bidirectional temporal transformer} to learn how to interpolate and extrapolate object appearance across time.  This extends the factorized spatial-temporal attention paradigm of video transformers~\cite{arnab2021vivit,bertasius2021space} and VidSGG methods like STTran~\cite{cong2021spatialtemporaltransformerdynamicscene} from visible-only 2D reasoning to full 4D (3D~+~time) world-state estimation.

    \item \textbf{MWAE} (Masked World Auto-Encoder, Sec.~\ref{sup:sec:mwae}): Introduces a \emph{masked auto-encoder} framework for scene graph completion, applying the encoder--decoder factorization of MAE~\cite{he2022masked} and VideoMAE~\cite{tong2022videomae} to relational scene understanding---where occlusions and camera motion provide natural masking, and a 3D geometric scaffold provides the complete structural prior.
\end{enumerate}

All methods operate on the same input: pre-extracted per-object DINOv2/v3~\cite{oquab2024dinov2learningrobustvisual} visual features, 3D oriented bounding box corners from a persistent global wireframe, and camera extrinsic matrices.  Relationship predictions are made for three categories: \textit{attention} (3 classes), \textit{spatial} (6 classes), and \textit{contacting} (17 classes).

\subsection{Shared Architectural Components}
\label{sup:sec:shared}

All methods reuse the following modules, which are defined centrally and instantiated by each method.  We describe each in detail.

\vspace{2mm}
\noindent\textbf{Global Structural Encoder.} The \textit{Global Structural Encoder} converts world-frame 3D oriented bounding boxes (OBBs) into per-object structural tokens and a permutation-invariant global scene summary.  This module grounds every method in the 3D world geometry, ensuring that even objects without visual evidence receive a rich spatial representation.

\paragraph{Input.} An OBB is parameterized by its $8$ corner vertices $\mat{C}_n \in \mR^{8 \times 3}$ for each of $N$ objects.

\paragraph{Local Geometry.}  To obtain a translation-invariant shape encoding, we center each box:
\begin{align}
    \bar{\mat{c}}_n &= \frac{1}{8}\sum_{i=1}^{8}\mat{c}_{n,i} \in \mR^{3}, \label{eq:center} \\
    \tilde{\mat{c}}_{n,i} &= \mat{c}_{n,i} - \bar{\mat{c}}_n \in \mR^{3}, \quad i=1,\dots,8. \label{eq:local}
\end{align}
The local corners are flattened to a 24-dimensional vector and concatenated with the absolute center, yielding a $27$-dimensional per-object input:
\begin{equation}
    \mat{x}_n = \bigl[\op{flatten}(\tilde{\mat{c}}_{n,1},\dots,\tilde{\mat{c}}_{n,8}) \;\|\; \bar{\mat{c}}_n \bigr] \in \mR^{27}.
    \label{eq:gse_input}
\end{equation}
This preserves the full rigid geometry (dimensions, orientation) that per-corner max-pooling would discard.

\paragraph{Per-Object Tokens.}  A shared MLP ($27 \!\to\! d_h \xrightarrow{\text{ReLU}} 2d_h \xrightarrow{\text{ReLU+LN}} d_{\text{struct}}$) with ReLU activations and an intermediate LayerNorm produces:
\begin{equation}
    \mat{g}_n = \op{MLP}_{\text{obj}}\bigl(\mat{x}_n\bigr) \in \mR^{d_{\text{struct}}}.
    \label{eq:gse_obj}
\end{equation}

\paragraph{Global Summary.}  A permutation-invariant global token is obtained via max-pooling over the set of valid objects, followed by a refinement MLP:
\begin{equation}
    \mat{g}_{\text{global}} = \op{MLP}_{\text{global}}\!\Bigl(\max_{n \in \mathcal{V}} \mat{g}_n\Bigr) \in \mR^{d_{\text{struct}}},
    \label{eq:gse_global}
\end{equation}
where $\mathcal{V}$ is the set of valid (non-padding) objects.  Invalid objects are zeroed out before pooling to prevent bias leakage.

\paragraph{Connection to Prior Work.}  The structural encoder plays an analogous role to the bounding-box encoders used in image-level SGG methods~\cite{xu2017scenegraphgenerationiterative,li2017scenegraphgenerationobjects}, but operates in 3D world coordinates rather than 2D image space.  Our encoder processes full OBB geometry, capturing object orientation and 3D extent, information critical for spatial relationship reasoning (e.g., distinguishing ``in front of'' from ``beside'').

\vspace{2mm}
\noindent\textbf{Spatial Positional Encoding.} Rather than treating spatial position as absolute coordinates, we compute \emph{pairwise} geometric features inspired by 3D spatial reasoning.  For each pair of objects $(i, j)$:

\begin{align}
    d_{ij} &= \sqrt{\|\bar{\mat{c}}_i - \bar{\mat{c}}_j\|^2 + \epsilon}, \label{eq:dist} \\
    \hat{\mat{d}}_{ij} &= \frac{\bar{\mat{c}}_i - \bar{\mat{c}}_j}{d_{ij}} \in \mR^{3}, \label{eq:dir} \\
    \rho_{ij} &= \log V_i - \log V_j, \label{eq:volratio}
\end{align}
where $V_n = \ell_a \cdot \ell_b \cdot \ell_c$ is the exact OBB volume computed from the edge lengths of the bottom face and vertical edge, and $\epsilon = 10^{-6}$ prevents NaN gradients at zero distance. The 5-dimensional pairwise feature $[d_{ij},\;\hat{\mat{d}}_{ij},\;\rho_{ij}]$ is processed by a shared MLP, and the result is mean-aggregated over valid neighbors:
\begin{equation}
    \mat{s}_i = \op{Linear}\Bigg(\frac{1}{|\mathcal{N}_i|}\sum_{j \in \mathcal{N}_i}\op{MLP}_{\text{pair}}\bigl([d_{ij},\,\hat{\mat{d}}_{ij},\,\rho_{ij}]\bigr)\Bigg) \in \mR^{d_{\text{model}}},
    \label{eq:spe}
\end{equation}
where $\mathcal{N}_i$ denotes all valid objects excluding $i$. This formulation can be seen as a continuous analogue of discrete spatial predicates (above, near, larger-than), learned end-to-end.  It shares the flavor of SE(3)-equivariant positional encodings~\cite{fuchs2020se3transformers3drototranslationequivariant} but operates on bounding-box-level geometry rather than point clouds.

\vspace{2mm}
\noindent\textbf{Spatial GNN.} Intra-frame object interactions are modeled by a Transformer encoder~\cite{vaswani2023attentionneed} with additive spatial positional encoding:
\begin{equation}
    \mat{H}^{(t)} = \op{TransformerEncoder}\bigl(\mat{X}^{(t)} + \mat{S}^{(t)},\; \text{mask}=\overline{\mat{v}}^{(t)}\bigr) \in \mR^{N \times d_{\text{model}}},
    \label{sup:eq:spatial_gnn}
\end{equation}
where $\mat{X}^{(t)}$ are the input tokens, $\mat{S}^{(t)}$ are the spatial positional encodings (Eq.~\ref{eq:spe}), and $\overline{\mat{v}}^{(t)}$ is the validity-based padding mask.  Pre-LN~\cite{xiong2020layernormalizationtransformerarchitecture} architecture is used with a final LayerNorm.

This corresponds to a fully-connected GNN where every valid object attends to every other valid object in the same frame, with 3D geometry informing the attention via additive positional bias.  It can be viewed as implementing a single-frame ``spatial reasoning'' step akin to the graph convolution stage in prior scene graph methods~\cite{xu2017scenegraphgenerationiterative,yang2018graphrcnnscenegraph,li2018factorizablenetefficientsubgraphbased}, but using self-attention in place of message passing.  Compared to the edge-conditioned message passing used in~\cite{yang2018graphrcnnscenegraph}, our additive positional encoding approach is more parameter-efficient while allowing the attention mechanism to implicitly learn geometry-dependent message weights.

\vspace{2mm}
\noindent\textbf{Node Predictor.} Object class prediction is performed by a two-layer MLP head with ReLU activation and Dropout($0.1$) between the layers:
\begin{equation}
    \hat{\mat{y}}_n^{\text{obj}} = \op{MLP}_{\text{node}}(\mat{h}_n) \in \mR^{C_{\text{obj}}},
    \label{eq:node_pred}
\end{equation}
where $C_{\text{obj}}=37$ is the number of object categories.  This is only used under the SGDet mode; in PredCls mode, ground-truth labels are used directly.

\vspace{2mm}
\noindent\textbf{Relationship Predictor.} The \emph{Relationship Predictor} is responsible for the complete edge prediction pipeline.  It operates in three phases.

\paragraph{Phase 1: Relationship Token Formation.}
For each human--object pair $(p, o)$, a relationship token is formed by concatenating:
\begin{equation}
    \mat{r}_{po} = \op{Proj}\bigl([\mat{h}_p \;\|\; \mat{h}_o \;\|\; \mat{u}_{po} \;\|\; \mat{e}_p^{\text{txt}} \;\|\; \mat{e}_o^{\text{txt}}]\bigr) \in \mR^{d_{\text{rel}}},
    \label{eq:rel_token}
\end{equation}
where $\mat{h}_p, \mat{h}_o \in \mR^{d_{\text{model}}}$ are the spatio-temporal aware node representations, $\mat{u}_{po} \in \mR^{d_{\text{union}}}$ are projected union ROI features (or a learnable ``no-union'' embedding), and $\mat{e}_p^{\text{txt}}, \mat{e}_o^{\text{txt}} \in \mR^{d_{\text{text}}}$ are projections of frozen CLIP~\cite{radford2021learning} text embeddings indexed by the predicted object class. The union features are obtained by projecting raw DINO union ROI features ($d_{\text{union\_roi}} \to d_{\text{union}} = d_{\text{rel}}/4$) through a linear layer with ReLU and LayerNorm.  The CLIP text embeddings provide semantic priors: the model can leverage the language grounding that ``a person \textit{standing next to} a chair'' implies spatial proximity.  This integration of language priors with visual and geometric features echoes the multi-modal fusion strategies explored in recent SGG work~\cite{lu2016visualrelationshipdetectionlanguage,chen2019knowledgeembeddedroutingnetworkscene}, but adapted for the 3D world setting.

\paragraph{Phase 2: Relationship Self-Attention.} This can be viewed as a ``message passing over the edge graph'' each relationship can attend to other concurrent relationships to resolve ambiguities (e.g., if person $p$ is \emph{holding} object $o_1$, they are less likely to also be \emph{holding} $o_2$).  This joint reasoning over candidate edges follows the relational context modeling principle~\cite{li2018factorizablenetefficientsubgraphbased,zellers2018neuralmotifsscenegraph}. All $K$ relationship tokens within a frame self-attend to enable inter-edge reasoning:
\begin{equation}
    \mat{R}^{(t)} = \op{RelTransformer}\bigl(\{\mat{r}_{po}\}_{(p,o) \in \mathcal{P}_t}\bigr) \in \mR^{K_t \times d_{\text{rel}}}.
    \label{eq:rel_attn}
\end{equation}

\paragraph{Phase 3: Prediction Heads.}
Three independent two-layer MLP heads ($d_{\text{rel}} \!\to\! d_{\text{rel}}/2 \xrightarrow{\text{ReLU+Dropout}} C$) produce per-class predictions:
\begin{align}
    \hat{\mat{y}}_{\text{att}} &= \op{MLP}_{\text{att}}(\mat{r}) \in \mR^{3}, & &\text{(attention: looking at, not looking at, unsure)} \label{eq:att_head} \\
    \hat{\mat{y}}_{\text{spa}} &= \sigma\!\bigl(\op{MLP}_{\text{spa}}(\mat{r})\bigr) \in [0,1]^{6}, & &\text{(spatial: above, beneath, in front of, \dots)} \label{eq:spa_head} \\
    \hat{\mat{y}}_{\text{con}} &= \sigma\!\bigl(\op{MLP}_{\text{con}}(\mat{r})\bigr) \in [0,1]^{17}. & &\text{(contacting: holding, touching, \dots)} \label{eq:con_head}
\end{align}
Attention uses softmax (mutually exclusive classes), while spatial and contacting use sigmoid (multi-label).
The raw logits are also returned for loss computation.

\vspace{2mm}
\noindent\textbf{Temporal Edge Attention.} Per-pair temporal self-attention enables cross-frame edge reasoning.  For each unique object pair $(p, o)$ observed across multiple frames, the relationship tokens from all frames are gathered into a temporal sequence, augmented with learnable temporal positional embeddings, and processed by a Transformer encoder:
\begin{equation}
    \tilde{\mat{r}}_{po}^{(t)} = \op{TemporalEncoder}\bigl(\{{\mat{r}}_{po}^{(\tau)} + \mat{e}_{\tau}\}_{\tau \in \mathcal{T}_{po}}\bigr),
    \label{eq:tea}
\end{equation}
where $\mathcal{T}_{po}$ is the set of frames in which pair $(p,o)$ is valid, and $\mat{e}_{\tau}$ is the learned temporal positional embedding for frame index $\tau$.

This temporal edge attention is closely related to the approach taken in STTran~\cite{cong2021spatialtemporaltransformerdynamicscene}, which similarly applies temporal self-attention to relationship features across frames.  Our key extension is that the set $\mathcal{T}_{po}$ now includes frames where one or both entities may be \emph{unseen}; their relationship tokens having been completed by the upstream world-state estimation module.  This enables temporal consistency enforcement even across occlusion boundaries, a capability absent from VidSGG methods that only process visible pairs.

\vspace{2mm}
\noindent\textbf{Camera Pose Encoder.} Camera extrinsic matrices $\mat{T} = [\mat{R}\,|\,\boldsymbol{\tau}] \in \mR^{4 \times 4}$ are encoded into:

\paragraph{Global Camera Token.}  The rotation matrix is converted to its continuous 6D representation~\cite{zhou2019continuity} (first two columns of $\mat{R}$) and concatenated with the translation:
\begin{equation}
    \mat{c}_{\text{cam}} = \op{MLP}_{\text{cam}}\bigl([\op{6D}(\mat{R}) \;\|\; \boldsymbol{\tau}]\bigr) \in \mR^{d_{\text{camera}}}.
    \label{eq:cam_global}
\end{equation}

\paragraph{Per-Object Camera-Relative Features.}  For each object $n$, we compute:
\begin{align}
    \mat{d}_n &= \bar{\mat{c}}_n - \boldsymbol{\tau}, & &\text{camera-to-object direction} \label{eq:cam_dir} \\
    \alpha_n &= \hat{\mat{d}}_n \cdot (-\mat{R}_{:,2}), & &\text{view alignment (cosine)} \label{eq:view_align} \\
    \beta_n^{\sin},\,\beta_n^{\cos} &= \hat{\mat{d}}_n \cdot \mat{R}_{:,0},\; \alpha_n, & &\text{azimuth (sin/cos)} \label{eq:azimuth}
\end{align}
These are combined and projected:
\begin{equation}
    \mat{c}_n = \op{MLP}\bigl([\log\|\mat{d}_n\|,\;\alpha_n,\;\beta_n^{\sin},\;\beta_n^{\cos}]\bigr) \in \mR^{d_{\text{camera}}}.
    \label{eq:cam_per_obj}
\end{equation}

Encoding camera-relative geometry for each object is inspired by the view-dependent appearance modeling in neural scene representations~\cite{sitzmann2020scenerepresentationnetworkscontinuous,mildenhall2020nerfrepresentingscenesneural}, adapted to operate at the bounding-box level.  This is a critical differentiator from prior VidSGG methods, which either assume a fixed camera or ignore ego-motion entirely.  In the WorldSGG setting, understanding \emph{why} an object disappeared (left the field of view vs.\ occluded by another object) requires explicit camera-relative reasoning.

\vspace{2mm}
\noindent\textbf{Camera Temporal Encoder.} A dedicated ego-motion encoder processes the \emph{full sequence} of camera poses.  For consecutive frames $(t{-}1, t)$, the relative pose is computed without matrix inversion:
\begin{align}
    \mat{R}_{\text{rel}} &= \mat{R}_t \mat{R}_{t-1}^{\!\top}, \label{eq:rel_rot} \\
    \boldsymbol{\tau}_{\text{rel}} &= \boldsymbol{\tau}_t - \mat{R}_{\text{rel}}\boldsymbol{\tau}_{t-1}. \label{eq:rel_trans}
\end{align}
Each relative pose is encoded via MLP, augmented with learnable temporal positional embeddings, and the full sequence is processed by a self-attention encoder to capture long-range camera motion patterns (e.g., panning, orbiting):
\begin{equation}
    \mat{e}_t = \op{EgoAttn}\bigl(\{\op{MLP}([\op{6D}(\mat{R}_{\text{rel}}^{(\tau)}) \;\|\; \boldsymbol{\tau}_{\text{rel}}^{(\tau)}]) + \mat{e}_{\tau}\}_{\tau=0}^{T-1}\bigr).
    \label{eq:ego_motion}
\end{equation}

Ego-motion encoding enables the model to distinguish between objects that leave view due to camera motion (recoverable by reversing the camera trajectory) vs.\ physical occlusion (requiring associative retrieval from memory).  This distinction is central to world modeling; a capability that prior VidSGG methods~\cite{cong2021spatialtemporaltransformerdynamicscene,ji2019actiongenomeactionscomposition} lack as they assume all object disappearances are equivalent.

\vspace{2mm}
\noindent\textbf{Motion Feature Encoder.} Per-object 3D motion is encoded from finite differences of OBB centers:
\begin{align}
    \mat{v}_n^{(t)} &= \bar{\mat{c}}_n^{(t)} - \bar{\mat{c}}_n^{(t-1)}, \label{eq:velocity} \\
    \mat{a}_n^{(t)} &= \mat{v}_n^{(t)} - \mat{v}_n^{(t-1)}. \label{eq:accel}
\end{align}
Optionally, camera-relative velocity is computed by rotating into the camera frame:
$\mat{v}_n^{\text{cam}} = \mat{R}^{\!\top}\mat{v}_n$, providing parallax-aware motion signals.  The full feature vector $[\mat{v}_n,\,\mat{a}_n,\,\|\mat{v}_n\|,\,\mat{v}_n^{\text{cam}},\,\|\mat{v}_n^{\text{cam}}\|] \in \mR^{11}$ is processed by a shared MLP.  The first frame receives a learnable ``no-motion'' embedding.  Velocity and acceleration are only computed where consecutive frames have valid objects, preventing spurious spikes from padding transitions. 3D motion features extend the 2D optical-flow-based motion signals used in prior VidSGG methods like TRACE~\cite{teng2021targetadaptivecontextaggregation}, which encode pixel-level motion between frames.  By operating in 3D world coordinates, our motion features capture true physical displacement independent of camera motion; essential for distinguishing contact relationships (e.g., ``carrying'' implies co-moving) from spatial coincidence.

\vspace{2mm}
\noindent\textbf{Label Smoother.} VLM pseudo-labels for unseen object relationships are inherently noisy.  We apply label smoothing~\cite{szegedy2015rethinkinginceptionarchitecturecomputer} to soften confidence:
\begin{align}
    \text{CE target:} &\quad [0,0,1,0] \to [\tfrac{\epsilon}{C},\,\tfrac{\epsilon}{C},\,1{-}\epsilon{+}\tfrac{\epsilon}{C},\,\tfrac{\epsilon}{C}], \label{eq:smooth_ce} \\
    \text{BCE target:} &\quad y_c \to \begin{cases} 1 - \epsilon & \text{if } y_c = 1, \\ \frac{\epsilon}{C - k} & \text{if } y_c = 0, \end{cases} \label{eq:smooth_bce}
\end{align}
where $k$ is the number of active labels per sample and $C$ is the total number of classes.

\newpage
\subsection{Method 1: PWG - Persistent World Graph}
\label{sup:sec:pwg}

\begin{figure}[H]
    \centering
    \includegraphics[width=\textwidth]{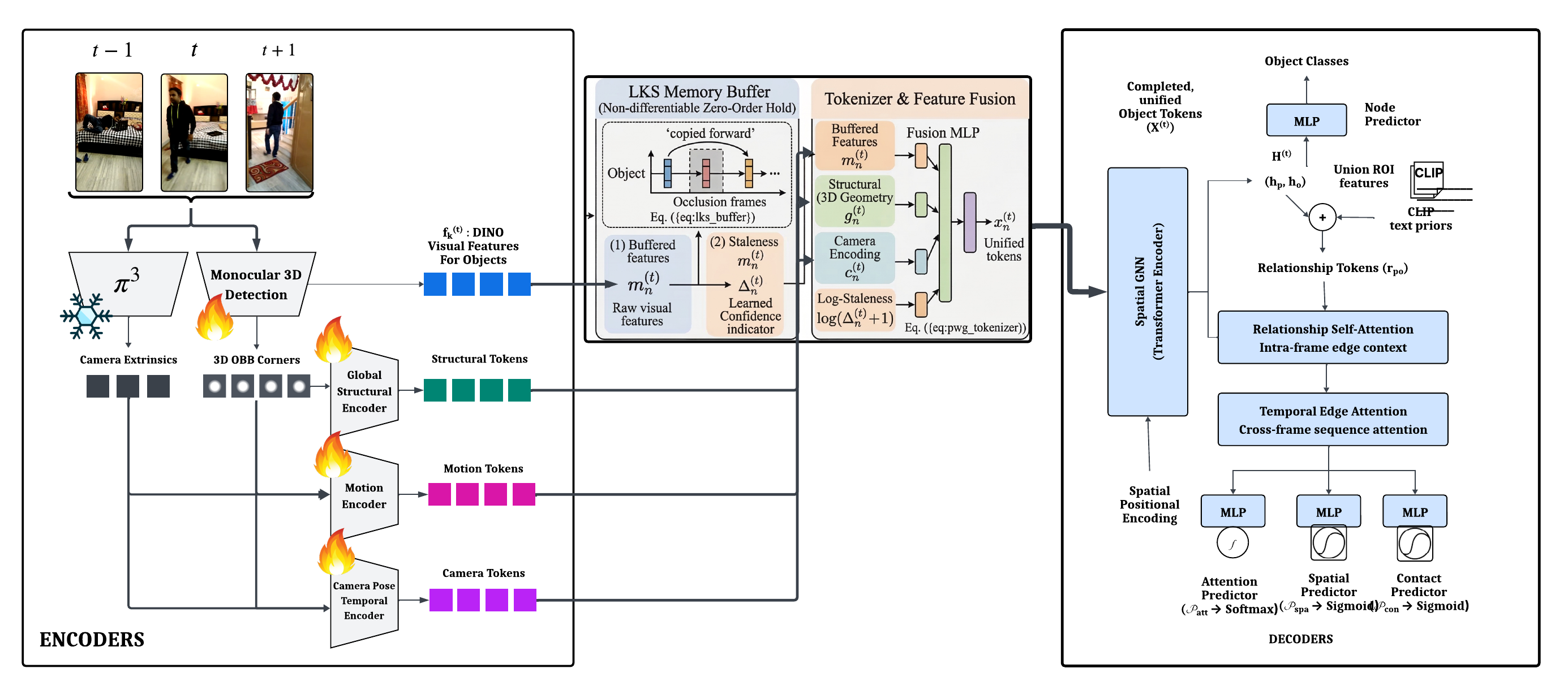}
    \caption{Pipeline of the proposed WorldSGG framework. The model extracts DINO-based object features, 3D OBB corners, and camera pose cues, encodes them into structural, motion, and camera tokens, and fuses them with buffered memory features to handle occlusions and missing observations. The resulting unified tokens are processed by a spatial GNN and relationship decoders to predict node categories and scene graph relations over time. }
    \label{sup:fig:pwg_architecture}
\end{figure}

\vspace{2mm}
\noindent\textbf{High-Level Goal.} PWG takes a \emph{memory-based} approach to WorldSGG, answering the question: \emph{``What if we freeze each object's visual appearance at the last time it was seen, and reason about the full world graph using 3D geometry as scaffolding?''}

This design is rooted in the cognitive science concept of \textbf{object permanence}~\cite{piloto2022intuitive,Spelke1990ObjectPerception} the understanding that objects continue to exist even when they are no longer directly perceived. In developmental psychology, object permanence is considered a foundational prerequisite for physical reasoning. PWG implements this principle at the feature level: when an object leaves the camera's field of view or becomes occluded, the model continues to reason about it using its last-known visual features, anchored by its persistent 3D world position.

\paragraph{Relation to Tracking-Based Methods.}  From the multi-object tracking perspective, PWG can be understood as an extension of tracking-by-detection paradigms \cite{bewley2016simple,wojke2017simpleonlinerealtimetracking} from the \emph{localization} domain to the \emph{relational reasoning} domain.  SORT \cite{bewley2016simple} maintains object states (position, velocity) through occlusions using Kalman filtering; PWG extends this to maintain \emph{visual appearance features} and \emph{relationship predictions} through occlusions using a zero-order hold.

\paragraph{Relation to VidSGG Methods.}  Prior VidSGG approaches such as STTran \cite{cong2021spatialtemporaltransformerdynamicscene} and its variants construct dynamic scene graphs only from \emph{currently visible} detections.  When an object disappears, it simply drops out of the graph, a significant limitation for downstream tasks like activity understanding and navigation that require persistent world state.  PWG addresses this by maintaining a \emph{complete} world graph at every frame, regardless of visibility.

\vspace{2mm}
\noindent\textbf{LKS Memory Buffer.} The memory buffer is a \emph{non-differentiable} zero-order hold that operates on raw DINO features.  For each object $n$ at frame $t$:
\begin{equation}
    \mat{m}_n^{(t)} = \begin{cases}
        \mat{f}_n^{(t)} & \text{if object $n$ is visible at $t$}, \\
        \mat{f}_n^{(\tau^*)} & \text{where } \tau^* = \arg\min_{\tau:\text{vis}(n,\tau)} |t - \tau|, \\
        \bm{0} & \text{if object $n$ is never seen (fog of war)}.
    \end{cases}
    \label{sup:eq:lks_buffer}
\end{equation}
Here, $\mat{f}_n^{(t)} \in \mR^{d_{\text{roi}}}$ denotes the raw DINO ROI features for object $n$ at frame $t$, and the nearest visible frame $\tau^*$ is the bidirectional nearest-neighbor.

\paragraph{Implementation.}  All operations are $O(T \cdot N)$, and the buffer is implemented via vectorized \texttt{cummax} operations over the full temporal axis: (1) \textbf{Forward pass:} $\text{last\_seen}(t,n) = \max_{\tau \le t} \{\tau : \text{vis}(n,\tau)\}$ via forward cummax; (2) \textbf{Backward pass:} $\text{next\_seen}(t,n) = \min_{\tau \ge t} \{\tau : \text{vis}(n,\tau)\}$ via reversed cummax on negated frame indices; (3) \textbf{Nearest selection:} Pick the direction with smaller staleness.

\paragraph{Staleness.}  The buffer also outputs a per-object staleness value $\Delta_n^{(t)} = |t - \tau^*| \in \mathbb{N}_0$, capped at a fixed sentinel ($\Delta=1000$ for new objects) to decouple from video length.  This staleness value serves as a learned confidence indicator: the model can learn to weight stale features differently from fresh observations, analogous to the age-based gating used in DeepSORT~\cite{wojke2017simpleonlinerealtimetracking} for track management.

\vspace{2mm}
\noindent\textbf{Tokenizer.} The tokenizer fuses geometry, buffered features, camera information, and staleness metadata:
\begin{equation}
    \mat{x}_n^{(t)} = \op{FusionProj}\bigl([\mat{g}_n^{(t)} \;\|\; \mat{m}_n^{(t)} \;\|\; \mat{c}_n^{(t)} \;\|\; \log(\Delta_n^{(t)} + 1)]\bigr) \in \mR^{d_{\text{model}}},
    \label{eq:pwg_tokenizer}
\end{equation}
where $\op{FusionProj}$ is a single-layer linear projection followed by LayerNorm and GELU activation, and the log-staleness term provides a smooth, bounded indicator of feature freshness.  The visual projection from $d_{\text{roi}}$ to $d_{\text{model}}$ happens \emph{inside} the fusion MLP, allowing gradients to flow through the alignment layer.

\vspace{2mm}
\noindent\textbf{Full Pipeline.} The PWG pipeline processes all $T$ frames in a single batched forward pass (with $B=T$):
\begin{enumerate}
    \item \textbf{Vectorized LKS Buffer} (non-differentiable): $\mat{f}_{\text{all}} \to (\mat{m}_{\text{all}},\,\bm{\Delta}_{\text{all}})$
    \item \textbf{Structural Encoding}: $\mat{C}_{\text{all}} \to \mat{G}_{\text{all}} \in \mR^{T \times N \times d_{\text{struct}}}$
    \item \textbf{Camera Encoding}: $\mat{T}_{\text{all}} \to \mat{c}_{\text{all}} \in \mR^{T \times N \times d_{\text{camera}}}$
    \item \textbf{Tokenization}: $[\mat{G}, \mat{m}, \mat{c}, \log \bm{\Delta}] \to \mat{X} \in \mR^{T \times N \times d_{\text{model}}}$
    \item \textbf{Spatial GNN}: $\mat{X} \to \mat{H} \in \mR^{T \times N \times d_{\text{model}}}$
    \item \textbf{Node Prediction}: $\mat{H} \to \hat{\mat{Y}}^{\text{obj}} \in \mR^{T \times N \times C_{\text{obj}}}$
    \item \textbf{Batched Edge Formation + Self-Attention}: $(\mat{H}, \hat{\mat{Y}}^{\text{obj}}) \to \mat{R} \in \mR^{T \times K_{\max} \times d_{\text{rel}}}$
    \item \textbf{Temporal Edge Attention}: $\mat{R} \to \tilde{\mat{R}} \in \mR^{T \times K_{\max} \times d_{\text{rel}}}$
    \item \textbf{Prediction Heads}: $\tilde{\mat{R}} \to (\hat{\mat{y}}_{\text{att}},\,\hat{\mat{y}}_{\text{spa}},\,\hat{\mat{y}}_{\text{con}})$
\end{enumerate}

\subsubsection{PWG Loss Function}
\label{sup:sec:pwg_loss}

The loss partitions human--object pairs into three visibility buckets based on whether the constituent entities are directly observed: (1) \textbf{Vis--Vis}: Both person and object are visible $\to$ clean manual ground-truth labels, full weight. (2) \textbf{Vis--Unseen}: Exactly one entity is visible $\to$ VLM pseudo-labels, $\lambda_{\text{vlm}}$-weighted. (3) \textbf{Unseen--Unseen}: Neither entity is visible $\to$ VLM pseudo-labels, $\lambda_{\text{vlm}}$-weighted. The latter two buckets receive identical treatment (both use $\lambda_{\text{vlm}}$ scaling and optional label smoothing).

All buckets share a single global denominator $N_{\text{total}} = \sum_{t}\sum_k \mathbb{1}[\text{pair}_{t,k} \text{ is valid}]$, ensuring each pair contributes equally to the gradient:
\begin{align}
    \mathcal{L}_{\text{att}}^{\text{vis}} &= \frac{1}{N}\sum_{(p,o) \in \text{visible}} \op{CE}\bigl(\hat{\mat{y}}_{\text{att}},\, y_{\text{att}}\bigr), \label{eq:pwg_att_vis} \\
    \mathcal{L}_{\text{spa}}^{\text{vis}} &= \frac{1}{N}\sum_{(p,o) \in \text{visible}} \op{BCE}\bigl(\hat{\mat{y}}_{\text{spa}},\, y_{\text{spa}}\bigr), \label{eq:pwg_spa_vis} \\
    \mathcal{L}_{\text{con}}^{\text{vis}} &= \frac{1}{N}\sum_{(p,o) \in \text{visible}} \op{BCE}\bigl(\hat{\mat{y}}_{\text{con}},\, y_{\text{con}}\bigr). \label{eq:pwg_con_vis}
\end{align}
For unseen buckets, the VLM pseudo-labels are down-weighted by $\lambda_{\text{vlm}}$ and optionally smoothed:
\begin{align}
    \mathcal{L}_{\text{att}}^{\text{unseen}} &= \frac{\lambda_{\text{vlm}}}{N}\sum_{(p,o) \in \text{unseen}} \op{KL}\bigl(\log\op{softmax}(\hat{\mat{y}}_{\text{att}}),\, \tilde{y}_{\text{att}}\bigr), \label{eq:pwg_att_unseen} \\
    \mathcal{L}_{\text{spa}}^{\text{unseen}} &= \frac{\lambda_{\text{vlm}}}{N}\sum_{(p,o) \in \text{unseen}} \op{BCE}\bigl(\hat{\mat{y}}_{\text{spa}},\, \tilde{y}_{\text{spa}}\bigr), \label{eq:pwg_spa_unseen} \\
    \mathcal{L}_{\text{con}}^{\text{unseen}} &= \frac{\lambda_{\text{vlm}}}{N}\sum_{(p,o) \in \text{unseen}} \op{BCE}\bigl(\hat{\mat{y}}_{\text{con}},\, \tilde{y}_{\text{con}}\bigr), \label{eq:pwg_con_unseen}
\end{align}
where $\tilde{y}$ denotes the label-smoothed target. The full loss is:
\begin{equation}
    \mathcal{L}_{\text{PWG}} = \sum_{r \in \{\text{att, spa, con}\}} \bigl(\mathcal{L}_{r}^{\text{vis}} + \mathcal{L}_{r}^{\text{unseen}}\bigr) + \mathcal{L}_{\text{node}}.
    \label{sup:eq:pwg_total}
\end{equation}
In SGDet mode, we include node classification loss $\mathcal{L}_{\text{node}} = \op{CE}(\hat{\mat{y}}^{\text{obj}}, y^{\text{obj}}) / N_{\text{nodes}}$.

\newpage
\subsection{Method 2: 4DST --- 4D Scene Transformer}
\label{sup:sec:4dst}


\vspace{2mm}
\noindent\textbf{High-Level Goal.} 4DST takes a \emph{temporal-attention} approach to WorldSGG.  Where PWG uses a non-differentiable memory buffer, 4DST instead employs a \emph{Temporal Object Transformer} that performs per-object bidirectional self-attention over the entire video, enabling the model to learn \emph{how} to interpolate, extrapolate, and contextualize object appearance across time in a differentiable manner.

\paragraph{Relation to Video Transformers.}  The 4DST architecture draws direct inspiration from the factorized spatial-temporal attention paradigm that has proven highly effective in video understanding.  ViViT~\cite{arnab2021vivit} and TimeSformer~\cite{bertasius2021space} demonstrated that factorizing attention into spatial (within-frame) and temporal (across-frame) components yields better efficiency and performance than joint space-time attention.  4DST adopts this factorization at the scene graph level: the \emph{Spatial GNN} performs within-frame object-to-object attention, while the \emph{Temporal Object Transformer} performs per-object across-frame attention.  The critical extension is that both modules operate over the \emph{complete} set of objects (visible and unseen), enabling temporal reasoning to fill in representations at occlusions.

\paragraph{Relation to VidSGG Methods.}  STTran~\cite{cong2021spatialtemporaltransformerdynamicscene} was the first to apply temporal transformers to video scene graph generation, using self-attention across frames for both node-level and edge-level features.  However, STTran operates exclusively on visible detections---when an object disappears, its temporal sequence simply ends.  4DST extends this approach in two key ways: (i) the temporal sequence includes \emph{all} frames, with a learned visibility embedding distinguishing observed from inferred tokens; and (ii) the temporal attention operates over 3D world-frame features enriched with camera pose and motion information, enabling view-invariant temporal reasoning.

\paragraph{Relation to 4D Scene Understanding.}  Recent work on 4D scene graph generation~\cite{wald2020learning3dsemanticscene} has explored jointly reasoning over 3D space and time for scene understanding.  4DST can be viewed as extending this line of work by operating in an \emph{egocentric video} setting with dynamic camera motion, rather than static scene scans.  The camera temporal encoder explicitly models ego-motion, enabling the model to disentangle object motion from camera-induced apparent motion.

\vspace{2mm}
\noindent\textbf{Temporal Object Transformer.} For each object $n$, the Temporal Object Transformer builds a sequence of $T$ tokens (one per frame) from multi-modal fused features and processes them with bidirectional self-attention.

\paragraph{Feature Fusion.}  Per-frame tokens are formed by concatenating projected visual features, structural tokens, camera features, motion features, and broadcast ego-motion tokens:
\begin{equation}
    \mat{z}_n^{(t)} = \op{InputProj}\bigl([\op{VisProj}(\mat{f}_n^{(t)}) \;\|\; \mat{g}_n^{(t)} \;\|\; \mat{c}_n^{(t)} \;\|\; \mat{\mu}_n^{(t)} \;\|\; \mat{e}_t]\bigr) \in \mR^{d_{\text{memory}}},
    \label{eq:4dst_fusion}
\end{equation}
where $\mat{\mu}_n^{(t)}$ is the motion feature and $\mat{e}_t$ is the ego-motion token.

\paragraph{Temporal Positional Encoding.}  Sinusoidal positional encoding~\cite{vaswani2023attentionneed} is added to each token:
\begin{equation}
    \mat{z}_n^{(t)} \mathrel{+}= \op{PE}_{\text{sin}}(t) \in \mR^{d_{\text{memory}}}.
    \label{eq:4dst_pe}
\end{equation}

\paragraph{Visibility Embedding.}  A learned embedding distinguishes directly observed vs.\ unobserved tokens:
\begin{equation}
    \mat{z}_n^{(t)} \mathrel{+}= \mat{e}_{\text{vis}}[\mathbb{1}[\text{vis}(n, t)]] \in \mR^{d_{\text{memory}}}.
    \label{eq:vis_emb}
\end{equation}

This binary indicator replaces explicit gating mechanisms (e.g., GRU-based seen/unseen branches), allowing the transformer to learn its own visibility-conditioned attention patterns.  Tokens marked as unseen still receive full geometric and camera features from the 3D scaffold, providing the transformer with a rich structural context from which to infer the missing visual representation.

\paragraph{Bidirectional Self-Attention.}  Each object's $T$-length sequence is independently processed by a Transformer encoder \emph{without causal masking}:
\begin{equation}
    \{\tilde{\mat{z}}_n^{(t)}\}_{t=1}^{T} = \op{TransformerEncoder}\bigl(\{\mat{z}_n^{(t)}\}_{t=1}^{T}\bigr).
    \label{eq:4dst_biattn}
\end{equation}
The Transformer batches over $N$ objects (batch dim $=N$, sequence length $=T$).  Invalid (padding) frames are masked via \texttt{src\_key\_padding\_mask}.

\paragraph{Connection to Video Object Memory.}  This per-object temporal attention can be viewed as a differentiable analogue of the memory mechanisms used in long-term video object segmentation~\cite{cheng2022xmemlongtermvideoobject,oh2019videoobjectsegmentationusing}.  XMem~\cite{cheng2022xmemlongtermvideoobject} maintains a multi-scale memory bank for each tracked object and uses attention-based retrieval to handle occlusions; 4DST achieves a similar effect through per-object bidirectional self-attention, but with the critical advantage that the entire pipeline is end-to-end differentiable and jointly optimized for relationship prediction.

\subsubsection{Full Pipeline}

The 4DST pipeline processes all $T$ frames in a single batched forward pass:
\begin{enumerate}
    \item \textbf{Structural Encoding}: $\mat{C}_{\text{all}} \to \mat{G}_{\text{all}} \in \mR^{T \times N \times d_{\text{struct}}}$
    \item \textbf{Camera Encoding}: $\mat{T}_{\text{all}} \to \mat{c}_{\text{all}} \in \mR^{T \times N \times d_{\text{camera}}}$
    \item \textbf{Ego-Motion Encoding}: $\mat{T}_{\text{all}} \to \mat{e}_{\text{all}} \in \mR^{T \times d_{\text{camera}}}$
    \item \textbf{Motion Feature Encoding}: $(\mat{v}_{\text{all}},\, \mat{a}_{\text{all}}) \to \mat{\mu}_{\text{all}} \in \mR^{T \times N \times d_{\text{motion}}}$
    \item \textbf{Temporal Object Transformer}: $[\mat{f}, \mat{G}, \mat{c}, \mat{\mu}, \mat{e}] \to \mat{Z} \in \mR^{T \times N \times d_{\text{memory}}}$
    \item \textbf{Spatial GNN}: $\mat{Z} \to \mat{H} \in \mR^{T \times N \times d_{\text{memory}}}$
    \item \textbf{Node Prediction}: $\mat{H} \to \hat{\mat{Y}}^{\text{obj}} \in \mR^{T \times N \times C_{\text{obj}}}$
    \item \textbf{Batched Edge Formation + Self-Attention}: $(\mat{H}, \hat{\mat{Y}}^{\text{obj}}) \to \mat{R} \in \mR^{T \times K_{\max} \times d_{\text{rel}}}$
    \item \textbf{Temporal Edge Attention}: $\mat{R} \to \tilde{\mat{R}} \in \mR^{T \times K_{\max} \times d_{\text{rel}}}$
    \item \textbf{Prediction Heads}: $\tilde{\mat{R}} \to (\hat{\mat{y}}_{\text{att}},\,\hat{\mat{y}}_{\text{spa}},\,\hat{\mat{y}}_{\text{con}})$
\end{enumerate}

\subsubsection{4DST Loss Function}

The loss structure mirrors PWG (Sec.~\ref{sup:sec:pwg_loss}) with visible and unobserved buckets:
\begin{equation}
    \mathcal{L}_{\text{4DST}} = \sum_{r \in \{\text{att, spa, con}\}} \bigl(\mathcal{L}_{r}^{\text{visible}} + \lambda_{\text{vlm}} \cdot \mathcal{L}_{r}^{\text{unobserved}}\bigr) + \mathcal{L}_{\text{node}}.
    \label{sup:eq:4dst_total}
\end{equation}
Label smoothing is applied identically for unobserved pairs.

\newpage
\subsection{Method 3: MWAE - Masked World Auto-Encoder}
\label{sup:sec:mwae}

\vspace{2mm}
\noindent\textbf{High-Level Goal.} MWAE introduces a \emph{masked auto-encoder} paradigm to \\ WorldSGG, reframing the prediction of relationships for unseen objects as a \emph{structured completion problem}.  Given partial observations (visible objects) and a complete geometric scaffold (the persistent 3D wireframe), the model must ``fill in'' the visual representations and relationship predictions for occluded objects.

\paragraph{Relation to Masked Representation Learning.}  The MWAE architecture is directly inspired by the masked auto-encoder (MAE) framework~\cite{he2022masked}, which demonstrated that masking a large fraction of image patches and training an encoder-decoder architecture to reconstruct them produces powerful visual representations.  VideoMAE~\cite{tong2022videomae} extended this to the space-time domain, showing that the temporal redundancy in video allows even more aggressive masking ratios. MWAE transposes this paradigm from the \emph{pixel/patch} domain to the \emph{object/relationship} domain.  The key conceptual parallels are:

\begin{itemize}
    \item \textbf{Masking source:}  In MAE/VideoMAE, patches are randomly masked.  In MWAE, objects are naturally masked by \emph{occlusion}, \emph{camera motion}, and \emph{field-of-view limitations} (the masking arises from the physics of perception).  During training, additional stochastic masking of visible objects supplements the natural masking to create a robust reconstruction signal.
    \item \textbf{Structural scaffold:}  MAE's decoder receives positional embeddings for masked patches.  MWAE's scaffold tokenizer receives \emph{full 3D geometric information} for all objects at all times (even those never observed), providing a far richer structural prior than learned positional embeddings alone.
    \item \textbf{Encoder-decoder factorization:}  MAE processes only visible patches \\ through the encoder, then applies a lightweight decoder to the full set (visible + masked).  MWAE similarly uses its \emph{Associative Retriever} as a decoder that cross-attends from masked object positions to visible memory entries, retrieving view-consistent representations for occluded objects.
\end{itemize}

\paragraph{Relation to Scene Completion.}  MWAE also connects to the semantic scene completion literature~\cite{song2016semanticscenecompletionsingle,li2021bipartitegraphnetworkadaptive}, which aims to predict the full 3D semantic layout of a scene from partial observations.  While those methods operate in voxel space, MWAE performs \emph{relational completion} predicting the relationships that hold between all objects, including those not directly observed.  The scaffold tokenizer's use of complete 3D geometry for all objects (including unseen ones) mirrors the volumetric priors used in scene completion, but at the object-relational level.

\paragraph{Relation to VidSGG.}  Compared to the temporal attention used in STTran~\cite{cong2021spatialtemporaltransformerdynamicscene} and our 4DST model, MWAE makes a distinct architectural choice: rather than having masked and visible tokens jointly participate in self-attention (where masked tokens might attend to each other and propagate noise), MWAE uses \emph{asymmetric cross-attention} where masked tokens can only attend to visible memory entries.  This mirrors the design difference between BERT~\cite{devlin2019bertpretrainingdeepbidirectional} (joint encoding of masked and visible tokens) and MAE~\cite{he2022masked} (separate encoder for visible, decoder for masked), where the latter was shown to produce superior representations by preventing masked tokens from attending to each other's noise.

\vspace{2mm}
\noindent\textbf{Scaffold Tokenizer.} The Scaffold Tokenizer performs top-down initialization of the world graph.  Every object receives a token at every frame, regardless of visibility:
\begin{equation}
    \mat{x}_n^{(t)} = \op{FusionMLP}\bigl([\mat{g}_n^{(t)} \;\|\; \mat{v}_n^{(t)} \;\|\; \mat{c}_n^{(t)} \;\|\; \mat{\mu}_n^{(t)} \;\|\; \mat{e}_t]\bigr),
    \label{eq:scaffold_token}
\end{equation}
where the visual component is:
\begin{equation}
    \mat{v}_n^{(t)} = \begin{cases}
        \op{VisProj}(\mat{f}_n^{(t)}) & \text{if visible and not masked}, \\
        \mat{e}_{\text{[MASK]}} & \text{otherwise},
    \end{cases}
    \label{eq:scaffold_visual}
\end{equation}
with $\mat{e}_{\text{[MASK]}} \in \mR^{d_{\text{visual}}}$ being a learnable mask embedding.

\paragraph{Training-Time Stochastic Masking.}  During training, a fraction $p_{\text{mask}}$ of \emph{visible} objects are artificially masked (visual features replaced with $\mat{e}_{\text{[MASK]}}$) to force the retriever to learn associative recovery.  This ensures the reconstruction loss has non-trivial targets even when few natural occlusions occur.  This augmentation strategy mirrors the random masking in MAE~\cite{he2022masked}, but targets individual objects rather than image patches, and supplements natural masking from occlusion.

\vspace{2mm}
\noindent\textbf{Associative Retriever.} The Associative Retriever performs per-object bidirectional cross-attention to ``fill in'' masked tokens by consulting that object's visible appearances across all frames.  This retrieval mechanism acts as the ``decoder'' in the MAE analogy: it takes incomplete representations (where some time steps have only geometric scaffolding) and completes them by attending to time steps where the object was actually observed.

\paragraph{Architecture.}  Each object $n$'s temporal sequence (length $T$) is processed by stacked cross-attention layers with pre-norm (LayerNorm before attention and FFN) and GELU activation in the feed-forward network.  The attention structure is: (a) \textbf{Query}: All tokens (visible + masked) with additive camera pose bias $\mat{q}_n^{(t)} = \mat{x}_n^{(t)} + \op{QProj}(\mat{c}_n^{(t)})$; (b) \textbf{Key}: Visible tokens with camera pose bias $\mat{k}_n^{(t)} = \mat{x}_n^{(t)} + \op{KProj}(\mat{c}_n^{(t)})$; (c) \textbf{Value}: Visible tokens (clean, without bias): $\mat{v}_n^{(t)} = \mat{x}_n^{(t)}$. The key padding mask ensures unseen tokens are excluded from the key/value pool.

\paragraph{View-Aware Bias.}  The camera pose projections ($\op{QProj}$, $\op{KProj}$) bias the attention so that retrieval naturally favors memory entries captured from \emph{similar viewpoints}.  This is consistent with the observation that visual features are more transferable between nearby camera positions, and connects to the view-dependent attention mechanisms used in multi-view reconstruction~\cite{wang2021ibrnetlearningmultiviewimagebased,yu2021pixelnerfneuralradiancefields}.

\paragraph{Iterative Key/Value Refinement.}  After each cross-attention layer, visible-position keys and values are updated with the evolved query values:
\begin{align}
    \mat{k}_n^{(t)} &\leftarrow \begin{cases} \mat{q}_n^{(t)} + \op{KProj}(\mat{c}_n^{(t)}) & \text{if visible}, \\ \mat{k}_n^{(t)} & \text{otherwise}. \end{cases} \label{eq:kv_update}
\end{align}
This allows subsequent layers to attend to progressively richer representations, creating an iterative refinement loop analogous to the recurrent refinement used in RAFT~\cite{teed2020raftrecurrentallpairsfield} for optical flow and in iterative attention mechanisms for point cloud processing~\cite{zhao2021pointtransformer}.

\paragraph{Visibility Embedding.}  After retrieval, a learned visibility embedding is added:
\begin{equation}
    \hat{\mat{x}}_n^{(t)} = \mat{x}_n^{(t)} + \mat{e}_{\text{vis}}[\mathbb{1}[\text{vis}(n,t) \wedge \neg\text{masked}(n,t)]].
    \label{eq:mwae_vis_emb}
\end{equation}
The effective visibility excludes artificially masked objects, to prevent a shortcut where the model could bypass retrieval by relying on the visibility indicator.

\vspace{2mm}
\noindent\textbf{Cross-View Reconstruction.} A linear projection head predicts the original DINO visual features from the completed tokens:
\begin{equation}
    \hat{\mat{f}}_n^{(t)} = \op{Linear}(\hat{\mat{x}}_n^{(t)}) \in \mR^{d_{\text{visual}}}.
    \label{eq:recon_pred}
\end{equation}
The reconstruction loss is only applied to \emph{artificially masked} objects (not padding or genuinely unseen objects, which lack meaningful targets).  This cross-view reconstruction objective forces the retriever to produce representations that faithfully capture the object's visual appearance, not just its category---ensuring that fine-grained visual cues relevant to relationship prediction (e.g., hand pose for ``holding'' vs.\ ``touching'') are preserved through the completion process.

\vspace{2mm}
\noindent\textbf{MWAE Loss Function.} MWAE uses a multi-objective loss:
\begin{equation}
    \mathcal{L}_{\text{MWAE}} = \mathcal{L}_{\text{SG}} + \lambda_{\text{recon}} \cdot \lambda_{\text{dom}} \cdot \mathcal{L}_{\text{recon}} + \mathcal{L}_{\text{sim}},
    \label{eq:mwae_total}
\end{equation}
where the terms are:

\paragraph{1. Scene Graph Loss ($\mathcal{L}_{\text{SG}}$).}  Same visible/unseen split as 4DST (Eq.~\ref{sup:eq:4dst_total}).

\paragraph{2. Reconstruction Loss ($\mathcal{L}_{\text{recon}}$).}
\begin{equation}
    \mathcal{L}_{\text{recon}} = \frac{1}{|\mathcal{M}_{\text{art}}|}\sum_{(t,n) \in \mathcal{M}_{\text{art}}} \|\hat{\mat{f}}_n^{(t)} - \mat{f}_n^{(t)}\|^2,
    \label{eq:recon_loss}
\end{equation}
where $\mathcal{M}_{\text{art}}$ is the set of artificially masked (not genuinely unseen) object-frame pairs.  The dominance factor $\lambda_{\text{dom}}$ prevents the reconstruction gradient from being overwhelmed by the scene graph loss in early training.  This reconstruction objective plays the same role as the pixel reconstruction loss in MAE~\cite{he2022masked}, but operates in DINO feature space rather than pixel space.

\paragraph{3. Simulated-Unseen Loss ($\mathcal{L}_{\text{sim}}$).}
For artificially masked objects, the model re-predicts relationships using the completed representations and compares against the \emph{clean} manual ground truth (since these objects were actually visible):
\begin{equation}
    \mathcal{L}_{\text{sim}} = \frac{1}{|\mathcal{S}|}\sum_{(p,o) \in \mathcal{S}} \bigl[\op{CE}(\hat{\mat{y}}_{\text{att}}, y_{\text{att}}) + \op{BCE}(\hat{\mat{y}}_{\text{spa}}, y_{\text{spa}}) + \op{BCE}(\hat{\mat{y}}_{\text{con}}, y_{\text{con}})\bigr].
    \label{eq:sim_loss}
\end{equation}
This creates a powerful self-supervised signal: the model is trained to predict relationships for objects it \emph{can} see but has been told to pretend are unseen.  Since the ground truth is clean (not VLM pseudo-labels), this provides high-quality gradients for learning the completion-to-prediction pathway.  This ``simulated occlusion'' training strategy is conceptually related to the dropout-based regularization used in~\cite{chen2019counterfactualcriticmultiagenttraining} for counterfactual reasoning in SGG, but applied at the object level rather than the feature level.

\vspace{2mm}
\noindent\textbf{Full Pipeline.} The MWAE pipeline processes all $T$ frames in a single batched forward pass:
\begin{enumerate}
    \item \textbf{Structural Encoding}: $\mat{C}_{\text{all}} \to \mat{G}_{\text{all}} \in \mR^{T \times N \times d_{\text{struct}}}$
    \item \textbf{Camera Encoding}: $\mat{T}_{\text{all}} \to \mat{c}_{\text{all}} \in \mR^{T \times N \times d_{\text{camera}}}$
    \item \textbf{Ego-Motion Encoding}: $\mat{T}_{\text{all}} \to \mat{e}_{\text{all}} \in \mR^{T \times d_{\text{camera}}}$
    \item \textbf{Motion Feature Encoding}: $(\mat{v}_{\text{all}},\, \mat{a}_{\text{all}}) \to \mat{\mu}_{\text{all}} \in \mR^{T \times N \times d_{\text{motion}}}$
    \item \textbf{Scaffold Tokenization}: $[\mat{G}, \mat{f}\;\text{or}\;\mat{e}_{\text{[MASK]}}, \mat{c}, \mat{\mu}, \mat{e}] \to \mat{X} \in \mR^{T \times N \times d_{\text{model}}}$
    \item \textbf{Associative Retriever}: $\mat{X} \to \hat{\mat{X}} \in \mR^{T \times N \times d_{\text{model}}}$ \quad (per-object cross-attention)
    \item \textbf{Visibility Embedding}: $\hat{\mat{X}} \mathrel{+}= \mat{e}_{\text{vis}}[\text{effective\_vis}]$
    \item \textbf{Spatial GNN}: $\hat{\mat{X}} \to \mat{H} \in \mR^{T \times N \times d_{\text{model}}}$
    \item \textbf{Node Prediction}: $\mat{H} \to \hat{\mat{Y}}^{\text{obj}} \in \mR^{T \times N \times C_{\text{obj}}}$
    \item \textbf{Cross-View Reconstruction}: $\hat{\mat{X}} \to \hat{\mat{f}} \in \mR^{T \times N \times d_{\text{visual}}}$
    \item \textbf{Batched Edge Formation + Self-Attention}: $(\mat{H}, \hat{\mat{Y}}^{\text{obj}}) \to \mat{R} \in \mR^{T \times K_{\max} \times d_{\text{rel}}}$
    \item \textbf{Temporal Edge Attention}: $\mat{R} \to \tilde{\mat{R}} \in \mR^{T \times K_{\max} \times d_{\text{rel}}}$
    \item \textbf{Prediction Heads}: $\tilde{\mat{R}} \to (\hat{\mat{y}}_{\text{att}},\,\hat{\mat{y}}_{\text{spa}},\,\hat{\mat{y}}_{\text{con}})$
\end{enumerate}
The remaining components (Spatial GNN, Node Predictor, Relationship Predictor, Temporal Edge Attention) are identical to PWG and 4DST.

\subsection{Architectural Relationships and Design Comparisons}
\label{sec:relationships}

The three methods offer complementary design philosophies for handling the core challenge of WorldSGG---reasoning about objects that are not currently visible.  Each makes different trade-offs:

\begin{itemize}
    \item \textbf{PWG vs.\ 4DST - Memory buffer vs.\ temporal attention.}  PWG uses a non-differentiable zero-order hold that preserves each object's last-seen features, offering simplicity and a strong inductive bias analogous to rule-based trackers like SORT~\cite{bewley2016simple}.  4DST instead uses a differentiable Temporal Object Transformer that can interpolate and extrapolate object state using full temporal context, at the cost of additional parameters and the need to learn temporal reasoning end-to-end.  4DST additionally incorporates ego-motion and per-object 3D motion encoding.

    \item \textbf{4DST vs.\ MWAE - Self-attention vs.\ asymmetric cross-attention.}  4DST processes visible and unseen object tokens jointly via self-attention, allowing all tokens to attend to each other.  MWAE instead uses an explicit encoder--decoder factorization inspired by MAE~\cite{he2022masked}: unseen tokens query visible memory entries through asymmetric cross-attention, preventing masked tokens from attending to each other's noise.  The scaffold tokenizer explicitly disentangles geometry (always available from the 3D wireframe) from visual evidence (only available when the object is visible).  MWAE's cross-view reconstruction and simulated-unseen losses additionally provide self-supervised training signals grounded in clean manual annotations.

    \item \textbf{PWG vs.\ MWAE - Feature freezing vs.\ associative retrieval.}  PWG and MWAE represent two ends of the design spectrum for unseen-object completion.  PWG freezes the last-known features as-is, while MWAE actively reconstructs them via cross-attention retrieval over all visible appearances.  This gives MWAE richer completed representations at the cost of a more complex training objective.
\end{itemize}

%% file: sup_tex_files/sup_mllm_wsg.tex
This section assesses how well current Vision-Language Models (VLMs)
perform on the task of \emph{World Scene Graph Generation} (WSGG).
Whereas Sec.~\ref{sup:sec:setup} describes the annotation pipeline
used to produce pseudo-labels, here we formalise the evaluation
protocol, define the evaluation modes, and present the metrics and
stratified analysis used to benchmark VLM performance.

\vspace{2mm}
\noindent\textbf{Task Definition.} Given a video $V$ with $K$ annotated 
frames $\{f_1, \ldots, f_K\}$ and a video-level object set $\mathcal{O}_v$, 
the goal is to predict, for \emph{every} annotated frame $f_k$ and \emph{every} 
object $o \in \mathcal{O}_v$, the three relationship types between the person
and object $o$:
\begin{equation}
  \hat{R}(f_k, o) = \bigl(\hat{a}_{k,o},\;\hat{C}_{k,o},\;\hat{S}_{k,o}\bigr)
\end{equation}
where $\hat{a}$ is the predicted attention label (single-label),
$\hat{C}$ is the predicted set of contacting labels (multi-label), and
$\hat{S}$ is the predicted set of spatial labels (multi-label).

\vspace{2mm}
\noindent\textbf{Object Set Resolution: PredCls vs.\ SGDet.}
\label{sup:sec:mllm:mode} The evaluation supports two standard modes:

\paragraph{PredCls (Predicate Classification).}
Ground-truth object labels from the annotations are used directly
as the video-level object set $\mathcal{O}_v$.  This isolates the
relationship prediction ability of the VLM.

\paragraph{SGDet (Scene Graph Detection).}
Objects are \emph{estimated} from the video content using the VLM
itself.  The estimation proceeds as follows:

\begin{enumerate}[nosep]
  \item \textbf{Subtitle-guided estimation.} The VLM is prompted with
        all available subtitles plus the canonical AG object vocabulary (36
        classes) as a candidate list.  It returns a JSON list of object
        name strings.
  \item \textbf{Vocabulary filtering.} The estimated object set is
        intersected with the canonical AG vocabulary to ensure all
        labels conform to the closed label space:
        $\mathcal{O}_\text{est} = \hat{\mathcal{O}} \cap \mathcal{V}$.
  \item \textbf{Discriminative verification.} For each estimated
        object $o \in \mathcal{O}_\text{est}$, a binary Yes/No
        verification query is sent to the VLM:
        ``\textit{Is there a \{o\} in this scene? Answer only Yes
        or No.}''
        Using batched single-token inference with log-probability
        extraction, a confidence score $p_\text{yes}(o)$ is obtained
        for every object.
\end{enumerate}

\vspace{2mm}
\noindent\textbf{Evaluation Protocol.}
\label{sup:sec:mllm:eval}

\paragraph{Matching Procedure.}
For each frame and each object, predictions are matched against
ground truth by exact object name matching.  For objects present in
the GT but absent from the VLM predictions (or vice versa), the
corresponding entry is recorded as a miss (false negative) or
false positive, respectively.

\paragraph{Metrics.}
We compute the following metrics for each relationship axis: For each label $l$ in the vocabulary of an axis (e.g., each of the
3 attention labels, 17 contacting labels, or 6 spatial labels):
\begin{align}
  \text{Precision}(l) &= \frac{|\text{TP}(l)|}{|\text{TP}(l)| + |\text{FP}(l)|} \\
  \text{Recall}(l) &= \frac{|\text{TP}(l)|}{|\text{TP}(l)| + |\text{FN}(l)|} \\
  \text{F1}(l) &= 2 \cdot \frac{\text{Precision}(l) \cdot \text{Recall}(l)}{\text{Precision}(l) + \text{Recall}(l)}
\end{align}

\paragraph{Macro-averaged and micro-averaged F1.}
Macro-average computes the mean F1 across all labels;
micro-average computes precision and recall over
all instances globally.

\paragraph{Attention accuracy.}
Since attention is single-label, we additionally report
top-1 accuracy.

\paragraph{Verification-aware metrics.}
When verification scores are available ($p_\text{yes}$), we evaluate
at multiple confidence thresholds $\tau \in \{0.3, 0.5, 0.7, 0.9\}$
to generate precision--recall trade-off curves.  A predicted label
is accepted only if $p_\text{yes} \geq \tau$.

\subsection{Evaluation Methods}
\label{sup:sec:mllm:evaluation_methods}

This section provides expanded details on the two evaluation methods summarised in Sec.~\ref{sup:sec:methods}.

\subsubsection{Method 1: Subtitle-Only Generation}
\label{sup:sec:mllm:methods:subtitle}

The subtitle-only method provides a lighter-weight alternative that bypasses graph construction, keyword extraction, and node retrieval entirely.

\paragraph{Subtitle-Enriched Prompting.}
For each relationship query, subtitle context is prepended to the prompt.  If a specific frame index is available, subtitles are \emph{sorted by temporal proximity}:
\begin{equation}
  \text{sorted}(\mathcal{S}, \text{key}=|i_s - i_f|)
\end{equation}
so that the most relevant subtitles appear first.  The concatenated prompt is:
\[
  \text{prompt}_\text{full} = \text{subtitle\_context} \oplus \text{relationship\_query}
\]

\paragraph{Annotation-Driven Per-Frame Clips.}
Each query uses the annotation-driven clip for its target frame.  If no per-frame clip is available, the full-video tensor is used as fallback.

\paragraph{Single Batched Inference.}
All subtitle-enriched prompts across all frames and all objects are collected into a single batch and processed via one batched VLM call (chunked into sub-batches of 64).  Subsequently, one bulk verification call (if not skipped) scores all predictions.

\subsubsection{Method 2: RAG-Based Generation}
\label{sup:sec:mllm:methods:rag}

The RAG (Retrieval-Augmented Generation) method leverages a precomputed scene graph from Phase~1 to produce semantically-grounded pseudo-annotations.

\paragraph{Precomputed Graph Loading.}
Per-clip graphs produced by Phase~1 are loaded and merged into a single video-level directed graph.  Node IDs are re-indexed with offsets to avoid collisions across clips.  An entity graph (inverted index mapping entity names to node sets) is reconstructed from all clip-level entities, actions, and scenes.  Subtitles are collected as $(frame\_index, text)$ pairs.  Clip intervals are also extracted to enable annotation-driven per-frame clip loading.

\paragraph{4-Step Query Resolution Pipeline.}
For efficiency, the pipeline deduplicates queries across frames since the same object prompts recur on every frame, only \emph{unique} prompts are processed through the first three steps.  The results are then broadcast back to all entries for the frame-specific final answer.

\paragraph{Step 1: Keyword Extraction.}
One batched VLM call extracts keywords from all unique query prompts.  Each response contains entity names, scene descriptors, and action keywords used for graph retrieval.

\paragraph{Step 2: Cached Node Retrieval.}
Graph node retrieval uses a BGE embedding model with precomputed embeddings (computed once per video).  Entity matching (cosine similarity $> 0.5$) and content re-ranking identify the top-$N_\text{retrieval}=20$ graph nodes per query.  No VLM call is required in this step.

\paragraph{Step 3: Templatised Node Refinement.}
Retrieved nodes are verified using fixed template sub-questions:
(i)~``\textit{Is the `\{object\}' visible or interacted with in this video segment?}''
(ii)~``\textit{Is there a person visible in this video segment?}''
One batched VLM call processes all node-check prompts.  Nodes are ranked by the number of ``yes'' answers.

\paragraph{Step 4: Frame-Specific Final Answer.}
Each (frame, object) pair receives its answer grounded on the \emph{annotation-driven clip} for that specific frame (same temporal segmentation as graph construction).  If no per-frame clip is available, the RAG-selected clip from node refinement is used; as a last resort, the full-video tensor serves as fallback.  Graph-node context (entities, actions, scenes from the top-ranked node) is prepended to the prompt.  One batched VLM call generates all relationship predictions.

\paragraph{Subtitle Context Integration.}
Both the generative and verification prompts are augmented with subtitle context.  Subtitles are sorted by temporal proximity.

\paragraph{LLM Call Summary.}
In total, the RAG method uses:
(1)~one batched keyword extraction (unique prompts);
(2)~cached node retrieval (CPU-only, no VLM call);
(3)~one batched node checking (unique prompts);
(4)~one batched final answer (frame-specific clips);
(5)~one bulk relationship verification (optional, all frames and queries).
All batched calls are chunked into sub-batches of 64.

\begin{table}[t]
\centering
\small
\caption{Comparison of the two pseudo-annotation generation methods for World Scene Graph Generation.}
\label{sup:tab:mllm:comparison}
\begin{tabularx}{\textwidth}{@{}lXX@{}}
\toprule
\textbf{Property}
  & \textbf{Subtitle-Only (Method 1)}
  & \textbf{RAG-Based (Method 2)} \\
\midrule
\textsc{Query scope}
  & All objects $\times$ all frames
  & All objects $\times$ all frames \\
\textsc{Uses precomputed graph?}
  & No (subtitles only)
  & Yes \\
\textsc{Deduplication}
  & N/A (no retrieval steps)
  & Yes (unique prompts for steps 1--3) \\
\textsc{VLM calls}
  & 1 batched + 1 verify
  & 4 batched + 1 verify \\
\textsc{Per-frame clips}
  & Annotation-driven (fallback: full video)
  & Annotation-driven (fallback: RAG clip) \\
\textsc{Grounding signal}
  & Subtitles + video
  & Graph context + subtitles + video \\
\bottomrule
\end{tabularx}
\end{table}

\vspace{2mm}
\noindent\textbf{Implementation Details.} All VLMs are deployed via the vLLM inference engine with tensor parallelism. Key performance optimisations include: (i)~8$\times$ frame subsampling for videos with $>$120 frames, (ii)~query deduplication (RAG only), processing unique prompts across steps~1--3 for an ${\sim}K{\times}$ reduction, (iii)~precomputed graph embeddings shared across queries (RAG only), (iv)~annotation-driven clips matching Phase~1 segmentation, and (v)~chunked batching of all VLM calls into sub-batches of 64 to prevent OOM.

%% file: sup_tex_files/sup_conclusion.tex
\noindent\textbf{Limitations.} ActionGenome4D inherits the indoor, single-person setting of Action Genome, and \PiThree{} may produce noisy geometry for textureless or reflective surfaces. Semantic labels rely on corrected VLM pseudo-labels with possible residual noise, and the object vocabulary is fixed to Action Genome categories.

\vspace{1mm}
\noindent\textbf{Future Work.} Several promising directions emerge from this study.
\emph{Online temporal reasoning:} adapting \FDST{} to streaming, variable-length observation
windows would enable real-time world-state tracking that can be used for applications such as error detection in procedural activities \cite{peddi2024captaincook4ddatasetunderstandingerrors,france2025position}.
\emph{End-to-end 3D grounding:} replacing the current multi-stage geometric pipeline with
unified 3D-aware detectors could improve both efficiency and accuracy. This can be potentially used for tasks such as navigation \cite{france2026chasing,allu2024modular}.
\emph{Open-vocabulary \WSGG{}:} broadening the object and predicate vocabularies through
vision-language grounding would move toward unconstrained scene understanding. This can be potentially used for tasks such as task guidance \cite{rheault2024predictive}.
\emph{Downstream applications:} assessing how world scene graphs benefit activity
recognition~\cite{peddi2024captaincook4ddatasetunderstandingerrors,france2025position,rheault2024predictive}, embodied navigation \cite{france2026chasing,allu2024modular}, and robotic manipulation \cite{xiang2024grasping} remains an important validation step.
\emph{Long-tail predicate balance:} the substantial gap between micro- and macro-averaged F1
scores (e.g.\ 53.3 vs.\ 26.6) indicates that VLMs disproportionately favour frequent
predicates; mitigating this distributional skew is a key open challenge. Techniques from robust optimization literature~\cite{Peddi2022RobustLearning,peddi2022distributionally} can potentially act as data augmentation strategies that could be used to address the issue.